\documentclass[10pt,twocolumn,letterpaper]{article}
 
\usepackage{cvpr}

\definecolor{cvprblue}{rgb}{0.21,0.49,0.74}
\usepackage[pagebackref,breaklinks,colorlinks,allcolors=cvprblue]{hyperref}
\usepackage{graphicx}
\usepackage{subcaption}
\usepackage{algorithm}
\usepackage{tabularx}
\usepackage{algpseudocode}
\usepackage{placeins}
\usepackage{amsmath}
\definecolor{tabfirst}{rgb}{1, 0.7, 0.7}
\definecolor{tabsecond}{rgb}{1, 0.85, 0.7}
\definecolor{tabthird}{rgb}{1, 1, 0.7}
\usepackage{tikz}
\usepackage{pifont}
\newcommand{\redxmark}{{\color{red} \ding{55}}}
\newcommand{\greencheck}{{\color{ForestGreen} \ding{51}}}
\newcommand{\orangetriangle}{{\color[HTML]{FF7F00}$\boldsymbol{\triangle}$}}
 
\usepackage{array}
\newcolumntype{C}[1]{>{\centering\let\newline\\\arraybackslash\hspace{0pt}}m{#1}}
 
\usepackage{color, colortbl}
\definecolor{Gray}{gray}{0.85}
\usepackage{lipsum}
\usepackage{bbm}
\usepackage{verbatim}
\usepackage{multirow}
\usepackage{amsmath}
\usepackage[most]{tcolorbox}
\usepackage{multirow}
\usepackage{algpseudocode}
\usepackage{algorithm}
\usepackage{algorithmicx}
\usepackage{setspace}
\usepackage{graphicx}
\usepackage{adjustbox}
\usepackage{comment}
\usepackage[normalem]{ulem}
\usepackage{bbm}
\usepackage{helvet}
\usepackage{tikz}
\usetikzlibrary{fadings,shapes.arrows,shadows}
\tikzset{arrowstyle/.style={draw=black,single arrow,minimum height=#1, single arrow,
single arrow head extend=.4cm,align=center}}
\usepackage{amsthm}
\usepackage{amssymb}
\usepackage{wrapfig}
\usepackage[utf8]{inputenc}
\usepackage[T1]{fontenc}

\usepackage{multicol}
\usepackage{cellspace}
\usepackage{booktabs}
\usepackage{caption}
\usepackage{bm}
\usepackage{dsfont}
\usepackage{xcolor}
\usepackage{chngcntr}
\newcommand\blfootnote[1]{%
  \begingroup
  \renewcommand\thefootnote{}\footnote{#1}%
  \addtocounter{footnote}{-1}%
  \endgroup
}
 
\def\paperID{7791}
\def\confName{CVPR}
\def\confYear{2026}
 
\begin{document}
 
 
\title{Erasing Thousands of Concepts: Towards Scalable and Practical\\
Concept Erasure for Text-to-Image Diffusion Models}
 
\author{Hoigi Seo$^{*1}$ \qquad Byung Hyun Lee$^{*1}$ \qquad Jaehyun Cho$^{3}$ \qquad Sungjin Lim$^{3}$ \qquad Se Young Chun$^{1,2,3\dag}$ \\
$^1$Dept. of Electrical and Computer Engineering, $^2$INMC, $^3$IPAI \\
Seoul National University, Republic of Korea \\
{\tt\small \{seohoiki3215, ldlqudgus756, tommycjh, sjin.lim, sychun\}@snu.ac.kr}
}
 
\maketitle

\blfootnote{* Authors contributed equally. $\dag$ Corresponding author.}

\begin{abstract}
Large-scale text-to-image (T2I) diffusion models deliver remarkable visual fidelity but pose  safety risks due to their capacity to reproduce undesirable content, such as copyrighted ones. Concept erasure has emerged as a mitigation strategy, yet existing approaches struggle to balance scalability, precision, and robustness, which restricts their applicability to erasing only a few hundred concepts. To address these limitations, we present Erasing Thousands of Concepts (ETC), a scalable  framework capable of erasing thousands of concepts while preserving generation quality. Our method first models low-rank concept distributions via a Student’s t-distribution Mixture Model (tMM). It enables pin-point erasure of target concepts via affine optimal transport while preserving others by anchoring the boundaries of target concept distributions without pre-defined anchor concepts. We then train a Mixture-of-Experts (MoE)–based module, termed MoEraser, which removes target embeddings while preserving the anchor embeddings. By injecting noise into the text embedding projector and fine-tuning MoEraser for recovery, our framework achieves robustness to white-box attack such as module removal. Extensive experiments on over 2,000 concepts across heterogeneous domains and diffusion models demonstrate state-of-the-art scalability and precision in large-scale concept erasure.
\end{abstract}

\section{Introduction}
Text-to-image (T2I) diffusion models generate high-quality, prompt-aligned images~\cite{nichol2022glide, saharia2022photorealistic, rombach2022high, ramesh2022hierarchical,chen2024pixart,comanici2025gemini,frontier2024flux1dev}. Their power also raises safety risks, including undesirable content and violations of portrait or copyright~\cite{dalle2preview2022, rando2022red, schramowski2023safe, somepalli2023diffusion}. To mitigate these risks, recent works adopt fine-tuning to remove unwanted concepts while preserving the rest~\cite{gandikota2023erasing, fan2023salun, gandikota2024unified, lu2024mace, huang2023receler, zhang2024defensive, lee2025concept}, advancing erasure efficacy~\cite{kumari2023ablating, zhang2023forget, gandikota2024unified} and specificity~\cite{lyu2023one, lu2024mace, bui2024erasing}. Despite these improvements, scalability and pin-point erasure remain limited in practice.

Effective concept erasure requires \textit{scalability}-removing many concepts-and \textit{pin-pointness}-erasing only the target while preserving others~\cite{moon2024holistic}. A practical framework must also handle heterogeneous domains. Prior work either tunes the base model~\cite{kumari2023ablating, gandikota2023erasing, heng2024selective, liu2024safetydpo} or adds linear modules~\cite{lu2024mace, lyu2023one}, 
which could potentially distort non-target concepts or limit scalability.
Recent non-linear approaches~\cite{lee2025concept, lee2025localized} improve scalability but still cannot be merged into model weights, can be bypassed by removing the external module, and incur inference cost that grows with the number of erased concepts. In addition, current methods depend on heuristic anchor selection to preserve remaining concepts, lack evaluation on heterogeneous concepts across diverse domains, and lose pin-pointness at large erasure scales.

Here, we propose Erasing Thousands of Concepts (ETC), a framework for scalable concept erasure with minimal alteration to the remaining ones and robustness to module removal. Our approach consists of two stages. 

The first stage models concept distributions. We extract text embeddings for each concept across diverse templates to capture per-concept embedding distributions. Because these embeddings are extremely low rank and heavy-tailed, we apply PCA for compression and fit a Student’s t-mixture model (tMM) \cite{peel2000robust, andrews2011model} for efficiency and fidelity. We then compute an affine optimal transport map (AOT) from the target to the mapping concept distribution, yielding mapping embeddings for the novel concept. Finally, we sample along the boundary of the target distribution to construct anchor embeddings that help preserve remaining concepts.

In the second stage, based on the obtained target, mapping, and anchor embeddings, we train an erasing module based on Mixture-of-Experts (MoE) \cite{shazeer2017outrageously,riquelme2021scaling,lepikhingshard,jiang2024mixtral}, termed \textit{MoEraser}, such that the target embeddings are transformed into the mapping embeddings while the anchor embeddings remain unchanged. We inject low-rank noise into text embedding projection layers of the diffusion model to damage the ability to generate high-fidelity images and fine-tune the MoEraser module to restore the corrupted embeddings. This design ensures that the model cannot generate normal images without our module, effectively increasing robustness against white-box attack (\textit{e.g.} module removal).

We evaluated our method on a wide range of concept-erasing tasks, spanning from the removal of 50 homogeneous concepts to over 2,000 concepts across three heterogeneous domains (\textit{i.e.} celebrity, character, and artistic style). We experimented on both Stable Diffusion v1.4~\cite{rombach2022high} (SDv1.4) and Stable Diffusion v3.5-L~\cite{esser2024scaling} (SDv3.5-L). 
Our comprehensive experiments demonstrate that our method achieves superior performance to prior arts in both relatively small-scale settings in the previous works as well as in our large-scale scenarios.
Our contribution is summarized as:
\begin{itemize}
    \item Proposing Erasing Thousands of Concepts (ETC), a framework for erasing large-scale concepts across diverse domains while preserving generation quality of others.
    \item Realizing pin-point erasing in an anchor concept-free manner by modeling concept distributions as tMMs and mapping them via affine optimal transport (AOT).   
    \item Proposing a novel robust training scheme for our MoE-based module, MoEraser, by integrating a Noise Injection–Restore (NIR) tuning strategy that induces resistance to removal against white-box attack.
    \item Validating our method on both SDv1.4 and SDv3.5-L, making this, to the best of our knowledge, the first work to effectively erase 2,000+ concepts from multiple domains.
\end{itemize}

\section{Related Works}
\textbf{Safe T2I diffusion models.}\,\,
Large-scale generative models risk producing undesirable content~\cite{abid2021persistent, birhane2021large, hutson2021robo}, a concern equally relevant to T2I diffusion models~\cite{rombach2022stable1.4, rombach2022stable2.0, esser2024scaling, chen2024pixart, comanici2025gemini, frontier2024flux1dev}. A prevalent mitigation strategy is data censoring~\cite{dalle2preview2022, schuhmann2022laion, rombach2022stable2.0}, but re-training on filtered datasets is computationally expensive and susceptible to residual biases or incomplete removal~\cite{ryan2022sd12, dixon2018measuring}. Post-generation filters~\cite{rando2022red, laborde2020nsfw, bedapudinudenet} and inference-time controls~\cite{schramowski2023safe, yoonsafree} offer alternatives but depend on pre-defined concept detectors, making them vulnerable to circumvention through altered or bypassed filtering mechanisms~\cite{rando2022red}.

\noindent\textbf{Concept erasing in T2I diffusion models.}
To address these limitations, recent works adopt fine-tuning-based concept erasure: FMN~\cite{zhang2023forget} redirects cross-attention activations, AblCon~\cite{kumari2023ablating} re-trains models to replace target concepts, and ESD~\cite{gandikota2023erasing} aligns target and mapping concepts. While effective at removal, preserving non-target concepts remains critical. SA~\cite{heng2024selective} applies continual-learning regularization~\cite{kirkpatrick2017overcoming, rolnick2019experience, lee2023online, lee2024doubly}, TIME~\cite{orgad2023editing} and UCE~\cite{gandikota2024unified} introduce closed-form preservation updates, and SPM~\cite{lyu2023one} leverages LoRA~\cite{hu2021lora} to retain concepts. Nonetheless, scalability and preservation of unrelated concepts remain challenging.

\noindent\textbf{Scalable concept erasure.}
Building on these efforts, scalability has emerged as a key focus for real-world deployment. MACE~\cite{lu2024mace} extends UCE with LoRA-based modules to erase up to 100 homogeneous concepts, while CPE~\cite{lee2025concept} reveals theoretical limits of linear module updates and demonstrates that non-linearity improves specificity and fidelity in large-scale erasure. SPEED~\cite{li2025speed} leverages null-space priors to minimize interference with retained concepts. Despite this progress, erasing thousands of concepts across heterogeneous domains such as celebrities, characters, and artistic styles remains challenging. A comparative summary is provided in Tab.~\ref{tab:property}.

\begin{table}[]
    \centering
    \resizebox{\linewidth}{!}{  
    \begin{tabular}{c|ccc}
    \toprule
         & Scalable erasing & Non-removable & Anchor \\
         & ($>2,000$) & erasing module & concept-free \\
    \midrule
    MACE~\cite{lu2024mace} & \large\redxmark & \large\greencheck & \large\redxmark \\
    CPE~\cite{lee2025concept} & \orangetriangle & \large\redxmark & \large\redxmark  \\
    SPEED~\cite{li2025speed} & \large\redxmark & \large\greencheck & \large\redxmark  \\
    ETC \textbf{(Ours)}     & \large\greencheck & \large\greencheck & \large\greencheck  \\
    \bottomrule
    \end{tabular}
    }
    \vspace{-1em}
    \caption{Property comparison of concept erasing methods.}
    \label{tab:property}
    \vspace{-1.5em}
\end{table}

\section{Method}
\subsection{Modeling concepts into distribution}\label{sec:text_modeling}
We model each concept as a distribution over text embeddings, which both enables explicit mapping between concepts and supports boundary-based identification of embeddings to erase targets and preserve anchors. With these models, we eliminate the need for predefined anchor concepts used in prior work~\cite{srivatsan2025stereo, lee2025localized, lee2025concept}. This design is motivated by the fact that a word’s embedding varies with the sentence-level semantics in which it appears, and prior studies interpret such variability using distributional modeling \cite{vilnis2014word, ath2017multimodal, ath2018probabilistic}.

Concretely, for a concept $c$, we construct a concept embedding matrix $X_c \in \mathbb{R}^{d \times N}$ by feeding thousands of templates (i.e., \textit{``As the snow covered the path, \{\} lit a small fire’’}) combined with the concept word into the text encoder, where $N$ is the number of templates and $d$ is the embedding dimension. Since $d$ is typically large ($\ge768$), directly leveraging $X_c$ is computationally inefficient. Observing that embeddings for a concept exhibit low-rank~\cite{lee2025localized}, we apply principal component analysis (PCA) to obtain a compact representation $Z_c \in \mathbb{R}^{d' \times N}$ from $X_c$ with basis matrix $V_c \in \mathbb{R}^{d' \times d}$, where $d' \ll d$.

\begin{figure}
    \centering
    \includegraphics[width=\linewidth]{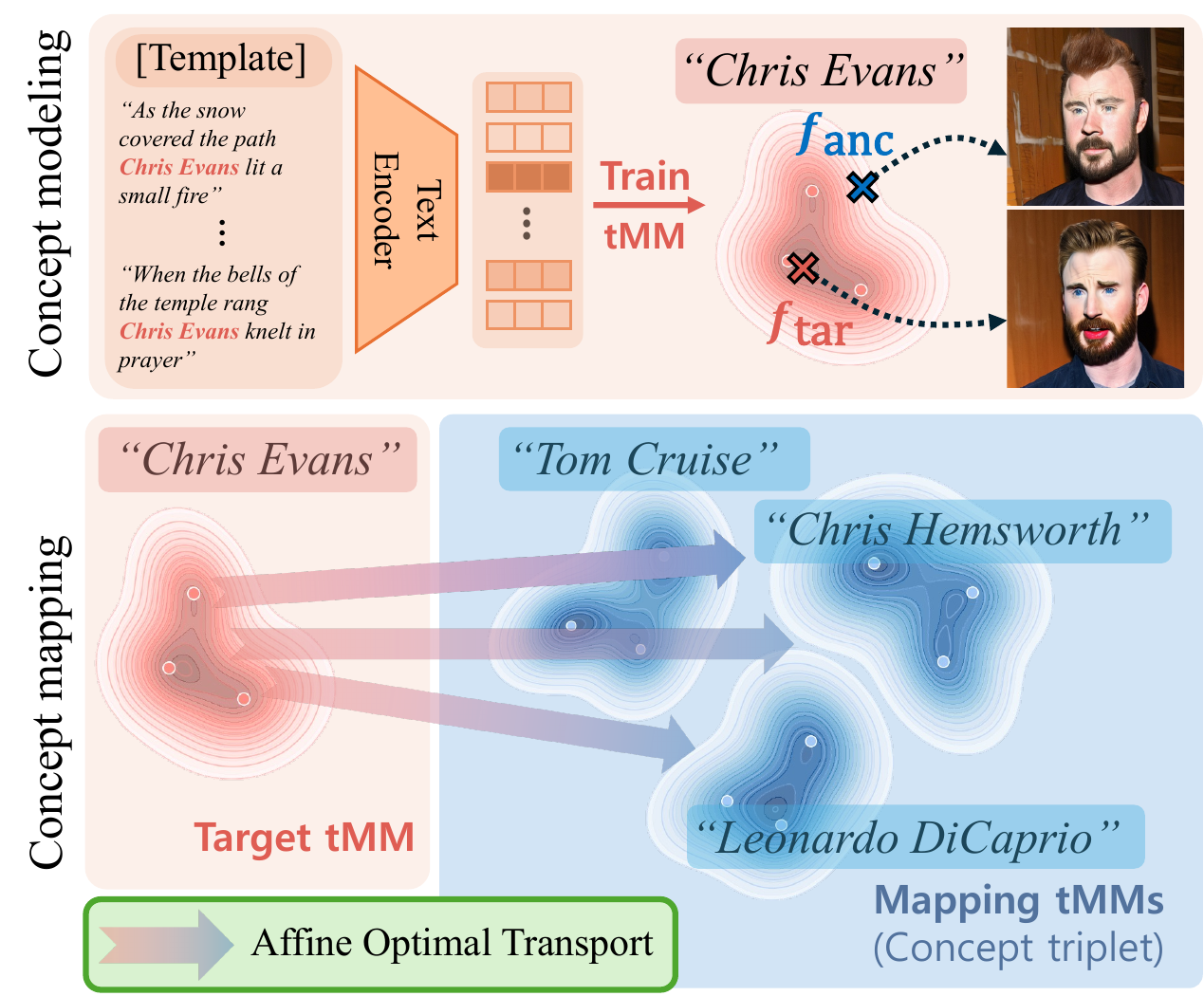}
    \vspace{-2em}
    \caption{\textbf{Concept distribution modeling and mapping.} (Top) Concept embeddings from templates are modeled with a Student's t-distribution Mixture Model (tMM). Embeddings in high-probability regions serve as target embeddings ($f_\text{tar}$), while those in low-probability regions serve as anchoring embeddings ($f_\text{anc}$). (Bottom) The target concept distribution is mapped to a merged distribution via Affine Optimal Transport (AOT).}
    \label{fig:text_modeling}
    \vspace{-1em}
\end{figure}

As illustrated in the top panel of Fig.~\ref{fig:text_modeling}, we then fit a Student’s t-distribution mixture model (tMM)~\cite{peel2000robust, andrews2011model} to $Z_c$:
\begin{equation}
P_c(z)=\sum_{i=1}^k \pi_{c, i}\cdot t(z|\mu_{c,i},\Sigma_{c,i}, \nu_{c,i}),
\end{equation}
where $\mu_{c,\cdot}$, $\Sigma_{c,\cdot}$, and $\nu_{c,\cdot}$ denote the mean, covariance, and degrees of freedom for each component, $k$ is the number of components, and $\pi_{c,\cdot}$ are the mixture weights. This distributional view captures contextual variability while providing a principled mechanism for constructing mappings across concepts and for selecting targets and anchors from probability boundaries without curated anchor sets.

\begin{figure}
    \centering
    \includegraphics[width=\linewidth]{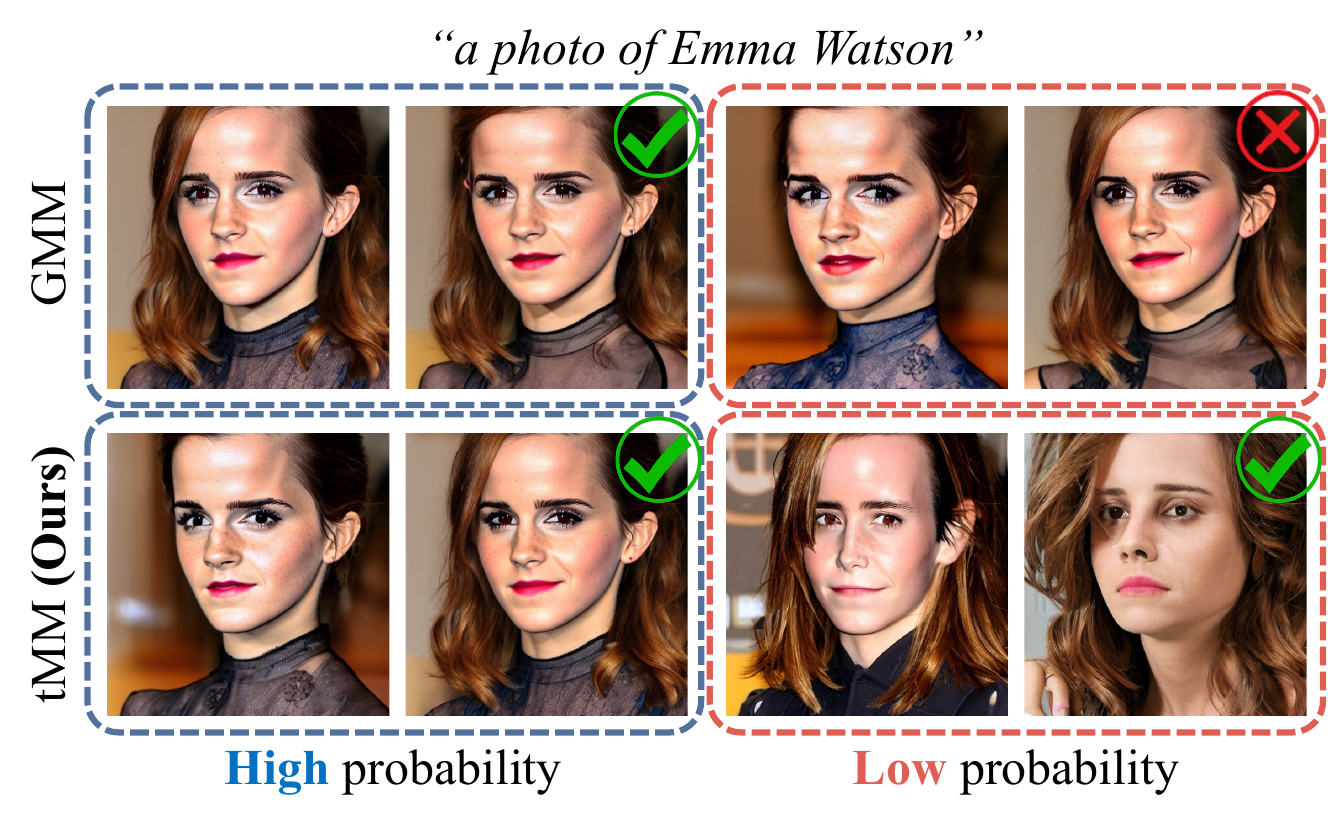}
    \vspace{-2em}
    \caption{\textbf{Qualitative results: tMM vs. GMM.} Distributions for \textit{``Emma Watson’’} are fitted with a Student-t Mixture Model (tMM) and a Gaussian Mixture Model (GMM). From each, we sample embeddings from high-probability cumulative intervals (0.05, 0.1) and low-probability intervals (0.9, 0.95), then generate images using SDv1.4. GMM preserves strong concept presence even for low-probability samples, whereas tMM attenuates it proportionally to probability. Degrees of freedom for each tMM are set to 2.}
    \label{fig:tmm_rationale}
    \vspace{-1em}
\end{figure}

Then why do we adopt the tMM, instead of the conventional Gaussian Mixture Model (GMM)? As shown in Fig.~\ref{fig:tmm_rationale}, the concept embedding distribution empirically exhibits a \textit{heavy-tailed behavior}. Intuitively, since the embeddings are inherently ``in-distribution'', genuine out-of-distribution samples that reflect true variability are scarce. The tMM naturally models heavier tails, enabling better modeling of variability within the concept.

\begin{tcolorbox}[before skip=2mm, after skip=0.0cm, boxsep=0.0cm, middle=0.0cm, top=0.1cm, bottom=0.1cm]
\textbf{Remark 1.} \textit{Concept distribution modeling.} A concept in a T2I diffusion model can be effectively represented as a Student's t-distribution mixture model.
\end{tcolorbox}

\subsection{Optimal transport for efficient mapping}
\begin{figure}[!t]
    \centering
    \includegraphics[width=\linewidth]{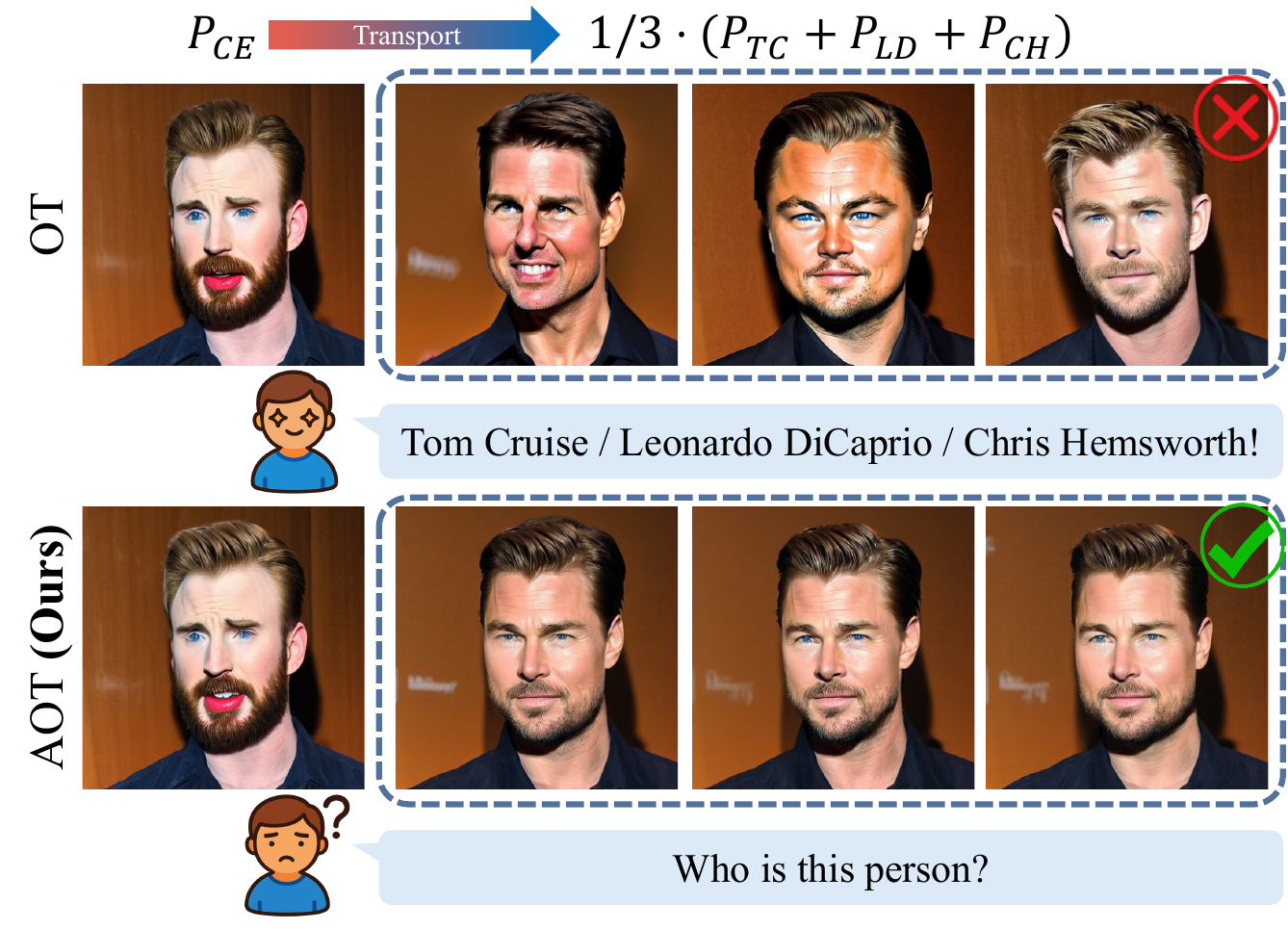}
    \vspace{-2em}
    \caption{\textbf{Qualitative rationale for AOT.} We formulate the distributions of \textit{``Chris Evans, ``Tom Cruise,'' ``Leonardo DiCaprio,''} and \textit{``Chris Hemsworth,''} as $P_{CE}$, $P_{TC}$, $P_{LD}$, and $P_{CH}$, respectively. We merge $P_{TC}$, $P_{LD}$, and $P_{CH}$ and subsequently map embeddings using optimal transport (OT) and affine optimal transport (AOT), followed by image generation with SDv1.4. Whereas OT yields images aligned with the modes of the mapped concepts, AOT 
    results in \textbf{novel visual features},
    thereby improving safety.}
    \label{fig:aot_rationale}
    \vspace{-1em}
\end{figure}

Recent studies have shown that mapping the target concept to an appropriate alternative concept significantly affects both erasing effectiveness and preservation quality~\cite{gong2024reliable, bui2025fantastic}. Given the known distributions of the target and mapping concepts, we model the concept mapping via optimal transport. Specifically, we formulate the affine distribution mapping from concept $p$ to concept $q$ as
\begin{equation}
    T_{p\mapsto q}(z, V_{pq})=AV_{pq}z+b,\; z\sim P_p,
\end{equation}
where $V_{pq}\triangleq V_qV_p^\top$ to ensure that the embeddings are represented in the correct space. Although affine transport does not guarantee perfect distributional alignment, (1) its linearity simplifies network training, and (2) when mapping to a merged multi-concept distribution, the result tends to align with a third concept, since the mapped distribution lies between the merged concepts, enhancing safety (Fig.~\ref{fig:aot_rationale}).

Under the affine model, we seek the optimal transport, which is formulated as the following optimization problem:
\begin{equation}
    (A^*, b^*)\in \arg\min_{A,b}W_2\big((AV_{pq}z+b)_\#P_p, P_q\big),
\end{equation}
where $(AV_{pq}z+b)_\# P_p$ is the pushforward of $P_p$ under the affine map $z\mapsto AV_{pq}z+b$ with the basis transformation matrix $V_{pq}$, $W_2$ denotes the 2-Wasserstein distance. Through this optimization, we obtain the optimal transport plan $T^*$ using the estimated parameters $(A^*, b^*)$. In our implementation, the mapping concept $q$ is represented as a merged distribution of three concepts (\textit{i.e. ``concept triplet''}), as illustrated in the bottom panel of Fig.~\ref{fig:text_modeling}.

\begin{tcolorbox}[before skip=2mm, after skip=0.0cm, boxsep=0.0cm, middle=0.0cm, top=0.1cm, bottom=0.1cm]
\textbf{Remark 2.} \textit{Affine Optimal Transport (AOT).} AOT efficiently maps the target concept distribution to an anonymous, novel concept distribution.
\end{tcolorbox}

\subsection{Training scheme for scalable concept erasing}
For training, we extract three embeddings: the target to erase, the mapping for replacement, and the anchor to preserve unrelated concepts. Prior works~\cite{lu2024mace, srivatsan2025stereo, lee2025localized, lee2025concept} show that anchor choice strongly affects performance, with anchors closer to the target yielding better results~\cite{lu2024mace, lee2025concept}. However, heuristic or manual construction of such anchors does not scale. As detailed in \cref{sec:text_modeling} and \cref{fig:text_modeling}, we instead sample anchor embeddings from low-probability regions of the target concept distribution, thereby eliminating the need for curated anchor concept sets.
\begin{equation}
    z_\text{tar}\sim P^\text{(high)}_\text{tar},\; z_\text{anc}\sim P^\text{(low)}_\text{tar},\; z_\text{map}=T_{\text{tar}\mapsto \text{map}}(z_\text{tar}).
\end{equation}
At this stage, each $z$ is represented in its respective projected basis. To obtain the final model input, we project back into the original embedding space with its basis.

\begin{figure*}[!t]
    \centering
    \begin{subfigure}{0.57\linewidth}
        \includegraphics[width=\textwidth]{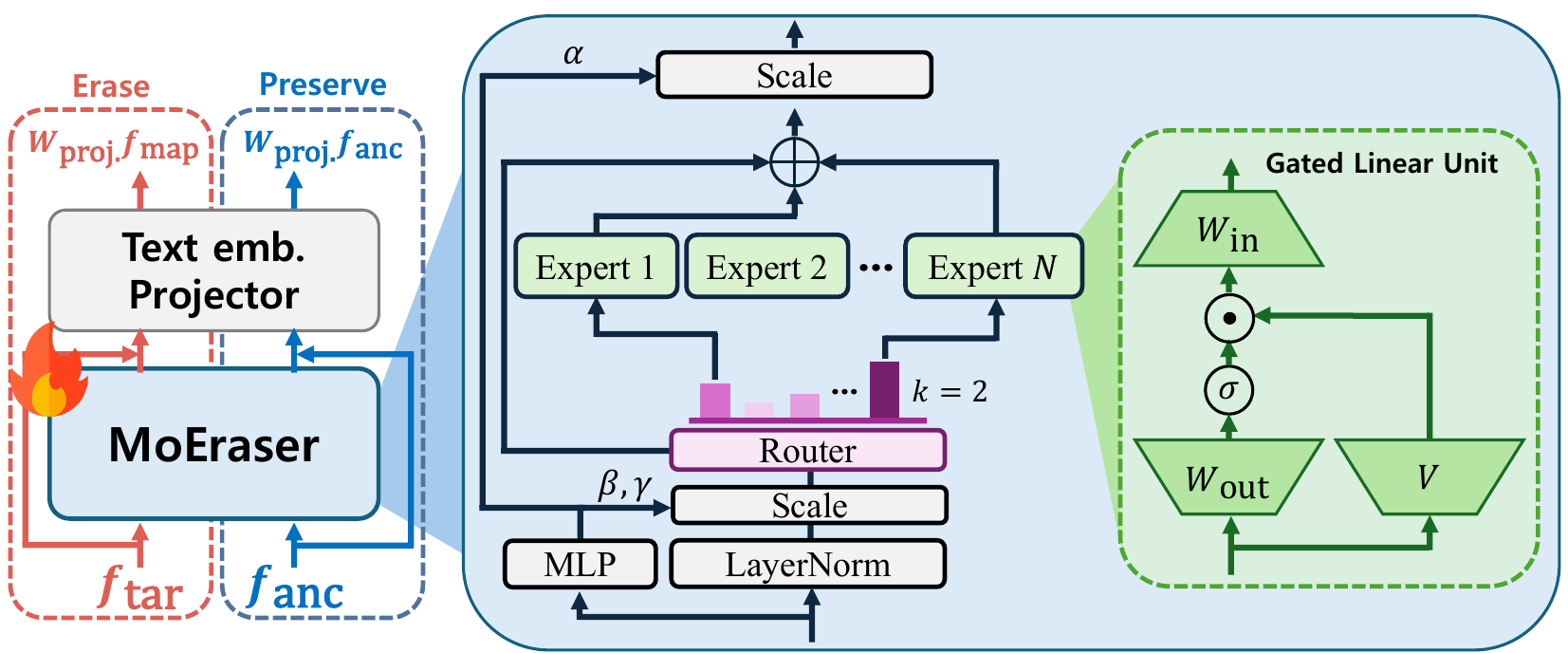}
        \caption{Illustration of our MoEraser architecture.}
        \label{fig:architecture}
    \end{subfigure}
    \hfill
    \begin{subfigure}{0.42\linewidth}
        \includegraphics[width=\textwidth]{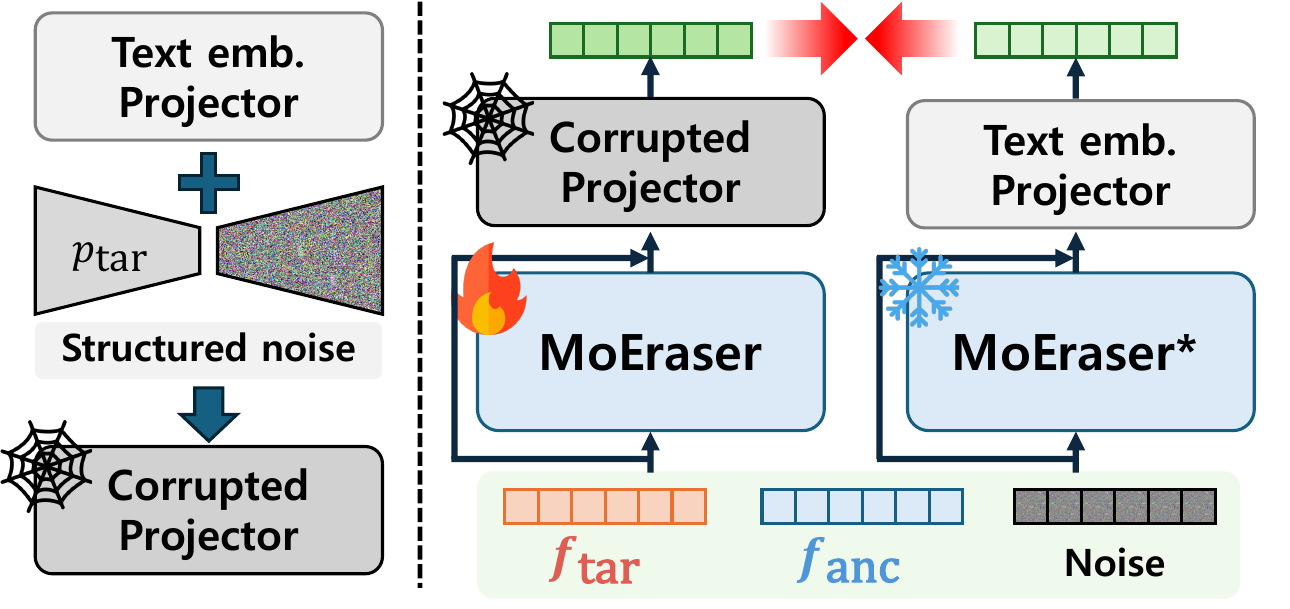}
        \caption{Illustration of noise injection-restore (NIR) tuning.}
        \label{fig:noise_training}
    \end{subfigure}
    \vspace{-1em}
    \caption{\textbf{MoEraser architecture and training.} (a) A MoE with GLU experts scales to heterogeneous domain concepts; training maps $f_\text{tar}$ to $f_\text{map}$ while leaving $f_\text{anc}$ unchanged. (b) To make the module non-removable, we inject structured noise into the text embedding projector and fine-tune the safety module to reconstruct the original embedding, improving robustness to white-box attacks such as module removal.}
    \label{fig:safety_model_training}
    \vspace{-1em}
\end{figure*}

With these samples, we train a network to learn the transport plan for the target concept while enforcing identity on non-target concepts.
A linear module or feed-forward network (FFN) may be employed as the network, but their efficiency and adaptability across domains could be limited. Instead, we adopt a Mixture-of-Experts (MoE) architecture \cite{riquelme2021scaling,lepikhingshard,jiang2024mixtral} with a router, rather than scaling a single model. We denote this as MoEraser, illustrated in \cref{fig:architecture}.
Following gating-based designs~\cite{lee2025concept}, each expert uses a Gated Linear Unit (GLU)~\cite{dauphin2017language}. 
MoEraser is trained to ensure that when the target embedding $f_{\text{tar}}$ is passed through the module and text embedding projection $W_{\text{proj.}}\in\mathbb{R}^{d_\text{out}\times d_\text{in}}$, the resulting representation aligns with $W_{\text{proj.}}f_{\text{map}}$, while the anchor $f_{\text{anc}}$ unchanged.
Then, we define the overall loss for erasing target concepts while preserving others as follows:
\begin{align}
    &\mathcal{L}_\text{Erase} = 
    \|W_\text{proj.}(\text{MoEraser}(f_\text{tar})+f_\text{tar}) -  W_\text{proj.}f_\text{map}  )\|_2^2 \nonumber \\
    &+ \lambda \cdot \|W_\text{proj.}(\text{MoEraser}(f_\text{anc})+f_\text{anc}) - W_\text{proj.}f_\text{anc}\|_2^2,
\end{align}
where $\lambda$ is a coefficient to balance erasing and preserving.

\begin{tcolorbox}[before skip=2mm, after skip=0.0cm, boxsep=0.0cm, middle=0.0cm, top=0.1cm, bottom=0.1cm]
\textbf{Remark 3.} \textit{Training scheme.} With concept distribution modeling and MoE–based architecture, we can train scalable, anchor-free concept-erasing module.
\end{tcolorbox}

\begin{figure}[!t]
    \centering
    \includegraphics[width=\linewidth]{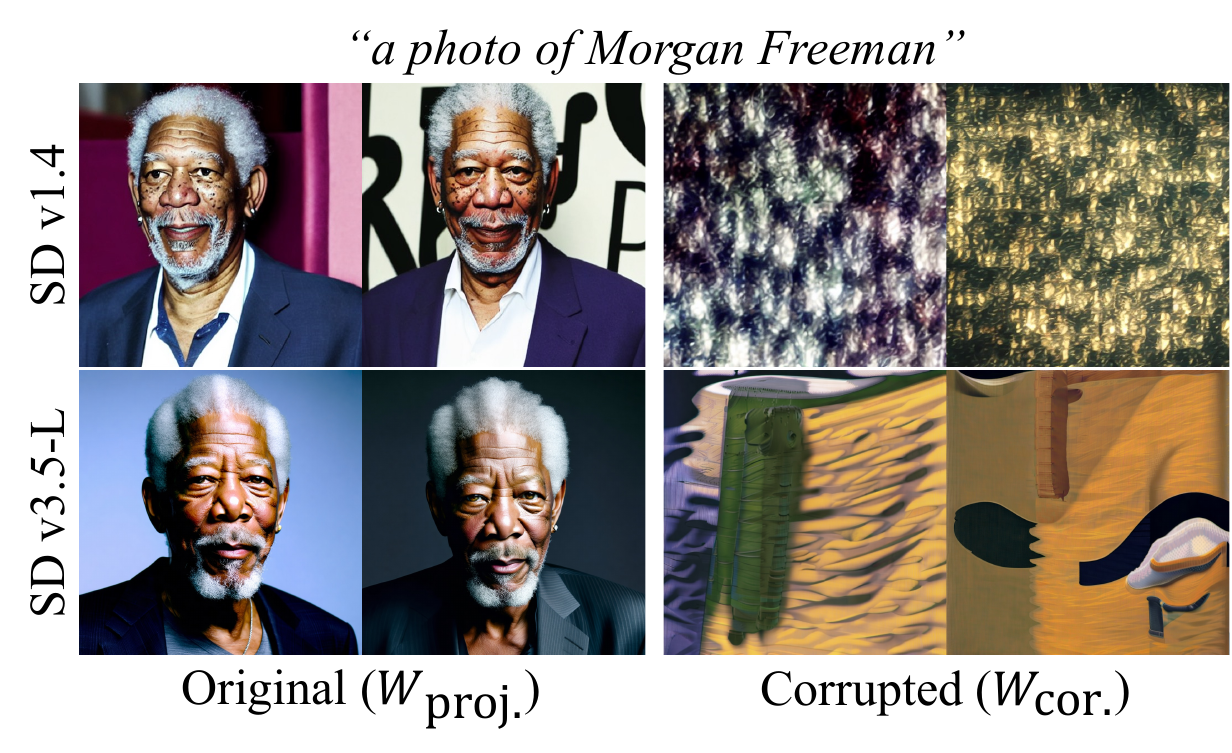}
    \vspace{-2em}
    \caption{\textbf{Qualitative rationale on NIR.} We generated images with the prompt \textit{``a photo of Morgan Freeman''} using the original text-embedding projection $W_\text{proj.}$ (left) and a corrupted weight $W_\text{cor.}$ (right) on SDv1.4 and SDv3.5-L. When sufficient noise is injected, the models fail to produce high-fidelity images; without the MoEraser module to restore the generation, the model becomes unusable, enhancing robustness to white-box attacks.}
    \label{fig:nir_rationale}
    \vspace{-1em}
\end{figure}

\subsection{Noise Injection-Restore tuning}
Prior concept erasure methods typically train all parameters or integrate linear modules with model weights~\cite{lu2024mace, li2025speed}. Recent studies reveal the limitations of such linear modules and promote non-linear alternatives~\cite{huang2023receler, lee2025concept, lee2025localized}, which, despite improved performance, remain vulnerable to malicious removal. We introduce a Noise Injection–Restore (NIR) scheme that perturbs model weights with noise and fine-tunes an erasure module to reconstruct the original embeddings. Removing this module disables normal image generation, thereby enforcing retention of the module.

Specifically, we inject structured noise into the projection linear layer, which is present in all T2I diffusion models, to obtain the corrupted weight $W_\text{cor.}$ as follows.
\begin{equation}
    W_\text{cor.} = W_\text{proj.} + \alpha_\text{noise} \cdot e p_\text{tar}^\top,
\end{equation} 
where $p_\text{tar}\in\mathbb{R}^{d_\text{in}\times r}$ is the top-$r$ PCA eigenvectors of target embeddings and $e\in\mathbb{R}^{d_\text{out}\times r}$ is a random matrix from $\mathcal{N}(0,I)$. The scalar $\alpha_\text{noise}$ controls noise magnitude. This structured noise perturbs weights more along target components, degrading image quality (Fig.~\ref{fig:nir_rationale}). MoEraser is trained to restore corrupted embeddings and resist white-box removal, as shown in Fig.~\ref{fig:noise_training}.

With the corrupted projector, we fine-tune the MoEraser with the following objective $\mathcal{L}_\text{NIR}$.
\begin{equation}
\begin{split}
    \mathcal{L}_\text{NIR}=
    &\ \|W_\text{cor.}(\text{MoEraser}(f) + f)\\
    &-W_\text{proj.}(\text{MoEraser}^*(f)+f)\|_2^2,
\end{split}
\end{equation}
where $f$ is sampled from $\{f_\text{tar}, f_\text{anc}, \epsilon\}$, and MoEraser$^*$ denotes the frozen pre-trained MoEraser module. Note that $\epsilon$ is noise sampled from a Gaussian distribution with the same dimensionality as $f_\text{tar}$ and $f_\text{anc}$, which is used to effectively preserve the priors that the diffusion model can generate beyond the target and anchor concepts. 
The overall training pipeline
is illustrated on the right side of Fig.~\ref{fig:noise_training}.
Further details on the method components are provided in Sec.~\ref{supp:method_and_ablations}.

\begin{tcolorbox}[before skip=2mm, after skip=0.0cm, boxsep=0.0cm, middle=0.0cm, top=0.1cm, bottom=0.1cm]
\textbf{Remark 4.} \textit{Noise Injection-Restore (NIR) tuning.} NIR makes MoEraser robust to removal, as removing the module leads to abnormal output images.
\end{tcolorbox}

\section{Experiments}
We provide further implementation details on hyperparameters for tMM modeling, model architecture, and other training configurations in Sec.~\ref{supp:implementation}.

\subsection{Experimental setups}
We evaluate ETC in three settings: large-scale concept erasure across heterogeneous domains, extension to advanced diffusion models, and small-scale erasure on 50 celebrities. The first and third use Stable Diffusion v1.4 (SDv1.4) \cite{rombach2022high}, and the extension uses Stable Diffusion v3.5-Large (SDv3.5-L)\cite{esser2024scaling}. We compare against seven baselines: FMN~\cite{zhang2023forget}, ESD~\cite{gandikota2023erasing}, UCE~\cite{gandikota2024unified}, MACE~\cite{lu2024mace}, CPE~\cite{lee2025concept}, SAFREE~\cite{yoonsafree}, and SPEED~\cite{li2025speed}. FMN and ESD are evaluated only in the small-scale setting. Since UCE, MACE, and CPE target cross-attention layers, SAFREE and SPEED are compared on the MMDiT-based SDv3.5-L. For details of each setups, please refer to Sec.~\ref{supp:exp_setup} and~\ref{supp:eval_metrics}. 

\subsection{Concept erasure at scale}

\noindent\textbf{Dataset.}\,\,
We conducted a large-scale experiment to simultaneously erase concept spanning three heterogeneous domains: celebrities, artistic styles, and characters. Specifically, we constructed concept pools consisting of 2,306, 1,734, and 1,016 concepts for each domain respectively. Then, we erased 949 celebrities, 693 artistic styles, and 430 characters, resulting in a total of 2,072 concepts from the concept pools. For remaining concepts, we used 325 celebrities, 430 artistic styles, and 279 characters to evaluate concept preservation performance. We selected mapping concepts to be mutually exclusive with respect to the remaining concepts for fair evaluations.

\noindent\textbf{Evaluation setup.}\,\,
Celebrity concepts can be partially evaluated with face-recognition networks, but artistic styles and characters lack reliable automatic assessment due to subjective semantics. Prior work often uses CLIP Score~\cite{hessel2021clipscore} and LPIPS~\cite{zhang2018unreasonable} to quantify representational change, but these can mistake over-erasure or general image degradation for improved erasure, distorting the evaluation of pinpoint removal at scale. To better match human perception, we conducted large-scale user studies. For each image, participants judged (1) whether the target concept was maintained and (2) whether image quality was preserved.
Based on the proportion of ``Yes'' responses to these two questions, we refer the proportion for the first and second criterion as Concept Retention Score (\textbf{CRS}) and Quality Score (\textbf{QS}), respectively. Importantly, we employed \textbf{the harmonic mean} of the CRS to comprehensively assess both target concept erasure and remaining concept preservation, which has been widely used in prior works \cite{lu2024mace, lee2025localized, li2025speed, amara2025erasing, huang2023receler}:
\begin{align}
H_{0} = \frac{2}{ (1-\text{CRS}_{\text{t}})^{-1} + (\text{CRS}_{\text{r}})^{-1} }
\end{align}
where $\text{CRS}_{\text{t}}$ and $\text{CRS}_{\text{r}}$ are for target and remaining concepts respectively.
For the study, we collected 8,120 responses from 203 users. To ensure fairness, we built platform that anonymously samples 40 images uniformly at random from a pool of 20,143 images across domains and methods for the evaluation. Further platform details and an example are provided in the supplementary materials.

\begin{table*}[t]
    \caption{\textbf{Quantitative results across diverse domains using SDv1.4 (Top) and SDv3.5 (Bottom).} We report the user study ``Yes’’ rate for concept preservation (CRS) and image quality (QS) for both target and remaining concepts on three domains - Celebrities, Artistic Styles, and Characters. To jointly evaluate target concept removal and preservation of remaining concepts, we provide the harmonic mean metric $H_0$. Numbers in parentheses indicate the number of concepts. $\uparrow/\downarrow$ indicates that higher/lower values correspond to better performance.}
    \label{tab:main_quanti}
    \resizebox{\textwidth}{!}{
    \begin{tabular}{c|cc|cc|c|cc|cc|c|cc|cc|c}
    \toprule
    \multirow{3}{*}{Method} 
    & \multicolumn{5}{c|}{Celebrities} 
    & \multicolumn{5}{c|}{Artistic Style} 
    & \multicolumn{5}{c}{Characters} \\
    
    \cmidrule(lr){2-16}

    & \multicolumn{2}{c|}{Target (949)}
    & \multicolumn{2}{c|}{Remain (325)}
    & \multirow{2}{*}{$H_0\uparrow$}
    & \multicolumn{2}{c|}{Target (693)}
    & \multicolumn{2}{c|}{Remain (430)}
    & \multirow{2}{*}{$H_0\uparrow$}
    & \multicolumn{2}{c|}{Target (430)}
    & \multicolumn{2}{c|}{Remain (279)}
    & \multirow{2}{*}{$H_0\uparrow$} \\ 
    \cmidrule(lr){2-5}
    \cmidrule(lr){7-10}
    \cmidrule(lr){12-15}

    & CRS$_{t}$$\downarrow$ & QS$\uparrow$
    & CRS$_{r}$$\uparrow$ & QS$\uparrow$
    &
    & CRS$_{t}$$\downarrow$ & QS$\uparrow$
    & CRS$_{r}$$\uparrow$ & QS$\uparrow$
    &
    & CRS$_{t}$$\downarrow$ & QS$\uparrow$
    & CRS$_{r}$$\uparrow$ & QS$\uparrow$ \\

    \midrule
    UCE \cite{gandikota2024unified} 
    & \cellcolor{tabthird}0.200 & 0.311 & 0.278 & 0.291 & 0.413 & 0.254 & 0.508 & 0.240 & 0.520 & 0.363 & 0.207 & 0.393 & 0.140 & 0.367 & 0.238 \\

    SAFREE \cite{yoonsafree}
    & 0.879 & \cellcolor{tabsecond}0.874 & \cellcolor{tabfirst}0.901 & \cellcolor{tabsecond}0.879 & 0.214 & 0.831 & \cellcolor{tabsecond}0.870 & \cellcolor{tabfirst}0.866 & \cellcolor{tabfirst}0.900 & 0.283 & 0.846 & \cellcolor{tabsecond}0.809 & \cellcolor{tabfirst}0.843 & \cellcolor{tabfirst}0.921 & 0.261 \\

    SPEED \cite{li2025speed}
    & 0.216 & \cellcolor{tabthird}0.647 & 0.217 & 0.602 & 0.340 & \cellcolor{tabsecond}0.143 & 0.714 & 0.194 & 0.725 & 0.316 & \cellcolor{tabfirst}0.098 & 0.515 & 0.127 & 0.633 & 0.222 \\

    MACE \cite{lu2024mace} 
    & 0.396 & 0.613 & 0.468 & \cellcolor{tabthird}0.701 & \cellcolor{tabthird}0.527 & 0.310 & 0.704 & 0.317 & 0.671 & \cellcolor{tabthird}0.434 & 0.162 & 0.473 & 0.162 & 0.643 & \cellcolor{tabthird}0.272 \\

    CPE \cite{lee2025concept}
    & \cellcolor{tabsecond}0.164 & 0.436 & \cellcolor{tabthird}0.544 & 0.667 & \cellcolor{tabsecond}0.659 & \cellcolor{tabthird}0.224 & \cellcolor{tabthird}0.784 & \cellcolor{tabthird}0.600 & \cellcolor{tabthird}0.777 & \cellcolor{tabsecond}0.677 & \cellcolor{tabsecond}0.115 & \cellcolor{tabthird}0.532 & \cellcolor{tabthird}0.430 & \cellcolor{tabthird}0.735 & \cellcolor{tabsecond}0.579 \\
    
    \textbf{Ours}
    & \cellcolor{tabfirst}0.099 & \cellcolor{tabfirst}0.895 & \cellcolor{tabsecond}0.688 & \cellcolor{tabfirst}0.936 & \cellcolor{tabfirst}0.780 & \cellcolor{tabfirst}0.130 & \cellcolor{tabfirst}0.894 & \cellcolor{tabsecond}0.719 & \cellcolor{tabsecond}0.829 & \cellcolor{tabfirst}0.787 & \cellcolor{tabthird}0.130 & \cellcolor{tabfirst}0.814 & \cellcolor{tabsecond}0.719 & \cellcolor{tabsecond}0.919 & \cellcolor{tabfirst}0.787 \\

    \bottomrule
    \end{tabular}
    }
    \centering
    \resizebox{\textwidth}{!}{
    \begin{tabular}{c|cc|cc|c|cc|cc|c|cc|cc|c}
    \toprule
    \multirow{3}{*}{Method} 
    & \multicolumn{5}{c|}{Celebrities} 
    & \multicolumn{5}{c|}{Artistic Style} 
    & \multicolumn{5}{c}{Characters} \\
    
    \cmidrule(lr){2-16}

    & \multicolumn{2}{c|}{Target (250)}
    & \multicolumn{2}{c|}{Remain (133)}
    & \multirow{2}{*}{$H_0\uparrow$}
    & \multicolumn{2}{c|}{Target (151)}
    & \multicolumn{2}{c|}{Remain (59)}
    & \multirow{2}{*}{$H_0\uparrow$}
    & \multicolumn{2}{c|}{Target (114)}
    & \multicolumn{2}{c|}{Remain (80)}
    & \multirow{2}{*}{$H_0\uparrow$} \\ 
    \cmidrule(lr){2-5}
    \cmidrule(lr){7-10}
    \cmidrule(lr){12-15}

    & CRS$_{t}$$\downarrow$ & QS$\uparrow$
    & CRS$_{r}$$\uparrow$ & QS$\uparrow$
    &
    & CRS$_{t}$$\downarrow$ & QS$\uparrow$
    & CRS$_{r}$$\uparrow$ & QS$\uparrow$
    &
    & CRS$_{t}$$\downarrow$ & QS$\uparrow$
    & CRS$_{r}$$\uparrow$ & QS$\uparrow$ \\

    \midrule
    SAFREE \cite{yoonsafree}
    & 0.948 & 0.957 & 0.960 & 0.960 & 0.099 & 0.939 & 0.921 & 0.921 & 0.921 & 0.115 & 0.942 & 0.969 & 0.938 & 0.938 & 0.109\\

    SPEED \cite{li2025speed}
    & 0.425 & 0.192 & 0.425 & 0.192 & 0.502 & 0.129 & 0.195 & 0.195 & 0.427 & 0.319 & 0.465 & 0.172 & 0.375 & 0.212 & 0.441 \\

    \textbf{Ours}
    & 0.231 & 0.729 & 0.729 & 0.854 & \textbf{0.748} & 0.281 & 0.800 & 0.740 & 0.810 & \textbf{0.616} & 0.117 & 0.693 & 0.693 & 0.813 & \textbf{0.777} \\

    \bottomrule
    \end{tabular}
    }
\end{table*}

\noindent\textbf{Results.}\,\,
We report quantitative erasure and preservation results for erasing 2,079 concepts in the top block of \cref{tab:main_quanti}. For target erasure, ETC achieves the best performance in Celebrities and Artistic Style while maintaining image quality; in Characters, ETC attains comparable accuracy to prior methods without degrading quality, whereas others reduce quality. For remaining concepts, ETC matches or exceeds SAFREE in CRS$_r$ and QS while SAFREE struggles with erasure, demonstrating strong preservation. The harmonic mean $H_0$ summarizes the erasing–preservation trade-off and indicates that ETC performs well on both objectives.

\subsection{Extension to advanced diffusion model}
\noindent\textbf{Dataset and evaluation setup.}\,\,
We further evaluated our method on SDv3.5-L model. To select concepts, we used the same concept pools as those employed in SDv1.4. In this setting, we erased 250 celebrities, 151 artistic styles, and 114 characters, totaling 515 concepts, and used 133 celebrities, 59 artistic styles, and 80 characters as remaining concepts for concept preservation performance. The evaluation protocol followed the same protocol as in the SDv1.4 experiments with 6,546 image pools, assessing both the efficacy of the erasure and the fidelity of the images. For the study, we collected 6,120 responses from 153 users.

\begin{figure*}
    \centering
    \includegraphics[width=\linewidth]{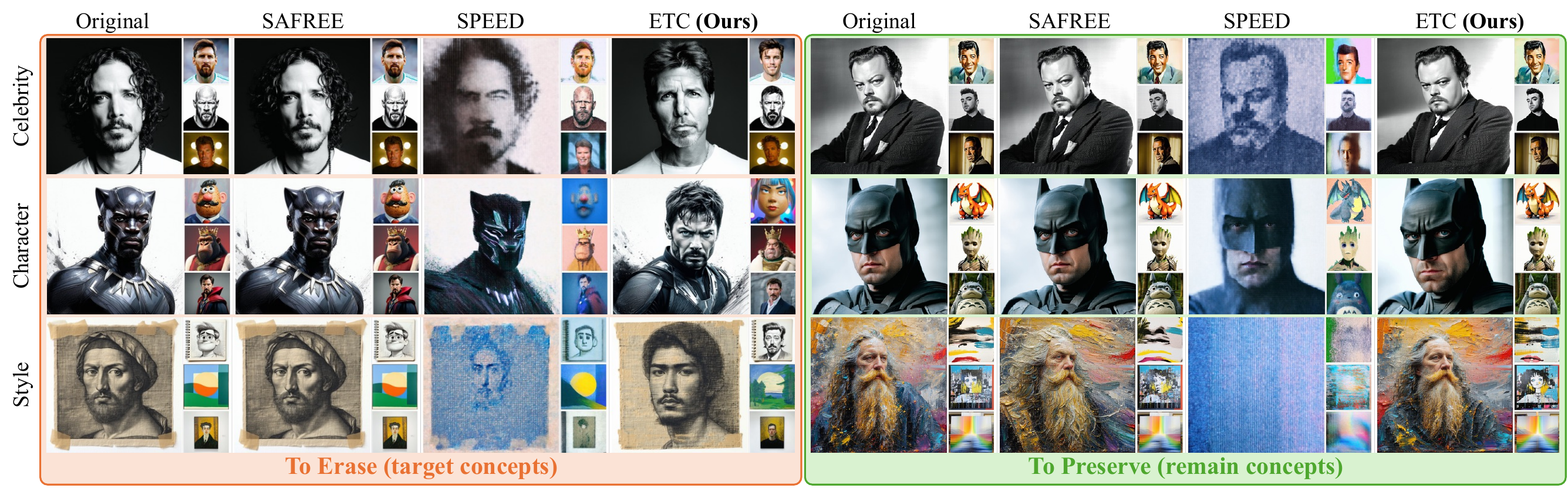}
    \vspace{-2em}
    \caption{\textbf{Qualitative results on erasing 515 concepts from SDv3.5-L.} SAFREE reproduces the original image and fails to remove the target concept. SPEED removes the target concept but degrades fidelity, and this degradation also affects the remaining concepts. In contrast, ETC achieves accurate concept erasure while preserving remaining concepts on the SDv3.5-L, demonstrating applicability.}
    \vspace{-1em}
    \label{fig:sd35_quali}
\end{figure*}

\noindent\textbf{Results.}\,\,
Bottom results in \cref{tab:main_quanti} show consistent trends across domains: SAFREE fails to erase targets with persistently high CRS$_t$. SPEED seems to erases targets but drastically degrades image quality for both target and remaining concepts. ETC achieves strong erasure while preserving remaining concepts and overall image quality in both target and remaining concepts as seen in its $H_0$ and QS score. Qualitative results in \cref{fig:sd35_quali} support this results, with ETC preserving composition and color while altering only identity, showing its pin-pointness and broad applicability.

\begin{table}
    \caption{\textbf{Quantitative results on 50 celebrity concepts erasure.} We used GCD accuracy in percentage (ACC${_t}$ for target and ACC${_r}$ for remaining concepts) and harmonic mean $H_0$. We also measured FID and KID (scaled by 100) for remaining concepts.}
    \label{tab:small_scale_quanti}
    \centering
    \resizebox{\linewidth}{!}{
    \begin{tabular}{c|c|ccc|cc|cc}
    \toprule
    \multirow{3}{*}{Method} 
    & \multicolumn{1}{c|}{Target} 
    & \multicolumn{7}{c}{Remaining} \\
    
    \cmidrule(lr){2-9}

    & \multicolumn{1}{c|}{50 Celeb.}
    & \multicolumn{3}{c|}{100 Celeb.}
    & \multicolumn{2}{c|}{100 Styles}
    & \multicolumn{2}{c}{COCO-30K} \\ 

    \cmidrule(lr){2-9}
    
    & Acc$_{t}$ $\downarrow$ 
    & Acc$_{r}$ $\uparrow$ & KID $\downarrow$ & $H_0\uparrow$
    & CLIP $\uparrow$ & KID $\downarrow$ 
    & CLIP $\uparrow$ & FID $\downarrow$ \\

    \midrule
    FMN \cite{zhang2023forget}
    & 59.98 
    & 56.00 & \:\:0.30 & 0.467
    & 28.23 & \cellcolor{tabfirst}0.01
    & 30.94 & \cellcolor{tabsecond}12.53 \\
    
    ESD-x \cite{gandikota2023erasing}
    &\:\:7.30
    & 10.39 & \:\:2.66 & 0.187
    & 26.65 & 1.20 
    & 29.55 & 14.40 \\

    ESD-u \cite{gandikota2023erasing}
    & 21.20 
    & 28.16 & \:\:8.99 & 0.415
    & 25.39 & 2.38 
    & 28.55 & 15.98 \\

    UCE \cite{gandikota2024unified} 
    &\cellcolor{tabfirst}\:\:0.09 
    & \:\:1.43 & 24.76 & 0.028
    & 20.41 & 5.59 
    & 20.13 & 97.09 \\

    SAFREE \cite{yoonsafree}
    & 86.72 
    & \cellcolor{tabfirst}90.24 & \cellcolor{tabfirst}\:\:0.01 & 0.232
    & \cellcolor{tabthird}28.94 & \cellcolor{tabfirst}0.01 
    & \cellcolor{tabfirst}31.32 & 14.01 \\
    
    SPEED \cite{li2025speed}
    & \:\:3.75 
    & 73.82 & \:\:0.67 & 0.836
    & 28.84 & \cellcolor{tabsecond}0.07 
    & 30.89 & 13.98\\

    MACE \cite{lu2024mace} 
    &\:\:3.29
    & 84.64 &\:\:0.23 & \cellcolor{tabthird}0.903
    & 27.25 & \cellcolor{tabthird}0.47 
    & 30.38 & \cellcolor{tabfirst}12.40  \\
    
    CPE \cite{lee2025concept}
    &\cellcolor{tabthird} \:\:0.37
    & \cellcolor{tabthird}88.26 & \cellcolor{tabsecond}\:\:0.08 & \cellcolor{tabsecond}0.936
    & \cellcolor{tabsecond}29.01 & \cellcolor{tabfirst}0.01 & \cellcolor{tabsecond}31.29 & 14.13 \\

    \textbf{Ours}
    &\cellcolor{tabsecond}\:\:0.24
    & \cellcolor{tabsecond}89.37 & \cellcolor{tabthird}\:\:0.14 & \cellcolor{tabfirst}0.943
    & \cellcolor{tabfirst}29.14 & \cellcolor{tabfirst}0.01 & \cellcolor{tabthird}31.17 & \cellcolor{tabthird}13.61  \\

    \midrule
    SD v1.4 \cite{rombach2022stable1.4}
    & 91.35 & 90.86 & - & - & 28.96 & - & 31.34 & 14.04  \\
    
    \bottomrule
    \end{tabular}
    }
\end{table}

\subsection{Erasure at smaller scale}\label{sec:smaller}
\noindent\textbf{Dataset and evaluation setup.}\,\,
We selected 50 target celebrities using the list from MACE \cite{lu2024mace}. Following CPE \cite{lee2025concept}, we defined two remaining-concept domains: 100 additional celebrities and 100 artistic styles. We also included COCO-30K to evaluate general T2I synthesis capability.
We measure GCD~\cite{hasty_celeb_2024} top-1 detection on celebrity images: lower accuracy indicates stronger erasure of target celebrities, while higher accuracy reflects better preservation of remaining ones. For remaining concepts, we report CLIP Score. Visual quality is measured by Fréchet Inception Distance (FID)~\cite{heusel2017gans} on COCO-30K \cite{hessel2021clipscore} and by KID \cite{sutherland2018demystifying}.

\noindent\textbf{Results.}\,\,
\cref{tab:small_scale_quanti} reports quantitative results in the small-scale setting. In this condition, ETC shows best performance to previous concept erasure methods across erasure and preservation metrics seen in $H_0$ score, as well as image quality indicators such as kernel inception distance (KID).

\begin{figure*}
    \centering
    \includegraphics[width=\linewidth]{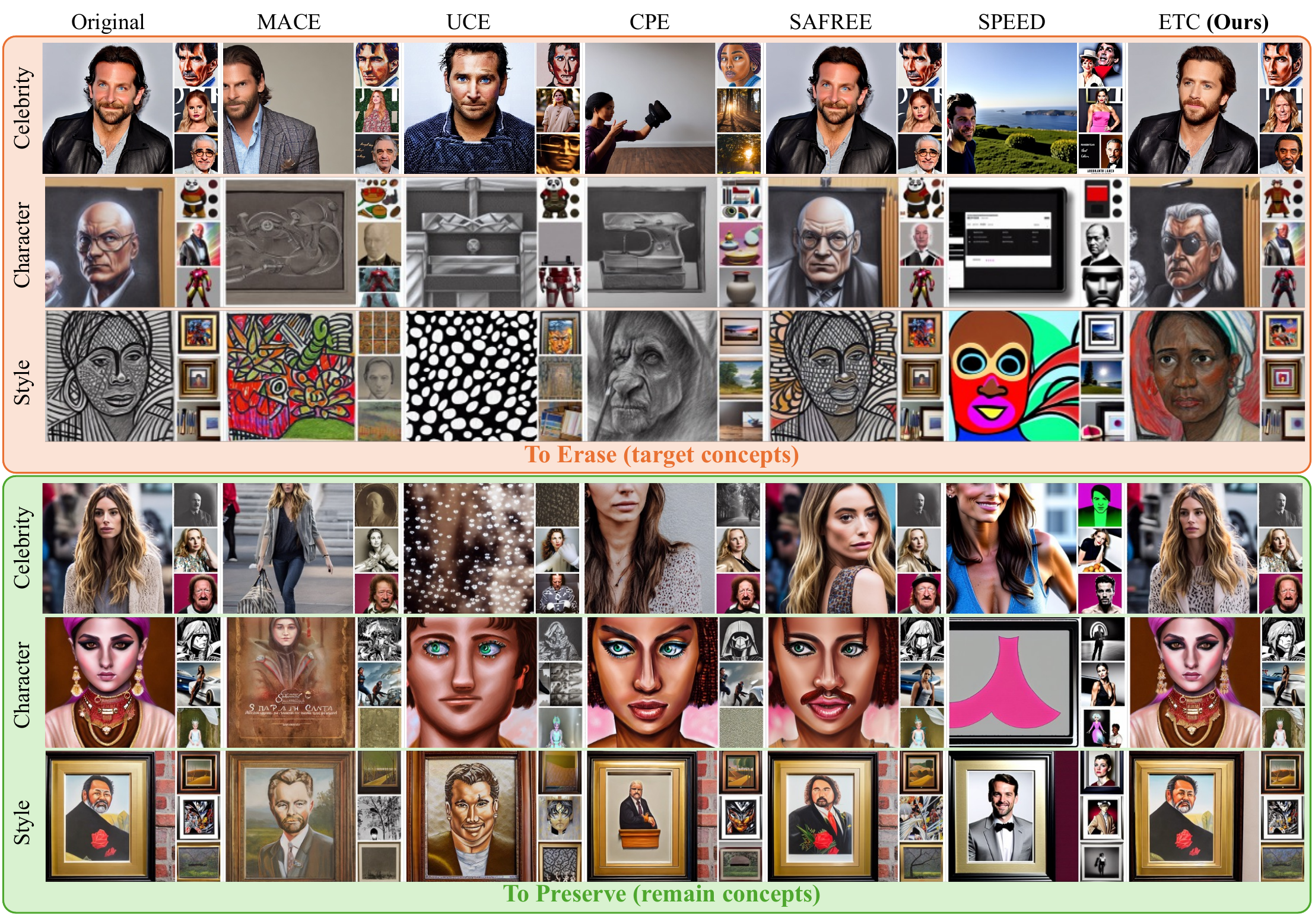}
    \vspace{-2em}
    \caption{\textbf{Qualitative results on erasing 2,072 concepts from SDv1.4.} Among baseline methods (MACE, UCE, CPE, SPEED, and SAFREE), most remove the target concept but often degrade image fidelity, and SAFREE struggles to erase concepts at the large scale. For the preservation of remaining concepts, baseline methods typically alter the original composition or distort remaining concepts. ETC achieves precise removal of the target concept while preserving the original composition and remaining concepts.}
    \label{fig:sd14_quali}
    \vspace{-1em}
\end{figure*}

\subsection{Additional experiments}
In addition to these experiments, we conducted further evaluations on computational cost, explicit content~\cite{schramowski2023safe} and robustness to adversarial attacks~\cite{tsai2024ring, zhang2024defensive}. Results for these experiments are provided in the Sec. \ref{supp:efficiency}, \ref{supp:additional_quanti}, and \ref{supp:additional_quali}.

\subsection{Ablation studies}
We conducted ablation studies on the key components in the ETC framework. All experiments were performed under the same settings described in \cref{sec:smaller}. Further ablation results on the tMM modeling, construction of mapping and anchor embeddings, MoEraser, and NIR are provided in Sec. \ref{supp:method_and_ablations}.

\noindent\textbf{Effect of distribution modeling and mapping.}\,\, To evaluate the effectiveness of tMM modeling and AOT mapping, we compared three distribution modeling methods—direct word-to-word mapping, Gaussian Mixture Model (GMM), and tMM—and two mapping strategies: the Surrogate approach (\textit{e.g.}, \textit{``a person''}, \textit{``an object''}, \textit{``a photograph''}) and our proposed AOT. As shown in \cref{tab:abl_dist_map}, direct mapping degrades at preservation, GMM demonstrates relatively weak target removal, and Surrogate mapping removes targets effectively but fails to retain remaining concepts.

\begin{table}[t]
    \caption{\textbf{Ablation on distribution modeling and AOT mapping.} Direct mapping effectively removes target concepts but fails to preserve remaining ones. GMM performs poorly in both erasure and preservation, suggesting its modeling is ill-suited for this task. While the surrogate mapping achieves strong erasure but weak preservation, combining tMM with AOT yields the best overall.}
    \label{tab:abl_dist_map}
    \vspace{-0.5em}
    \centering
    \setlength{\tabcolsep}{3pt}
    \renewcommand{\arraystretch}{0.9}
    \resizebox{\columnwidth}{!}{
    \begin{tabular}{cc|c|cc|cc|cc}
    \toprule
    \multicolumn{2}{c|}{\multirow{2}{*}{Method}}
    & \multicolumn{1}{c|}{Target} 
    & \multicolumn{6}{c}{Remaining} \\
    
    \cmidrule(lr){3-9}

    && \multicolumn{1}{c|}{50 Celebs}
    & \multicolumn{2}{c|}{100 Celebs}
    & \multicolumn{2}{c|}{100 Styles}
    & \multicolumn{2}{c}{COCO-1K} \\ 

    \cmidrule(lr){1-9}
    
    Modeling & Mapping   
    & Acc$_{t}$ $\downarrow$ 
    & Acc$_{r}$ $\uparrow$ & KID $\downarrow$ 
    & CLIP $\uparrow$ & KID $\downarrow$ 
    & CLIP $\uparrow$ & KID $\downarrow$ \\

    \midrule

    Direct & Surrogate
    & 0.64 & 85.84 & 0.18 & 28.53 & 0.02 & 30.61 & 0.08  \\
    GMM & Surrogate
    & 6.72 & 86.65 & 0.17 & 28.57 & 0.01 & 30.74 & 0.07  \\
    GMM & AOT
    & 8.96 & 89.49 & 0.12 & 29.05 & 0.01 & 30.95 & 0.06  \\
    tMM & Surrogate
    & 0.17 & 86.48 & 0.19 & 28.62 & 0.02 & 30.42 & 0.10  \\
    \textbf{tMM} & \textbf{AOT}
    & 0.24 & 89.37 & 0.14 & 29.14 & 0.01 & 30.87 & 0.06  \\
    
    \midrule
    \multicolumn{2}{c|}{SD v1.4 \cite{rombach2022stable1.4}}
    & 91.35 & 90.86 &  - & 28.96 & - & 31.01 & -  \\
    
    \bottomrule
    \end{tabular}
    }
    \vspace{-1em}
\end{table}

\noindent\textbf{Effect of anchor concept-free preservation.}\,\,
We evaluate anchor-free modeling of the concept distribution across five settings: using the mapping concept as the anchor, mapping to a similar concept, Gaussian noise as the anchor, using only $f_\text{anc}$ sampled from $P^\text{(low)}$, and using both $f_\text{anc}$ and noise. As shown in \cref{tab:abl_anc}, anchor concept-based settings achieve competitive erasure and preservation. In the anchor concept-free regime, using only $f_\text{anc}$ weakens erasure, whereas combining $f_\text{anc}$ with noise yields best on concept erasure and preservation.

\begin{table}[t]
    \caption{\textbf{Ablation study on the type of anchor samples.} We conducted ablation studies on several variants used as anchors and confirm that using embeddings sampled from the distribution boundary or Gaussian noise achieves performance comparable to or better than that obtained with anchor concepts.}
    \label{tab:abl_anc}
    \vspace{-0.5em}
    \centering
    \setlength{\tabcolsep}{3pt}
    \renewcommand{\arraystretch}{0.9}
    \resizebox{\columnwidth}{!}{
    \begin{tabular}{cc|c|cc|cc|cc}
    \toprule
    \multicolumn{2}{c|}{\multirow{2}{*}{Anchor Method}}
    & \multicolumn{1}{c|}{Target} 
    & \multicolumn{6}{c}{Remaining} \\
    
    \cmidrule(lr){3-9}

    && \multicolumn{1}{c|}{50 Celebs}
    & \multicolumn{2}{c|}{100 Celebs}
    & \multicolumn{2}{c|}{100 Styles}
    & \multicolumn{2}{c}{COCO-1K} \\ 

    \cmidrule(lr){1-9}
    
    Type & Concept-free   
    & Acc$_{t}$ $\downarrow$ 
    & Acc$_{r}$ $\uparrow$ & KID $\downarrow$ 
    & CLIP $\uparrow$ & KID $\downarrow$ 
    & CLIP $\uparrow$ & KID $\downarrow$ \\

    \midrule
    Mapping & \redxmark
    & 0.24 & 88.98 & 0.16 & 28.64 & 0.02 & 30.53 & 0.13 \\
    Similar & \redxmark
    & 0.64 & 89.52 & 0.13 & 28.35 & 0.05 & 29.86 & 0.49 \\
    Noise & \greencheck
    & 0.24 & 88.08 & 0.21 & 28.49 & 0.03 & 30.41 & 0.17 \\
    $f_{\text{anc}}$ & \greencheck
    & 1.12 & 88.53 & 0.18 & 28.26 & 0.12 & 29.97 & 0.45 \\
    \textbf{$f_{\text{anc}}$ + noise} & \greencheck
    & 0.24 & 89.37 & 0.14 & 29.14 & 0.01 & 30.87 & 0.06 \\
    
    \midrule
    \multicolumn{2}{c|}{SD v1.4 \cite{rombach2022stable1.4}}
    & 91.35 & 90.86 &  - & 28.96 & - & 31.01 & -  \\
    
    \bottomrule
    \end{tabular}
    }
    \vspace{-1em}
\end{table}

\noindent\textbf{Effect of NIR and noise structure.}\,\,
To assess how noise structure affects fidelity degradation and restoration in the Noise Injection–Restore (NIR) process, we compared three noise types: full-rank, low-rank, and the proposed structured noise. All were normalized to the same Frobenius norm for comparable energy, and performance was measured before and after restoration. As shown in \cref{tab:abl_noise}, all noise types caused similar degradation of the target concept, but structured noise better preserved other concepts, demonstrating more stable restoration in NIR.
\begin{table}[t]
    \caption{\textbf{Ablation on noise structure in NIR.} We compare full-rank, low-rank, and structured (Ours) noise within the NIR process. While all noise types cause similar fidelity degradation to target concepts without MoEraser, structured noise yields superior preservation of remaining concepts after restoration.}
    \label{tab:abl_noise}
    \vspace{-0.5em}
    \centering
    \setlength{\tabcolsep}{2pt}
    \renewcommand{\arraystretch}{0.9}
    \resizebox{\columnwidth}{!}{
    \begin{tabular}{cc|c|cc|cc|cc}
    \toprule
    \multicolumn{2}{c|}{\multirow{2}{*}{Method}}
    


    & \multicolumn{1}{c|}{Target} 
    & \multicolumn{6}{c}{Remaining} \\
    
    \cmidrule(lr){3-9}

    && \multicolumn{1}{c|}{50 Celebs}
    & \multicolumn{2}{c|}{100 Celebs}
    & \multicolumn{2}{c|}{100 Styles}
    & \multicolumn{2}{c}{COCO-1K} \\ 

    \cmidrule(lr){1-9}
    Noise Struc. & Restore 
    & Acc$_{t}$ $\downarrow$ 
    & Acc$_{r}$ $\uparrow$ & KID $\downarrow$ 
    & CLIP $\uparrow$ & KID $\downarrow$ 
    & CLIP $\uparrow$ & KID $\downarrow$ \\

    \midrule

    Full-rank & \redxmark
    & 0.08 & 3.72 & 10.72 & 18.21 & 9.35 & 21.12 & 6.32  \\
    Full-rank & \greencheck
    & 0.16 & 79.28 & \:\:0.27 & 27.05 & 0.76 & 29.42 & 0.24  \\
    Low-rank & \redxmark
    & 0.16 & 4.57 & \:\:9.43 & 19.03 & 7.53 & 20.57 & 7.01  \\
	Low-rank & \greencheck
    & 0.16 & 85.84 & \:\:0.21 & 27.41 & 0.42 & 30.16 & 0.17  \\
    Ours & \redxmark
    & 0.08 & 7.16 & \:\:9.72 & 22.46 & 4.38 & 23.36 & 2.69  \\
    Ours & \greencheck
    & 0.24 & 89.37 & \:\:0.14 & 29.14 & 0.01 & 30.87 & 0.06  \\

    \midrule
    SD v1.4 \cite{rombach2022stable1.4}
    & - & 91.35 & 90.86 &  - & 28.96 & - & 31.01 & -  \\
    
    \bottomrule
    \end{tabular}
    }
    \vspace{-1em}
\end{table}


\begin{figure}[!t]
    \centering
    \includegraphics[width=\linewidth]{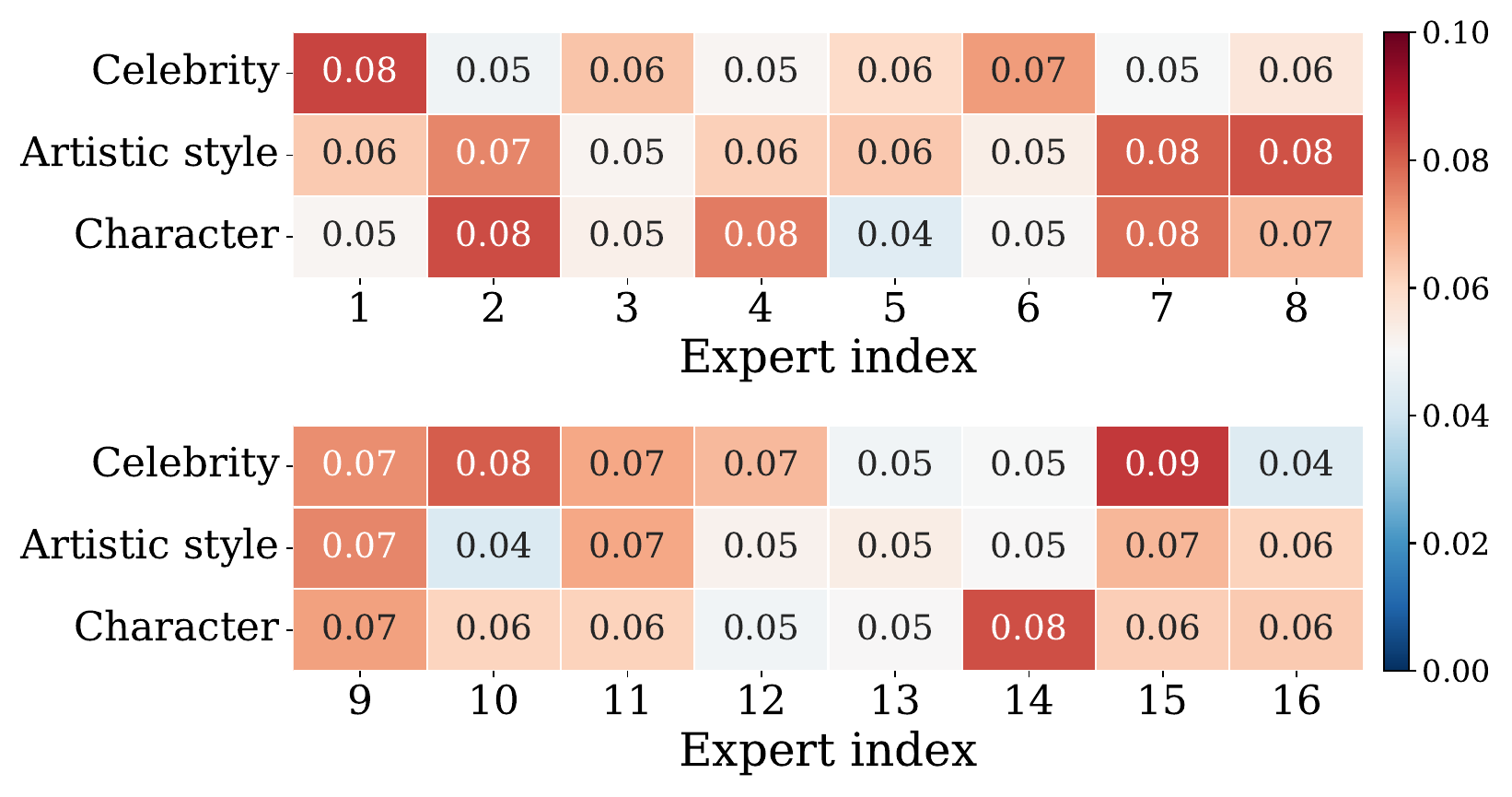}
    \vspace{-2em}
    \caption{\textbf{Load heatmap of experts.} We visualize the frequency ratio of selection of each expert for three domains where each column represents an expert, and each row corresponds to a domain. The relatively uniform load distribution across experts suggests that the router network effectively balances expert utilization.}
    \label{fig:dis_expert}
    \vspace{-1em}
\end{figure}

\section{Discussion}
\noindent\textbf{Experts selection analysis.}\,\,
We use a MoE architecture for erasure across heterogeneous concept domains. A risk is sparse expert usage per concept, which could reduce robustness or enable circumvention via specific expert paths. We therefore analyze expert activations by domain—celebrity, artistic style, and character—in \cref{fig:dis_expert}. Some experts are preferred within each domain, but none dominates, indicating the router distributes activations relatively evenly.

\section{Conclusion}
We introduced a scalable framework, Erasing Thousands of Concepts (ETC) for large-scale concept erasure in text-to-image (T2I) diffusion models. ETC models low-rank concept distributions using a Student's t-distribution Mixture Model (tMM) and aligns them through affine optimal transport (AOT). A MoEraser module removes target embeddings while preserving remaining concepts in an anchor concept-free manner, and a noise injection–restore (NIR) enhances robustness against module removal. Our framework achieves superior scalability and precision in large-scale erasure scenarios involving over 2,000 heterogeneous concepts across multiple domains. Moreover, it effectively generalizes to advanced T2I models such as Stable Diffusion v3.5-Large. In relatively smaller-scale settings, the method attains performance comparable to or exceeding existing approaches, demonstrating its versatility.

\section*{Acknowledgements}
This work was supported in part by Institute of Information \& communications Technology Planning \& Evaluation (IITP) grants funded by the Korea government(MSIT) [No.RS-2021-II211343, Artificial Intelligence Graduate School Program (Seoul National University) / No.RS-2025-02314125, Effective Human-Machine Teaming With Multimodal Hazy Oracle Models], the National Research Foundation of Korea(NRF) grant funded by the Korea government(MSIT) (No. RS-2025-02263628) and the BK21 FOUR program of the Education and Research Program for Future ICT Pioneers, Seoul National University.

{
    \small
    \bibliographystyle{ieeenat_fullname}
    \bibliography{main}
}

 
 
\onecolumn
\clearpage
\twocolumn[{%
\begin{center}
    {\Large \textbf{Erasing Thousands of Concepts: Towards Scalable and Practical\\
    Concept Erasure for Text-to-Image Diffusion Model}} \\[1em]
    {\Large \emph{Supplementary Materials}} \\[1em]
\end{center}
}]
 
\setcounter{section}{0}
\setcounter{figure}{0}
\setcounter{table}{0}
\setcounter{equation}{0}
\setcounter{proposition}{0}
\setcounter{theorem}{0}
\setcounter{corollary}{0}
 
\renewcommand{\thesection}{\Alph{section}}
\renewcommand{\thefigure}{\thesection.\arabic{figure}}
\renewcommand{\thetable}{\thesection.\arabic{table}}
\renewcommand{\theequation}{\thesection.\arabic{equation}}
\counterwithin{table}{section}
\counterwithin{figure}{section}

\section{Details on Method and Ablation Studies} \label{supp:method_and_ablations}

\subsection{Modeling concept embeddings via tMMs}
\paragraph{tMM modeling templates.}
To model the concept embedding distribution, we insert the word corresponding to each concept into a variety of prompts and extract the associated concept embeddings from the text encoder. The templates are automatically generated using ChatGPT-5, yielding a total of 2,582 templates. Since the number of tokens per concept word varies and the CLIP text encoder we use for all embedding extraction employs a causal mask, the information for each concept is concentrated in the last token. We therefore use only the last token of the concept word for concept modeling.
\paragraph{Parameters in tMMs.}
Our modeling of concepts as tMMs requires several parameters to define the distribution: the degrees of freedom (DOF), the number of modes, and the number or rank of principal components for efficient modeling and optimal transport to mapping embeddings. In a Student’s t-distribution, a smaller degrees of freedom value yields a more heavy-tailed distribution. A larger number of modes allows the distribution to capture a wider variety of visual or semantic variants for each concept. The number of principal components controls how simply the concept can be represented in a linear subspace. In this work, we use the same configuration for all concepts, setting the degrees of freedom to 2, the number of modes to 4, and the number of principal components to 32.

\cref{tab:abl_dof} and \cref{tab:abl_num_pca} report the results obtained under different modeling parameters. For the degrees of freedom (DOF), as the DOF increases, the Student’s t-distribution becomes mathematically closer to a Gaussian distribution, and we observe a corresponding performance degradation, consistent with the Gaussian-distribution ablation study presented in the main paper. For the rank, the performance remains largely stable across different values, indicating robustness with respect to the choice of rank. These results collectively support the validity of our hyperparameter.

\begin{table}[h]
    \caption{Effect of the degrees of freedom for modeling tMMs.}
    \label{tab:abl_dof}
    \centering
    \setlength{\tabcolsep}{3pt}
    \renewcommand{\arraystretch}{0.9}
    \resizebox{\columnwidth}{!}{
    \begin{tabular}{c|c|cc|cc|cc}
    \toprule
    \multirow{3}{*}{DOF} 
    & \multicolumn{1}{c|}{Target} 
    & \multicolumn{6}{c}{Remaining} \\
    
    \cmidrule(lr){2-8}

    & \multicolumn{1}{c|}{50 Celebs}
    & \multicolumn{2}{c|}{100 Celebs}
    & \multicolumn{2}{c|}{100 Styles}
    & \multicolumn{2}{c}{COCO-1K} \\

    \cmidrule(lr){2-8}
    
    & Acc$_{t}$ $\downarrow$ 
    & Acc$_{r}$ $\uparrow$ & KID $\downarrow$ 
    & CLIP $\uparrow$ & KID $\downarrow$ 
    & CLIP $\uparrow$ & KID $\downarrow$ \\

    \midrule

    1 
    & 0.24 & 89.02 & 0.15 & 29.09 & 0.01 & 30.89 & 0.06 \\
    \textbf{2} 
    & 0.24 & 89.37 & 0.14 & 29.14 & 0.01 & 30.87 & 0.06 \\
    4
    & 0.64 & 89.56 & 0.11 & 29.04 & 0.01 & 30.94 & 0.06 \\
    8 
    & 0.72 & 89.44 & 0.12 & 29.01 & 0.01 & 30.93 & 0.06 \\

    \midrule
    SD v1.4 \cite{rombach2022stable1.4}
    & 91.35 & 90.86 &  - & 28.96 & - & 31.02 & -  \\
    
    \bottomrule
    \end{tabular}
    }
\end{table}

    


    


    
    

\begin{table}[h]
    \caption{Effect of the number of principal components for modeling tMMs}
    \label{tab:abl_num_pca}
    \centering
    \setlength{\tabcolsep}{3pt}
    \renewcommand{\arraystretch}{0.9}
    \resizebox{\columnwidth}{!}{
    \begin{tabular}{c|c|cc|cc|cc}
    \toprule
    \multirow{3}{*}{Rank} 
    & \multicolumn{1}{c|}{Target} 
    & \multicolumn{6}{c}{Remaining} \\
    
    \cmidrule(lr){2-8}

    & \multicolumn{1}{c|}{50 Celebs}
    & \multicolumn{2}{c|}{100 Celebs}
    & \multicolumn{2}{c|}{100 Styles}
    & \multicolumn{2}{c}{COCO-1K} \\

    \cmidrule(lr){2-8}
    
    & Acc$_{t}$ $\downarrow$ 
    & Acc$_{r}$ $\uparrow$ & KID $\downarrow$ 
    & CLIP $\uparrow$ & KID $\downarrow$ 
    & CLIP $\uparrow$ & KID $\downarrow$ \\

    \midrule

    16   
    & 0.24 & 89.40 & 0.13 & 29.04 & 0.01 & 30.84 & 0.07 \\
    \textbf{32} 
    & 0.24 & 89.37 & 0.14 & 29.14 & 0.01 & 30.87 & 0.06 \\
    64   
    & 0.40 & 89.24 & 0.14 & 29.09 & 0.01 & 30.93 & 0.04 \\
    128  
    & 0.32 & 89.28 & 0.14 & 29.11 & 0.01 & 30.89 & 0.05 \\
        
    \midrule
    SD v1.4 \cite{rombach2022stable1.4}
    & 91.35 & 90.86 &  - & 28.96 & - & 31.02 & -  \\
    
    \bottomrule
    \end{tabular}
    }
\end{table}

\paragraph{Selection of mapping concepts.}
A sample from the training dataset consists of three components: target, anchor, and mapping embeddings. Target embeddings for each target concept are sampled from the tMM corresponding to that concept by drawing, for each mode, embeddings from the high-probability region, defined as the confidence interval determined by a confidence level $\tau_1$ for that mode; within this region, we uniformly sample embeddings. In contrast, anchor embeddings corresponding to a given target concept are randomly sampled from outside this confidence-interval region of the same tMM. Sampling mapping embeddings requires a more sophisticated selection rule, including deciding which mapping concepts to choose, how many concepts to select in order to construct a merged tMM for mapping embeddings, and how to adjust the norms of the sampled mapping concept embeddings.

For the selection of mapping concepts, we consider two main factors: they should be semantically similar to the target concepts while exhibiting visually distinct features, and they should enable the generation of non-identifiable visual features while maintaining high image fidelity. For the first factor, we draw inspiration from the observation in prior work \cite{bui2025fantastic} that choosing mapping concepts that are close to the target concepts yet differ in specific semantics is effective for pinpoint erasure of target concepts without degrading remaining concepts that share highly similar semantics. To satisfy the second factor, we select multiple mapping concepts but control the variance of their merged tMM so that the target embeddings are not mapped to regions associated with low-fidelity image generation.

\begin{algorithm}[t]
\caption{Selection rule of mapping concepts}
\begin{algorithmic}[1]
\State { \textbf{Input:} 
a target concept $c_{\text{tar}}$,
mapping concept pool $\{c_{\text{map}}^{m}\}_{m=1}^{M}$,
stacked embeddings of $c_{\text{tar}}$ as $X_{c_{\text{tar}}}$, stacked embeddings of $\{c_{\text{map}}\}_{m=1}^{M}$ as $\{ X_{c_{\text{map}}^{m}}\}_{m=1}^M$, the number of selected mapping concepts $M'$, $\tau_2$ for threshold of similarity, and $\tau_3$ for threshold of variance of merged mapping emddings.}

{\State $\mu_{c_{\text{tar}}} = \frac{1}{N} \sum_{n=1}^{N}[X_{c_{\text{tar}}}]_n$ 
}
{\State $\mu_{c_{\text{map}}}^{m} = \frac{1}{N} \sum_{n=1}^{N}[X_{c_{\text{map}}}^m]_n$ for $m=1,2, \cdots, M$ 
}
{\State $S = [\textbf{sim} (\mu_{c_{\text{tar}}}, \mu_{c_{\text{map}}}^{1}), \cdots, \textbf{sim} (\mu_{c_{\text{tar}}}, \mu_{c_{\text{map}}}^{M})]$ 
}
{\State $S_{\downarrow} = \textbf{sort}(S, \text{descending}=\text{True})$
}
{\State $I_{\downarrow} = \textbf{argsort}(S, \text{descending}=\text{True})$
}
{\For{$i \leftarrow 1,2, \cdots, M$}
    \If{$[S_{\downarrow}]_{i} \le \tau_2$}
        \State $j = [I_{\downarrow}]_{i}$
    
        \State $S = [\textbf{sim} (\mu_{c_{\text{map}}}^{j}, \mu_{c_{\text{map}}}^{1}), \cdots, \textbf{sim} (\mu_{c_{\text{map}}}^{j}, \mu_{c_{\text{map}}}^{M})]$
    
        \State $S'_{\downarrow} = \textbf{sort}(S', \text{descending}=\text{True})$
        
        \State $I'_{\downarrow} = \textbf{argsort}(S', \text{descending}=\text{True})$
    
        \State $I_{\text{map}} = [I'_{\downarrow}]_{:M'}$
    
        \State $X_{\text{map}} = [X_{c_{\text{map}}}^{[I_{\text{map}}]_{1}}, \cdots, X_{c_{\text{map}}}^{[I_{\text{map}}]_{M'}}]$
    
        \If{$\text{Tr}(\text{Cov}[[X_{\text{map}}]_n]) \le \tau_3$}
            \State $C_{\text{map}} = [c_{\text{map}}^{I_{\text{map},1}},
            \cdots,
            c_{\text{map}}^{I_{\text{map},M'}}]$
            \State \text{break}
        \EndIf
    \EndIf
\EndFor
}
\\
{\textbf{Output:} Selected mapping concepts $C_{\text{map}}$.}

\end{algorithmic}
\label{alg:select_map}
\end{algorithm}

\cref{alg:select_map} specifies the selection rule for mapping concepts for each target concept under the above criteria. We first use cosine similarity $\text{sim}$ and a threshold $\tau_2$ to identify one mapping concept that is deemed sufficiently similar to the target concept. We then select several additional mapping concepts that are closest to this mapping concept and evaluate the variance of the union of their embeddings. If this variance is smaller than a threshold $\tau_3$, meaning that their variation is limited and thus unlikely to span regions associated with low-fidelity image generation, we select this set of concepts as the final mapping concepts for constructing the merged tMM. In the experiments, we fixed $\tau_2 = 0.3$ and $\tau_3 = 0.5$ across all concepts.

\begin{table}[t]
    \caption{Effect of the number of selected mapping concepts.}
    \label{tab:abl_mapping_num_concepts}
    \centering
    \setlength{\tabcolsep}{3pt}
    \renewcommand{\arraystretch}{0.9}
    \resizebox{\columnwidth}{!}{
    \begin{tabular}{c|c|cc|cc|cc}
    \toprule
    \multirow{3}{*}{\# Mappings} 
    & \multicolumn{1}{c|}{Target} 
    & \multicolumn{6}{c}{Remaining} \\
    
    \cmidrule(lr){2-8}

    & \multicolumn{1}{c|}{50 Celebs}
    & \multicolumn{2}{c|}{100 Celebs}
    & \multicolumn{2}{c|}{100 Styles}
    & \multicolumn{2}{c}{COCO-1K} \\

    \cmidrule(lr){2-8}
    
    & Acc$_{t}$ $\downarrow$ 
    & Acc$_{r}$ $\uparrow$ & KID $\downarrow$ 
    & CLIP $\uparrow$ & KID $\downarrow$ 
    & CLIP $\uparrow$ & KID $\downarrow$ \\

    \midrule

    1   
    & 0.24 & 89.64 & 0.11 & 28.99 & 0.01 & 30.94 & 0.05 \\
    \textbf{3}   
    & 0.24 & 89.37 & 0.14 & 29.14 & 0.01 & 30.87 & 0.06 \\
    5   
    & 2.08 & 88.96 & 0.16 & 29.07 & 0.01 & 30.83 & 0.07 \\
    10  
    & 3.92 & 88.72 & 0.18 & 29.04 & 0.01 & 30.78 & 0.07 \\
        
    \midrule
    SD v1.4 \cite{rombach2022stable1.4}
    & 91.35 & 90.86 &  - & 28.96 & - & 31.02 & -  \\
    
    \bottomrule
    \end{tabular}
    }
\end{table}

In this setting, we forcibly rescale the norms of the mapped embeddings to match the norms of the corresponding target embeddings. Since CLIP is trained with a cosine-similarity-based objective, it approximately follows a hyper-sphere geometry, and AOT maps embeddings into a simplex residing on this hypersphere; under normal mapping, the norms tend to decrease, causing the mapped embeddings to become under-represented. This phenomenon has already been
well explored in the prior works \cite{seo2025geometricalpropertiestexttoken, hu2024token}.

\paragraph{Ablation studies on number of mapping concepts.}
The ETC framework constructs a mapping concept distribution using a triplet of concepts and then maps the target distribution to this mapping distribution via AOT. We perform an ablation study that varies the number of mapping concepts from this default triplet to a single concept, five concepts, and ten concepts, as reported in \cref{tab:abl_mapping_num_concepts}. We observe that using fewer mapping concepts tends to improve both erasure and preservation performance, whereas increasing the number of mapping concepts exhibits the opposite trend. To ultimately guide each target concept toward an anonymous identity while balancing these effects, we select three mapping concepts as an appropriate choice and adopt the concept triplet configuration in our final ETC framework.

\paragraph{Ablation studies on the confidence level.}
We define the boundary between target and anchor concept embeddings using a confidence level, that is, the point at which the cumulative probability of the distribution reaches a specified value; embeddings inside this boundary are treated as target concept embeddings, and those outside as anchor concept embeddings. We set this confidence level to 0.9 and conduct an ablation study over different values, as reported in \cref{tab:abl_confidence_level}. As the confidence level decreases, the range of the distribution to be erased shrinks, resulting in reduced erasure performance, while the region assigned to anchor concept embeddings expands, improving preservation performance. Conversely, increasing the confidence level enlarges the erased region, enhancing erasure performance but degrading preservation. Considering this trade-off, we select 0.9 as a sweet-spot value that allows ETC to achieve high performance in both erasure and preservation.

\begin{table}[t]
    \caption{Effect of confidence level for sampling target and anchor embeddings.}
    \label{tab:abl_confidence_level}
    \centering
    \setlength{\tabcolsep}{3pt}
    \renewcommand{\arraystretch}{0.9}
    \resizebox{\columnwidth}{!}{
    \begin{tabular}{c|c|cc|cc|cc}
    \toprule
    \multirow{3}{*}{$\tau_1$} 
    & \multicolumn{1}{c|}{Target} 
    & \multicolumn{6}{c}{Remaining} \\
    
    \cmidrule(lr){2-8}

    & \multicolumn{1}{c|}{50 Celebs}
    & \multicolumn{2}{c|}{100 Celebs}
    & \multicolumn{2}{c|}{100 Styles}
    & \multicolumn{2}{c}{COCO-1K} \\ 
    
    \cmidrule(lr){2-8}
    
    & Acc$_{t}$ $\downarrow$ 
    & Acc$_{r}$ $\uparrow$ & KID $\downarrow$ 
    & CLIP $\uparrow$ & KID $\downarrow$ 
    & CLIP $\uparrow$ & KID $\downarrow$ \\

    \midrule

    0.7   
    & 2.32 & 89.68 & 0.11 & 29.03 & 0.01 & 30.98 & 0.05 \\
    0.8   
    & 0.72 & 89.96 & 0.09 & 28.98 & 0.01 & 30.92 & 0.04 \\
    \textbf{0.9}   
    & 0.24 & 89.37 & 0.14 & 29.14 & 0.01 & 30.87 & 0.06 \\
    0.95  
    & 0.24 & 88.08 & 0.18 & 28.85 & 0.05 & 30.89 & 0.06 \\
    0.99  
    & 0.16 & 85.12 & 0.22 & 28.27 & 0.09 & 30.62 & 0.09 \\
    0.999 
    & 0.16 & 82.52 & 0.24 & 27.91 & 0.11 & 30.54 & 0.12 \\
        
    \midrule
    SD v1.4 \cite{rombach2022stable1.4}
    & 91.35 & 90.86 &  - & 28.96 & - & 31.02 & -  \\
    
    \bottomrule
    \end{tabular}
    }
\end{table}

\subsection{MoEraser}

\paragraph{Effect of MoE architecture.}
To evaluate the MoE architecture, we compared it with models of equal parameter counts: a linear layer, an FFN with ReLU, and an FFN with GLU activation. As shown in \cref{tab:abl_MoE_archi}, the linear model performed the worst, the GLU-based FFN outperformed the ReLU variant, and the MoE achieved the best erasure and preservation results, confirming its effectiveness.

\begin{table}[t]
    \caption{Ablation study on the effect of the MoE architecture. 
    We compare our MoEraser with comparable parameter counts: a linear layer, an FFN with ReLU, and a FFN with GLU.
    }
    \label{tab:abl_MoE_archi}
    \centering
    \setlength{\tabcolsep}{3pt}
    \renewcommand{\arraystretch}{0.9}
    \resizebox{\columnwidth}{!}{
    \begin{tabular}{c|c|cc|cc|cc}
    \toprule
    \multirow{3}{*}{Method} 
    & \multicolumn{1}{c|}{Target} 
    & \multicolumn{6}{c}{Remaining} \\
    
    \cmidrule(lr){2-8}

    & \multicolumn{1}{c|}{50 Celebs}
    & \multicolumn{2}{c|}{100 Celebs}
    & \multicolumn{2}{c|}{100 Styles}
    & \multicolumn{2}{c}{COCO-1K} \\ 

    \cmidrule(lr){2-8}
    
    & Acc$_{t}$ $\downarrow$ 
    & Acc$_{r}$ $\uparrow$ & KID $\downarrow$ 
    & CLIP $\uparrow$ & KID $\downarrow$ 
    & CLIP $\uparrow$ & KID $\downarrow$ \\

    \midrule

    Linear
    & 7.89 & 88.12 & 0.19 & 27.95 & 0.25 & 29.68 & 0.19  \\
    FFN-ReLU
    & 5.73 & 89.08 & 0.15 & 28.85 & 0.02 & 30.92 & 0.05  \\    
    FFN-GLU
    & 0.48 & 88.76 & 0.17 & 29.03 & 0.01 & 30.75 & 0.08  \\    
    \textbf{MoEraser}
    & 0.24 & 89.37 & 0.14 & 29.14 & 0.01 & 30.87 & 0.06  \\
    
    \midrule
    SD v1.4 \cite{rombach2022stable1.4}
    & 91.35 & 90.86 &  - & 28.96 & - & 31.01 & -  \\
    
    \bottomrule
    \end{tabular}
    }
\end{table}

\paragraph{Components in MoE architecture.}
Mixture-of-Experts has two key components: the number of experts and the number of selected experts. For the first factor, we conduct a study by increasing the number of experts while keeping the total number of parameters approximately fixed; the results are reported in \cref{tab:abl_num_of_expert}. As the number of experts increases, we observe a trade-off in which the performance of erasing the target concept degrades, whereas the preservation performance for the remaining concepts improves, and we choose 8 experts as a balanced operating point. For the second factor, the Top-$k$ selection of experts, we vary the value of $k$ as summarized in \cref{tab:abl_num_of_selected_expert} and adopt $k=6$, which achieves the best overall performance.

\begin{table}[t]
\caption{Effect of the number of experts}
\label{tab:abl_num_of_expert}
    \centering
    \setlength{\tabcolsep}{3pt}
    \renewcommand{\arraystretch}{0.9}
    \resizebox{\columnwidth}{!}{
    \begin{tabular}{c|c|cc|cc|cc}
    \toprule
    \multirow{3}{*}{\# Experts} 
    & \multicolumn{1}{c|}{Target} 
    & \multicolumn{6}{c}{Remaining} \\
    
    \cmidrule(lr){2-8}

    & \multicolumn{1}{c|}{50 Celebs}
    & \multicolumn{2}{c|}{100 Celebs}
    & \multicolumn{2}{c|}{100 Styles}
    & \multicolumn{2}{c}{COCO-1K} \\ 

    \cmidrule(lr){2-8}
    
    & Acc$_{t}$ $\downarrow$ 
    & Acc$_{r}$ $\uparrow$ & KID $\downarrow$ 
    & CLIP $\uparrow$ & KID $\downarrow$ 
    & CLIP $\uparrow$ & KID $\downarrow$ \\
    
    \midrule
    2  
    & 0.16 & 88.17 & 0.16 & 28.82 & 0.02 & 30.76 & 0.09 \\
    4  
    & 0.32 & 89.12 & 0.14 & 29.09 & 0.01 & 30.89 & 0.06 \\
    \textbf{8}  
    & 0.24 & 89.37 & 0.14 & 29.14 & 0.01 & 30.87 & 0.06 \\
    16 
    & 0.24 & 89.44 & 0.14 & 29.08 & 0.01 & 30.91 & 0.06 \\
    32 
    & 0.32 & 88.96 & 0.15 & 29.07 & 0.01 & 30.83 & 0.06 \\
    
    \bottomrule
    \end{tabular}
    }
\end{table}

\begin{table}[t]
    \caption{Effect of the number of selected experts}
    \label{tab:abl_num_of_selected_expert}
    \centering
    \setlength{\tabcolsep}{3pt}
    \renewcommand{\arraystretch}{0.9}
    \resizebox{\columnwidth}{!}{
    \begin{tabular}{c|c|cc|cc|cc}
    \toprule
    \multirow{3}{*}{\# Top-$k$ Exp.} 
    & \multicolumn{1}{c|}{Target} 
    & \multicolumn{6}{c}{Remaining} \\
    
    \cmidrule(lr){2-8}

    & \multicolumn{1}{c|}{50 Celebs}
    & \multicolumn{2}{c|}{100 Celebs}
    & \multicolumn{2}{c|}{100 Styles}
    & \multicolumn{2}{c}{COCO-1K} \\ 

    \cmidrule(lr){2-8}
    
    & Acc$_{t}$ $\downarrow$ 
    & Acc$_{r}$ $\uparrow$ & KID $\downarrow$ 
    & CLIP $\uparrow$ & KID $\downarrow$ 
    & CLIP $\uparrow$ & KID $\downarrow$ \\

    \midrule
    
    1 
    & 0.48 & 87.92 & 0.18 & 29.04 & 0.01 & 30.91 & 0.05  \\
    2 
    & 0.24 & 89.04 & 0.14 & 29.06 & 0.01 & 30.88 & 0.06 \\
    4 
    & 0.32 & 88.96 & 0.15 & 29.01 & 0.01 & 30.93 & 0.04 \\
    \textbf{6} 
    & 0.24 & 89.37 & 0.14 & 29.14 & 0.01 & 30.87 & 0.06 \\
    8 
    & 0.54 & 88.84 & 0.16 & 28.98 & 0.01 & 30.79 & 0.08 \\
    
    \bottomrule
    \end{tabular}
    }
\end{table}

\paragraph{Effect of AdaLN-Zero.}
We adopt AdaLN-Zero, originally proposed in DiT~\cite{peebles2023scalable}, as part of the MoEraser architecture, which leads to a slight improvement in preserving remaining concepts. We conduct an ablation study to assess its impact, and the results are reported in \cref{tab:abl_AdaLN}. The target concept erasure performance is similar with and without AdaLN-Zero, whereas the preservation performance on remaining concepts increases slightly when AdaLN-Zero is used. Since this component introduces only a negligible increase in the number of parameters, we include it in our final architecture.

\begin{table}[t]
\caption{Effect of applying AdaLN-Zero}
\label{tab:abl_AdaLN}
    \centering
    \setlength{\tabcolsep}{3pt}
    \renewcommand{\arraystretch}{0.9}
    \resizebox{\columnwidth}{!}{
    \begin{tabular}{c|c|cc|cc|cc}
    \toprule
    \multirow{3}{*}{AdaLN-Zero} 
    & \multicolumn{1}{c|}{Target} 
    & \multicolumn{6}{c}{Remaining} \\
    
    \cmidrule(lr){2-8}

    & \multicolumn{1}{c|}{50 Celebs}
    & \multicolumn{2}{c|}{100 Celebs}
    & \multicolumn{2}{c|}{100 Styles}
    & \multicolumn{2}{c}{COCO-1K} \\ 

    \cmidrule(lr){2-8}
    
    & Acc$_{t}$ $\downarrow$ 
    & Acc$_{r}$ $\uparrow$ & KID $\downarrow$ 
    & CLIP $\uparrow$ & KID $\downarrow$ 
    & CLIP $\uparrow$ & KID $\downarrow$ \\

    \midrule
    \redxmark
    & 0.24 & 89.32 & 0.14 & 29.11 & 0.01 & 30.83 & 0.07  \\
    \greencheck 
    & 0.24 & 89.37 & 0.14 & 29.14 & 0.01 & 30.87 & 0.06 \\
    
    \midrule
    SD v1.4 \cite{rombach2022stable1.4}
    & 91.35 & 90.86 &  - & 28.96 & - & 31.01 & -  \\
    
    \bottomrule
    \end{tabular}
    }
\end{table}

\subsection{Noise Injection-Restore tuning}
\paragraph{Effect of the scale parameter $\alpha_{\text{noise}}$.}
We obtain robustness to model removal through Noise-Injection Restoration tuning. The noise scale $\alpha_{\text{noise}}$ directly influences the degree to which image fidelity degrades and, consequently, the preservation performance for remaining concepts. To examine this behavior, we evaluate erasure and preservation performance across different noise scales in \cref{tab:abl_noise}. Smaller noise scales tend to improve both erasure and preservation, but the model becomes insufficiently perturbed and therefore less resistant to module-removal attacks. Conversely, excessively large values of $\alpha_{\text{noise}}$ lead to perturbations that are not fully recoverable during restoration. Considering this trade-off, we adopt moderate noise levels (\textit{e.g.}, $\alpha_{\text{noise}} = 2.0 \times 10^{-3}$).

\begin{table}[t]
    \caption{Effect of the noise scale $\alpha_\text{noise}$.}
    \label{tab:abl_noise}
    \centering
    \setlength{\tabcolsep}{3pt}
    \renewcommand{\arraystretch}{0.9}
    \resizebox{\columnwidth}{!}{
    \begin{tabular}{c|c|cc|cc|cc}
    \toprule
    \multirow{3}{*}{Scale of noise} 
    & \multicolumn{1}{c|}{Target} 
    & \multicolumn{6}{c}{Remaining} \\
    
    \cmidrule(lr){2-8}

    & \multicolumn{1}{c|}{50 Celebs}
    & \multicolumn{2}{c|}{100 Celebs}
    & \multicolumn{2}{c|}{100 Styles}
    & \multicolumn{2}{c}{COCO-1K} \\ 

    \cmidrule(lr){2-8}
    
    & Acc$_{t}$ $\downarrow$ 
    & Acc$_{r}$ $\uparrow$ & KID $\downarrow$ 
    & CLIP $\uparrow$ & KID $\downarrow$ 
    & CLIP $\uparrow$ & KID $\downarrow$ \\

    \midrule
    $2.0\times10^{-5}$
    & 0.16 & 89.44 & 0.12 & 29.02 & 0.01 & 30.90 & 0.05 \\
    $2.0\times10^{-4}$
    & 0.24 & 89.48 & 0.11 & 29.06 & 0.01 & 30.91 & 0.05 \\
    $\mathbf{2.0\times10^{-3}}$
    & 0.24 & 89.37 & 0.14 & 29.14 & 0.01 & 30.87 & 0.06 \\
    $2.0\times10^{-2}$
    & 0.16 & 85.21 & 0.21 & 28.07 & 0.19 & 29.35 & 0.29 \\
    $2.0\times10^{-1}$
    & 0.0 & 0.04 & 41.7 & 9.28 & 52.7 & 8.95 & 37.2 \\
    \midrule
    SD v1.4 \cite{rombach2022stable1.4}
    & 91.35 & 90.86 &  - & 28.96 & - & 31.01 & -  \\

    \bottomrule
    \end{tabular}
    }
\end{table}

For further illustration, \cref{fig:supp_noise_scale} shows qualitative examples showing how the noise scale of $\alpha_{\text{noise}}$ degrade the generated images and the NIR module subsequently restores them. Extremely small noise levels introduce only mild perturbations that are almost imperceptible. In contrast, larger noise scales produce increasingly severe distortions ranging from subtle texture corruption to complete structural collapse, which demonstrates how the injected noise disrupts the visual representations. We found that moderate noise levels like $\alpha_{\text{noise}} = 2.0 \times 10^{-3}$ are sufficient to corrupt the visual representations while still allowing the restoration module to recover high-fidelity images, thereby enabling reliable erasure of the target concept.

\section{Experimental Setups} \label{supp:exp_setup}

\subsection{Concept erasure at scale}

We conducted a large-scale experiment to simultaneously erase heterogeneous concepts from three domains with SDv1.4: celebrities, artistic styles, and characters. Specifically, we removed 949 celebrities, 693 artistic styles, and 430 characters, resulting in a total of 2,072 erased concepts. For celebrities, we selected targets from 2,306 identities which can be detected by the GIPHY Celebrity Detector (GCD)~\cite{hasty_celeb_2024}.
For artistic styles, we used 1,734 styles provided by UCE~\citep{gandikota2024unified}.
For characters, we constructed a list of 1,016 characters using a large language model and selected a subset as target concepts. For each domain, the mapping and remaining concepts used to evaluate the utility of the model were chosen from their respective concept pools, excluding the target concepts.
The mapping and remaining sets were selected to be mutually exclusive to ensure non-overlapping evaluations. The list of target, mapping, and remaining concepts are listed in Sec.~\ref{sec:concept_list_2072}

To construct the evaluation benchmark, we generated the full prompt pools by combining each target and remaining concept with the 50 prompt templates for each domain listed in Sec.~\ref{sec:prompt_temp}. 
From this pool, we randomly selected 20,143 prompts to form the final prompt set used for generation, ensuring that all combinations of heterogeneous domains and target/remaining concepts were sampled. 
For each prompt, the generation seed was sampled purely at random.

\subsection{Extension to advanced diffusion model}
For concept erasure with SDv3.5-L, we used the same concept pools and prompt pools described above.  In this setting, we erased 250 celebrities, 151 artistic styles, and 114 characters, for a total of 515 target concepts, and evaluated preservation performance using 133 celebrities, 59 artistic styles, and 80 characters designated as remaining concepts. From the prompt pools, we sampled 6,546 image pools to assess both the effectiveness of erasing the target concepts and the fidelity of the generated images.

\subsection{Erasure at smaller scale}

We selected 50 celebrities as target concepts from the celebrity list provided by \citet{lu2024mace}. For each target, we constructed 1,250 prompts using five prompt templates combined with five random seeds.
Following the experimental setup of \cite{lee2025concept}, we defined three domains for remaining concepts: 100 celebrities and 100 artistic styles from \citet{lu2024mace}, and 64 characters derived from Word2Vec embeddings by \citet{church2017word2vec}.
For each remaining domain, we generated 25 images from five templates and five seeds, producing a total of 2,500, 2,500, and 1,600 images, respectively. Additionally, we included COCO-30K to evaluate general capability on image generation. The concepts and prompt templates are listed in \cite{lee2025concept}.

\section{Evaluation Metrics} \label{supp:eval_metrics}
\begin{figure}
    \centering
    \includegraphics[width=\linewidth]{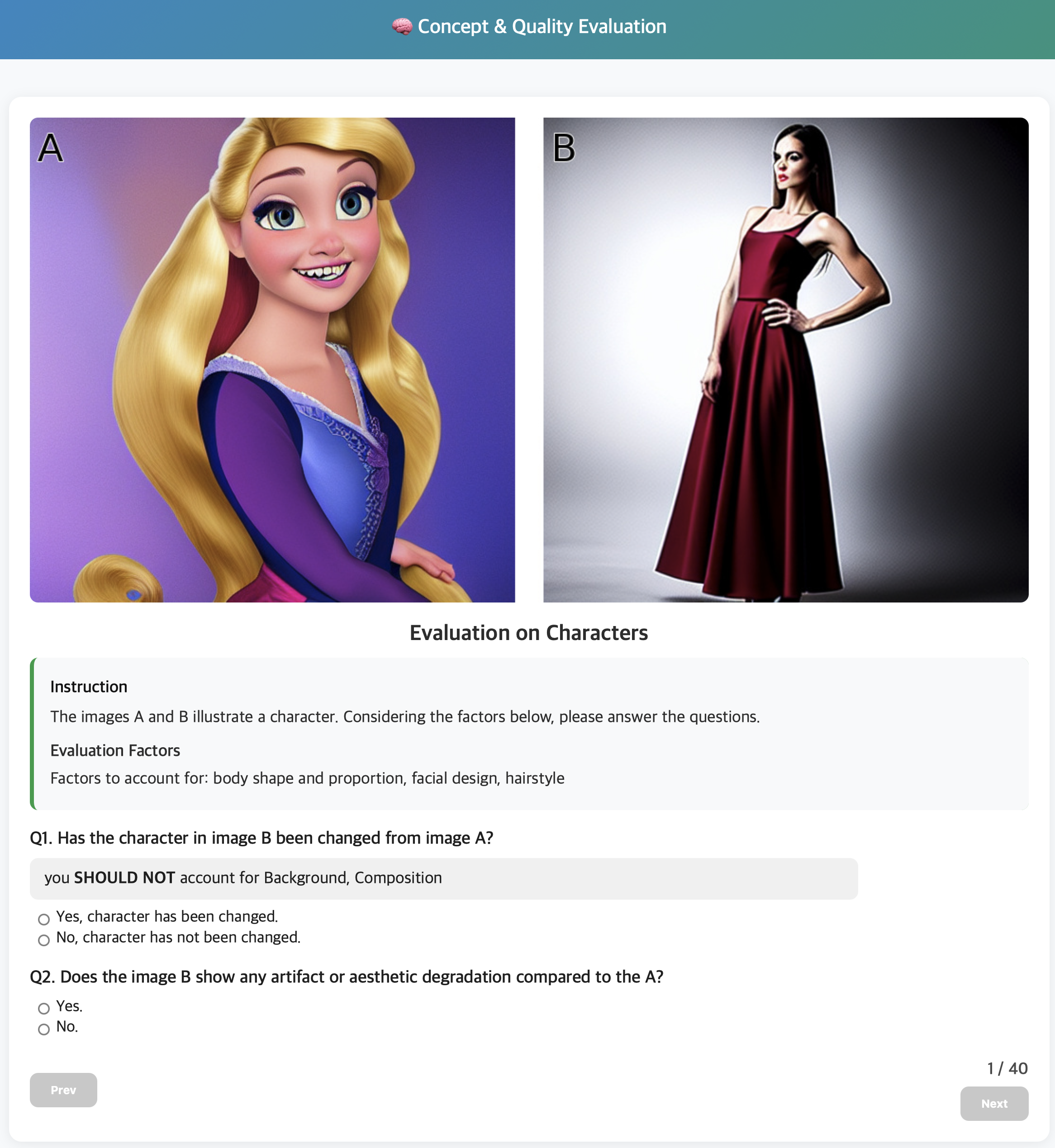}
    \caption{Example of a user study platform on character concept evaluation.}
    \label{fig:platform}
\end{figure}
\paragraph{User study platform.}
We evaluate whether a concept is successfully erased or preserved through a large-scale user study, as illustrated in \cref{fig:platform}. In our A/B interface, image A is always generated by the original model, while image B is generated by one of the comparison methods and configured to test either erasure of the target concept or preservation of the remaining concepts. Users are not given any information about which model produced image B or whether the queried concept is a target or a remaining concept, ensuring that concept erasing and preservation performance, as well as image quality, are assessed in a fully anonymous manner. For each session, the displayed images are refreshed and randomly sampled from a pool of more than 20,000 and 5,000 images for SD v1.4 and SD v3.5-L, respectively, enabling fairer and more reliable results.

\paragraph{CLIP Score (CLIP) \cite{hessel2021clipscore}.}\,\,
We quantify the alignment between a generated image and a text prompt using the CLIP model \cite{radford2021learning}. 
Let $\mathcal{I}$ denote a generated image and $\mathcal{T}$ the corresponding text prompt. 
We extract the image embedding $E_{\mathcal{I}}$ from the CLIP image encoder and the text embedding $E_{\mathcal{T}}$ from its text encoder. 
The CLIP score is defined as the cosine similarity between these embeddings:
\begin{align}
\text{CLIP}(\mathcal{I}, \mathcal{T}) 
= \max\!\left(100 \cdot \cos\!\left(E_{\mathcal{I}}, E_{\mathcal{T}}\right),\, 0\right).
\end{align}
Then, its lower scores indicate stronger suppression of erased concepts, while higher scores reflect faithful retention of the remaining concepts.

\paragraph{Fr{\'e}chet Inception Distance (FID) \cite{heusel2017gans}.}\,\,
We measure the distributional discrepancy between real and generated images using FID for remaining concepts. 
FID computes the Wasserstein-2 distance between feature distributions extracted from a pre-trained network:
\begin{align}
\text{FID}(p, q) 
&= \|\mu_p - \mu_q\|_2^2 \nonumber \\
&\quad + \text{Tr}\!\left(\Sigma_p + \Sigma_q - 2\sqrt{\Sigma_p \Sigma_q}\right),
\end{align}
where $\mu_p, \Sigma_p$ and $\mu_q, \Sigma_q$ denote the means and covariances of the feature distributions $p$ and $q$ for reference and generated images, respectively. 
It jointly captures the discrepancy in mean shift and covariance mismatch. 
Lower FID scores indicate a closer match between the two distributions.

\paragraph{Kernel Inception Distance (KID) \cite{sutherland2018demystifying}.}\,\,
It measures the divergence between reference and real images based on the squared Maximum Mean Discrepancy (MMD) in a learned feature space. For distributions $p$ and $q$, MMD is defined as:
\begin{align}
\text{MMD}(p, q)
&= \mathbb{E}_{x, x' \sim p}[K(x, x')] 
 + \mathbb{E}_{y, y' \sim q}[K(y, y')] \nonumber \\
&\quad - 2\,\mathbb{E}_{x \sim p, y \sim q}[K(x, y)],
\end{align}
where $K$ is a kernel function measuring similarity between feature embeddings. 
KID is unbiased and robust even with limited sample sizes, making it particularly suitable for evaluating generative models across diverse domains. In this work, we report KID values scaled by a factor of 100.

\paragraph{GIPHY Celebrity Detector (GCD)~\cite{hasty_celeb_2024}.}\,\,
It detects faces of celebrities in an image and outputs the top-5 predicted celebrity names along with their associated probabilities, specialized for evaluating concept erasure and preservation for celebrity identities. 
Let $\{(\mathcal{I}_{i}, c_{i})\}_{i=1}^{N}$ denote a set of generated images paired with their corresponding concepts. 
We define the accuracy for target concepts as:
\begin{align}
\text{Acc}_t 
= \frac{1}{N} \sum_{i=1}^{N} 
\mathbb{I}\!\left(\text{GCD}_{\text{top-1}}(\mathcal{I}_i) = c_i\right),
\end{align}
where a sample is considered successfully erased if the detector’s top-1 prediction differs from the true target celebrity.
For the remaining concepts, we include an additional confidence constraint to ensure reliable detection, which is given by:
\begin{align}
\text{Acc}_r = &\frac{1}{N} \sum_{i=1}^{N} 
\mathbb{I}\Big(
\text{GCD}_{\text{top-1}}(\mathcal{I}_i) = c_i \nonumber \\
& \;\land\; P(\text{GCD}_{\text{top-1}}(\mathcal{I}_i)\mid \mathcal{I}_i) \ge 0.9
\Big),
\end{align}
meaning that a remaining celebrity is considered preserved only when GCD identifies it as the top-1 prediction with a confidence of at least $0.9$.

\section{Implementation Details} \label{supp:implementation}

\subsection{Baselines}
For baselines, we conducted experiments using their code by adding the necessary configurations for multi-concept erasing across heterogeneous domains. We provide implementation details of the baselines.

\paragraph{UCE~\cite{gandikota2024unified}.}\,\,
Following the original implementation, we selected a regularization term $\lambda$ that enables the removal of thousands of concepts while maintaining image fidelity; the value was empirically found to be 70,000.

\paragraph{MACE~\cite{lu2024mace}.}\,\, 
In the MACE framework, the preservation scales for the closed-form refinement and multi-LoRA fusion stages are controlled by two hyperparameters, $\lambda_{1}$ and $\lambda_{2}$. We empirically tuned these coefficients and found that setting both to $1.0 \times 10^{-6}$, enabling the erasure of thousands of concepts while largely preserving image fidelity.

\paragraph{CPE~\cite{lee2025concept}.}\,\, 
We followed the CPE training procedure and adopted the hyperparameter configuration used for the artistic-style erasure setting, which applies stronger erasure than the celebrity domain. Specifically, we set the ranks of the shared attention gates to $s_{1}=16$ and $s_{2}=1$, and used an attention-anchor loss coefficient of $\lambda = 1.0 \times 10^{4}$.

\paragraph{SAFREE~\cite{yoonsafree}.}\,\,
SAFREE does not include editable hyperparameters for concept erasure, so we directly followed the official implementation and applied the method to both SDv1.4 and SDv3.5-L.

\paragraph{SPEED~\cite{li2025speed}.}\,\,
In SPEED, we used an augmentation times of $N_A = 10$ and set the effective augmentation rank via a singular-value threshold of $10^{-1}$. The default filtering scale $\alpha = 1$ is adopted, corresponding to the original prior-filtering strength. In addition, we performed $k$-means clustering with $k=3$ to construct the null prior basis from the unconditional prompt. The same configuration is applied to both SDv1.4 and SDv3.5-L.

\subsection{Details on ETC}

In this work, the proposed Erase-Then-Compose (ETC) framework is explored on four erasure benchmarks: 2,072 concepts from heterogeneous domains, 50 celebrities, and explicit content with SDv1.4 and 515 concepts from heterogeneous domains with SDv3.5-L. 
For all settings, we freeze the diffusion backbone and the text encoder and update only the parameters of the MoEraser modules. For all experiments, we used a single A6000 GPU and an AMD EPYC 7763 64-Core Processor for data construction, training, and generation.
The key hyperparameters, including the number of mapping concepts, the confidence level $\tau_1$ for separating target and anchor embeddings, and the MoE architecture parameters, are summarized in \cref{tab:appdx_impl_config}.

\noindent\textbf{2,072 concepts erasure.}\,\,
For this benchmark, we used three mapping concepts per target concept and set the confidence level to $\tau_1 = 0.8$ when sampling target and anchor embeddings. 
The MoEraser employs a hidden dimensionality of 1536 with a single layer, $16$ experts, and Top-$k$ routing with $k = 12$ experts activated per token.
We train ETC with anchoring coefficient $\lambda = 10$, a learning rate of $0.01$, a batch size of $512$, and $16{,}384$ training iterations.

\noindent\textbf{515 concepts erasure.}\,\,
We again use three mapping concepts and increase the confidence level to $\tau_1 = 0.9$.
The MoEraser is configured with a hidden dimensionality of 2048 with single depth, $8$ experts, and Top-$k$ routing with $k = 6$ selected experts.
We train with $\lambda = 10$, learning rate $0.01$, batch size $512$, and $16{,}384$ training iterations.

\noindent\textbf{50 celebrities erasure.}\,\,
We keep the number of mapping concepts at three and set the confidence level to $\tau_1 = 0.9$.
Because the number of target concepts is much smaller than in the heterogeneous benchmarks, we reduce the MoEraser capacity to a hidden dimensionality of 192 while maintaining one layer, $8$ experts, and Top-$k$ routing with $k = 6$ experts.
We train with $\lambda = 10$, a learning rate of $0.01$, batch size $256$, and $8{,}192$ iterations.

\noindent\textbf{Explicit content erasure.}\,\,
For explicit content erasure, we also employ SD v1.4 and use only one mapping concepts per target concept.
To enable erasure over a broader region of the embedding space, we used a higher confidence level of $\tau_1 = 0.98$.
The MoEraser is configured with a hidden dimensionality of 384, with single depth, $4$ experts, and Top-$k$ routing with $k = 3$ experts.
We set $\lambda = 300$ to strongly preserve the other regions.
The learning rate, batch size, and number of iterations are fixed to $0.01$, $256$, and $8{,}192$, respectively.

\begin{table}[t]
    \caption{Key configuration of ETC for each erasure task.}
    \centering    
    \resizebox{\columnwidth}{!}{
    \begin{tabular}{c|c|c|c|c}
    \toprule
      & 2,072 Hetero. & 515 Hetero. & 50 Celebs  & Explicit  \\
    
    \midrule

    Model & SDv1.4 & SDv3.5-L & SDv1.4 & SDv1.4 \\

    \midrule
    
    \# Mappings 
    & 3 & 3 & 3 & 1 \\ 
    $\tau_1$ 
    & 0.8 & 0.9 & 0.9 & 0.98  \\ 
    
    \midrule

    Hidden dim. 
    & 1536 & 2048 & 192 & 384 \\ 
    Depth.  
    & 1 & 1 & 1 & 1  \\
    \# Experts 
    & 16 & 8 & 8 & 4 \\ 
    \# Top-$k$ Exp. 
    & 12 & 6 & 6 & 3 \\ 
    
    \midrule
    $\lambda$ 
    & 10 & 10 & 30 & 300 \\ 
    LR 
    & 0.01 & 0.01 & 0.01 & 0.01 \\ 
    Batchsize 
    & 512 & 512 & 256 & 256 \\ 
    Train Iter. 
    & 16384 & 16384 & 8192 & 8192 \\ 
         
    \bottomrule
        
    \end{tabular}
    }
    \label{tab:appdx_impl_config}
\end{table}


\section{Efficiency studies} \label{supp:efficiency}

\paragraph{Memory efficiency.}\,\,
Table~\ref{tab:appdx_table_memory} summarizes the parameter overhead introduced by ETC across diverse concept-erasure benchmarks. 
Despite operating on large and heterogeneous concept pools (e.g., 2{,}072 mixed-domain concepts), our method requires only 75.03M trainable parameters on SDv1.4, corresponding to about $7.63\%$ of the original model. 
For erasing 515 concepts with SDv3.5-L, ETC accounts for only $1.62\%$ of the total parameters. 
Smaller benchmarks, including the 50 celebrities and explicit erasure settings, further reduce the overhead to below 1\%. 
These results demonstrate that ETC achieves scalable concept erasure while imposing only a marginal memory footprint, making it suitable for real-world deployments where parameter efficiency is critical.

\begin{table}[t]
    \centering
    \caption{Memory consumption of ETC on various concept erasure benchmarks, demonstrating the memory efficiency of ETC.}
    \label{tab:appdx_table_memory}
    \resizebox{1.0\columnwidth}{!}{
    \begin{tabular}{c|c|c|c|c}
        \toprule
         & 2,072 Hetero. & 515 Hetero. & 50 Celebs & Explicit  \\
        \midrule
        Model & SDv1.4 & SDv3.5-L & SDv1.4 & SDv1.4 \\
        \midrule
        \# Params & 75.01M & 130.18M & 5.48M & 5.94M \\
        Ratio of params & $\simeq 7.63\%$ & $\simeq 1.62\%$ & $\simeq 0.62\%$ & $\simeq 0.69\%$ \\
        \hline
    \end{tabular}
    }
\end{table}

\paragraph{Computation efficiency.}\,\,
Table~\ref{tab:appdx_table_flops} reports computational cost of ETC measured in A6000 GPU hours and additional FLOPs introduced by ETC. 
Data preparation and training incur 7.37 GPU hours in total for erasing 2{,}072-concept setting and the total times reduces to 0.27 hours for the explicit concept benchmark. 
Noticeably, the additional computational overhead introduced by ETC is negligible relative to the original diffusion model. For example, the additional FLOPs represent merely 10.91G on SDv1.4 and 17.45G on SDv3.5-L, which are negligible compared to the original FLOPs (678.02G and 22.52T, respectively). 
Since ETC is applied only once at the first denoising step during inference, the method adds virtually no extra cost to the generation pipeline. 
These results verify that ETC enables effective large-scale concept erasure without compromising computational efficiency. 

\begin{table}[t]
    \centering
    \caption{Computational costs on diverse concept erasure benchmarks in A6000 GPU hours, demonstrating the efficiency of ETC.}
    \label{tab:appdx_table_flops}
    \resizebox{1.0\columnwidth}{!}{
    \begin{tabular}{c|c|c|c|c}
        \toprule
         & 2,072 Hetero. & 515 Hetero. & 50 Celebs & Explicit \\
        \midrule
        Model & SDv1.4 & SDv3.5-L & SDv1.4 & SDv1.4 \\
        \midrule
        Data Prep. Time (h) & 5.58 & 1.71 & 0.73 & 0.10 \\
        Training Time (h) & 2.76 & 2.17 & 0.17 & 0.17 \\
        Total Time (h) & 7.37 & 3.88 & 0.91 & 0.27 \\
        \midrule
        Org. FLOPs & 678.02G & 22.52T & 678.02G & 678.02G \\
        Add. FLOPs & 10.91G & 17.45G & 0.547G & 0.682G \\
        \hline
    \end{tabular}
    }
\end{table}

\section{Additional Quantitative Results} \label{supp:additional_quanti}
\begin{table}
    \caption{CLIP Score and FID on erasing 2,072 heterogenous concept erasing.}
    \label{tab:coco}
    \centering
    \resizebox{0.6\linewidth}{!}{  
    \begin{tabular}{@{}l|ccc|c}
    \toprule
         &  MACE & CPE & ETC (Ours) & SDv1.4\\
    \midrule
        CLIP Score$\uparrow$ & \cellcolor{tabthird}18.24 & \cellcolor{tabsecond}30.64 & \cellcolor{tabfirst}31.28 & 31.34\\
        FID$\downarrow$ & \cellcolor{tabthird}88.34 & \cellcolor{tabsecond}16.38 & \cellcolor{tabfirst}13.96 & 14.04\\
    \bottomrule
    \end{tabular}
    }
    \vspace{-1em}
\end{table}
\subsection{FID and CLIP score on 2,072 concept erasing}
In the main manuscript, we measured the Kernel Inception Distance (KID) over 1,000 samples to examine distributional changes in images generated by the original model after erasing 2,072 concepts from SDv1.5. We further extend this study by evaluating on the COCO-30K dataset, where we generate images using the full set of captions for each concept erasure method and measure both FID and CLIP score, as reported in \cref{tab:coco}. The results show that ETC achieves significantly higher CLIP scores and lower FID compared to other baselines, indicating that it preserves the original model’s image generation capability.

\begin{table*}[t]
    \caption{Automated metric at scale on Celeb./Art./Char. }
    \label{tab:sup_automated}
    \centering
    \resizebox{\linewidth}{!}{  
    \begin{tabular}{l|ccc|ccc|ccc|ccc|ccc}
    \toprule
         & \multicolumn{3}{c|}{SAFREE} & \multicolumn{3}{c|}{SPEED} & \multicolumn{3}{c|}{MACE} & \multicolumn{3}{c|}{CPE} & \multicolumn{3}{c}{ETC \textbf{(Ours)}}\\
        & Celeb. & Art. & Char. & Celeb. & Art. & Char. & Celeb. & Art. & Char. & Celeb. & Art. & Char. & Celeb. & Art. & Char. \\
    \midrule
        $\text{CRS}_t\downarrow$ & 0.97 & 0.92 & 0.97 & \cellcolor{tabthird}0.20 & \cellcolor{tabthird}0.19 & \cellcolor{tabfirst}0.03 & 0.52 & 0.63 & 0.20 & \cellcolor{tabfirst}0.02 & \cellcolor{tabfirst}0.02 & \cellcolor{tabthird}0.27 & \cellcolor{tabsecond}0.15 & \cellcolor{tabsecond}0.13 & \cellcolor{tabsecond}0.14  \\
        $\text{CRS}_r\uparrow$ & \cellcolor{tabsecond}0.89 & \cellcolor{tabfirst}0.94 & \cellcolor{tabfirst}0.89 & 0.26 & 0.19 & 0.09 & 0.60 & 0.63 & \cellcolor{tabthird}0.38 & \cellcolor{tabthird}0.67 & \cellcolor{tabthird}0.76 & \cellcolor{tabsecond}0.44 & \cellcolor{tabfirst}0.96 & \cellcolor{tabsecond}0.87 & \cellcolor{tabfirst}0.89  \\
        $H_0\uparrow$ & 0.07 & 0.14 & 0.14 & 0.40 & 0.16 & 0.31 & \cellcolor{tabthird}0.53 & \cellcolor{tabthird}0.33 & \cellcolor{tabthird}0.47 & \cellcolor{tabsecond}0.80 & \cellcolor{tabsecond}0.74 & \cellcolor{tabsecond}0.61 & \cellcolor{tabfirst}0.90 & \cellcolor{tabfirst}0.87 & \cellcolor{tabfirst}0.88  \\
        KID $(\times10^3)\downarrow$ & \cellcolor{tabsecond}0.72 & \cellcolor{tabsecond}0.63 & \cellcolor{tabsecond}0.25 & 3.57 & 4.34 & 2.32 & 7.23 & 7.85 & \cellcolor{tabthird}1.34& \cellcolor{tabthird}2.69 & \cellcolor{tabthird}1.08 & 2.76 & \cellcolor{tabfirst}0.09 & \cellcolor{tabfirst}0.29 & \cellcolor{tabfirst}0.15 \\
    \bottomrule
    \end{tabular}
    }
\end{table*}

\subsection{Automated evaluation}
Because the notion of a concept can be subjective, it is difficult to determine quantitatively whether it has changed and equally difficult to automate such an assessment. Therefore, we evaluate the performance of ETC through a large-scale user study. Beyond this user study, however, we additionally introduce a more unbiased evaluation protocol by adopting an MLLM-based metric inspired by the Holistic Unlearning Benchmark~\cite{moon2024holistic}, and report the results in \cref{tab:sup_automated}. We use Qwen2.5-VL 32B~\cite{Qwen2.5-VL} as the MLLM evaluator, and the image set is identical to the image pool used in the user study. The results show that ETC effectively removes the target concept while preserving the remaining concepts, achieving the best performance in balancing erasure and preservation ($H_0$) while maintaining image quality (KID). These findings are also consistent with the user study results.

\subsection{Explicit content erasure}
Removing explicit concepts (\textit{e.g.}, nudity) is essential for the safe use of generative AI systems. Although the main paper focuses on identity concept erasure, we additionally evaluate ETC on explicit concept erasure using the I2P benchmark. The target explicit concepts and their corresponding mapping concepts used in this experiment are summarized in \cref{sec:explicit}, and the quantitative results are reported in \cref{tab:explicit}. Even though this experiment includes implicit prompts that do not contain explicit keywords and were not used during training, ETC successfully handles such prompts and achieves comparable or superior performance to state-of-the-art concept erasure methods, thereby demonstrating its robustness. We provide qualitative results of explicit concept erasure in \cref{fig:explicit}.

We hypothesize that ETC’s robustness to implicit prompts arises from modeling concept distributions and erasing these distributions. To verify this, we replace tMM modeling and AOT with a direct word-to-word mapping and evaluate the explicit concept erasing performance; the results are reported in \cref{tab:explicit_ablation}. Although direct word-to-word mapping is still effective at suppressing explicit concepts, it yields substantially lower explicit concept erasing performance, confirming that distribution modeling plays a critical role in both robustness and overall performance.

\begin{table}[t]
    \caption{Results of detected number of explicit contents using NudeNet detector on I2P, along with preservation performance on COCO-30K measured using CS and FID.}
    \label{tab:explicit}
    \centering
    
    \setlength{\tabcolsep}{2pt}
    \renewcommand{\arraystretch}{1.05}

    \resizebox{1.0\columnwidth}{!}{%
        \begin{tabular}{@{}c|ccccc|cc@{}}
        
            \hline
            
            \multirow{2}{*}{Method} & \multicolumn{5}{c}{Number of detections by NudeNet} & \multicolumn{2}{|c}{COCO-30K} \\
            
            \cmidrule(lr){2-6} \cmidrule(lr){7-8}
            
            & Buttocks & Breasts (F) & Genitalia (F) & Breasts (M) & Genitalia (M) & CS $\uparrow$ & FID $\downarrow$ \\
            
            \hline
            ESD-x \cite{gandikota2023erasing} 
            & 12 & 100 & 4 & 30 & 8 & 30.69 & 14.41 \\
            
            ESD-u \cite{gandikota2023erasing} 
            & \cellcolor{tabsecond}2 & 35 & \cellcolor{tabthird}3 & \cellcolor{tabthird}9 & \cellcolor{tabsecond}2 & 30.21 & 15.10 \\

            UCE \cite{gandikota2024unified} 
            & 7 & 35 & 5 & 11 & 4 & 30.85 & 14.07 \\
            
            MACE \cite{lu2024mace} 
            & \cellcolor{tabsecond}2 & 16 & \cellcolor{tabfirst}0 & \cellcolor{tabthird}9 & 7 & 29.41 & \cellcolor{tabfirst}13.42 \\
            
            RECE \cite{gong2024reliable} 
            & \cellcolor{tabthird}3 & \cellcolor{tabthird}10 & \cellcolor{tabfirst}0 & \cellcolor{tabthird}9 & \cellcolor{tabthird}3 & \cellcolor{tabthird}30.95 & - \\
            
            CPE \cite{lee2025concept}
            & \cellcolor{tabsecond}2 & \cellcolor{tabfirst}6 & \cellcolor{tabsecond}1 & \cellcolor{tabfirst}3 & \cellcolor{tabsecond}2 & \cellcolor{tabsecond}31.19 & \cellcolor{tabsecond}13.89  \\


            \textbf{Ours}
            & \cellcolor{tabfirst}0 & \cellcolor{tabsecond}8 & \cellcolor{tabfirst}0 & \cellcolor{tabsecond}5 & \cellcolor{tabfirst}1 & \cellcolor{tabfirst}31.24 & \cellcolor{tabthird}14.06  \\
            
            \hline
            SD v1.4 \citep{rombach2022stable1.4} 
            & 29 & 266 & 18 & 42 & 7 &  31.34 & 14.04  \\
            
            SD v2.1 \citep{rombach2022stable2.0} 
            & 17 & 177 & 9 & 57 & 2  & 31.53 & 14.87 \\
            \hline
        \end{tabular}
    }
    \vskip -0.1in
\end{table}

\begin{table}[t]
    \caption{Results of detected number of explicit contents using NudeNet detector on I2P, along with preservation performance on COCO-30K measured using CS and FID.}
    \label{tab:explicit_ablation}
    \centering
    \setlength{\tabcolsep}{2pt}
    \renewcommand{\arraystretch}{1.05}

    \resizebox{1.0\columnwidth}{!}{%
        \begin{tabular}{@{}c|ccccc|cc@{}}
        
            \hline
            
            \multirow{2}{*}{Modeling} & \multicolumn{5}{c}{Number of detections by NudeNet} & \multicolumn{2}{|c}{COCO-30K} \\
            
            \cmidrule(lr){2-6} \cmidrule(lr){7-8}
            
            & Buttocks & Breasts (F) & Genitalia (F) & Breasts (M) & Genitalia (M) & CS $\uparrow$ & FID $\downarrow$ \\
            
            \hline
            
            Direct \cite{gandikota2023erasing} 
            & 6 & 42 & 12 & 26 & 5 & 30.97 & 13.81 \\
            
            \textbf{tMM}
            & 0 & 8 & 0 & 5 & 1 & 31.24 & 14.06  \\
            
            \hline
            SD v1.4 \citep{rombach2022stable1.4} 
            & 29 & 266 & 18 & 42 & 7 &  31.34 & 14.04  \\
            
            \hline
        \end{tabular}
    }
    \vskip -0.1in
\end{table}


\begin{table}[t]
    \centering
    \setlength{\tabcolsep}{10pt}
    \renewcommand{\arraystretch}{1.0}
    \caption{Attack success rate (\%) to measure robustness against attack methods: Ring-A-Bell (RAB)~\cite{tsai2024ring} \& UnlearnDiff (UD)~\cite{zhang2023generate}.} 
    \label{tab:comparison_robustness}
    \vskip -0.05in
    \resizebox{1.0\columnwidth}{!}{%
        \begin{tabular}{c|ccccccc}
            \hline

            Method 
            & FMN & ESD & UCE 
            & MACE & RECE & CPE & ETC \\

            \hline

            RAB $\downarrow$ 
            & 80.85 & 61.70 & 35.46
            &\cellcolor{tabthird}4.26 & 13.38 & \cellcolor{tabfirst}0.00 
            & \cellcolor{tabsecond}1.42 \\

            UD $\downarrow$ & 
            97.89 & 76.05 & 79.58 
            & 66.90 &\cellcolor{tabthird}65.46 & \cellcolor{tabfirst}30.28 
            & \cellcolor{tabsecond}52.82 \\
            
            \hline
        \end{tabular}
        
    }
    \vskip -0.15in
\end{table}

\begin{table}
    \caption{Attack success rate (ASR) and train / sample time.}
    \label{tab:sup_asr}
    \centering
    \resizebox{\linewidth}{!}{  
    \begin{tabular}{l|ccccc}
    \toprule
         &  CPE w/o RARE & CPE & ETC & ETC w/ RARE\\
    \midrule
        ASR-RAB (\%) & \cellcolor{tabthird}10.64 & \cellcolor{tabfirst}0 & \cellcolor{tabsecond}1.42 & \cellcolor{tabfirst}0 \\
        ASR-UnlearnDiff (\%) & 85.21 & \cellcolor{tabsecond}30.28 & \cellcolor{tabthird}52.82 & \cellcolor{tabfirst}28.81 \\
        Training + Sampling time (hrs) & \cellcolor{tabsecond}0.36 & 5.46 & \cellcolor{tabfirst}0.35 & \cellcolor{tabthird}3.31 \\
    \bottomrule
    \end{tabular}
    }
\end{table}
\subsection{Robustness to adversarial attacks}
We achieve robustness by modeling concept embedding distributions and applying Noise-Injection Restoration tuning, which enhances resistance to white-box attacks such as model removal. Beyond these settings, we further evaluate ETC under advanced adversarial attacks, including Ring-A-Bell (RAB)~\cite{tsai2024ring} and the attack introduced in UnlearnDiff (UD)~\cite{zhang2023generate}, with results summarized in \cref{tab:comparison_robustness}. ETC demonstrates the second-highest robustness against both RAB and UD, following CPE. Notably, CPE explicitly improves robustness by incorporating adversarial attacks during training and introducing an additional erasure stage (RARE). In contrast, ETC achieves comparable robustness without such explicit robustness-enhancement procedures, highlighting the effectiveness of modeling and erasing concept embedding distributions. Furthermore, we verify in \cref{tab:sup_asr} that ETC can obtain additional robustness to adversarial attacks by incorporating RARE. We also provide qualitative comparisons of robustness under adversarial attacks in \cref{fig:robustness}.

\begin{table}
    \caption{Wall-clock time comparison for each methods' pipeline.}
    \label{tab:sup_efficiency}
    \centering
    \resizebox{\linewidth}{!}{  
    \begin{tabular}{l|ccc|l|ccc}
    \toprule
       50 Celeb.  & MACE & CPE & ETC & 2072 Hetero. & MACE & CPE & ETC\\
    \midrule
        Data prep. (hrs) & \cellcolor{tabthird}1.02 & \cellcolor{tabfirst}0.0 & \cellcolor{tabsecond}0.73 & Data prep. (hrs) & \cellcolor{tabthird}42.2 & \cellcolor{tabfirst}0.0 & \cellcolor{tabsecond}5.58 \\
        Train (hrs) & \cellcolor{tabsecond}0.93 & \cellcolor{tabthird}6.0 & \cellcolor{tabfirst}0.17 & Train (hrs) & \cellcolor{tabsecond}38.5 & \cellcolor{tabthird}241.73 & \cellcolor{tabfirst}2.76 \\
        Total (hrs) & \cellcolor{tabsecond}1.95 & \cellcolor{tabthird}6.0 & \cellcolor{tabfirst}0.91 & Total (hrs) & \cellcolor{tabsecond}80.7 & \cellcolor{tabthird}241.73 & \cellcolor{tabfirst}7.37 \\
    \midrule
        Latency (sec/img) & \cellcolor{tabfirst}5.61 & \cellcolor{tabthird}6.45 & \cellcolor{tabsecond}5.76 & Latency (sec/img) & \cellcolor{tabfirst}5.61 & \cellcolor{tabthird}7.03 & \cellcolor{tabsecond}6.17 \\
    \bottomrule
    \end{tabular}
    }
\end{table}

\subsection{Computational cost.}
In \cref{tab:sup_efficiency}, we compare the computational cost of ETC with other competitive methods, including MACE and CPE. ETC requires slightly more data preparation time than MACE, while CPE does not require any data, resulting in ETC exhibiting moderate efficiency in this stage. However, when considering the overall training time, ETC achieves the most efficient training performance. For inference latency, MACE can be merged into the model through linear operations, leading to latency that is independent of the number of concepts to be erased. In contrast, CPE attaches modules in parallel, causing latency to increase as the number of erased concepts grows. ETC also introduces a small amount of latency, but remains significantly more scalable than CPE in terms of latency.

\section{Additional Qualitative Results} \label{supp:additional_quali}
\begin{figure}[H]
    \centering
    \includegraphics[width=0.88\linewidth]{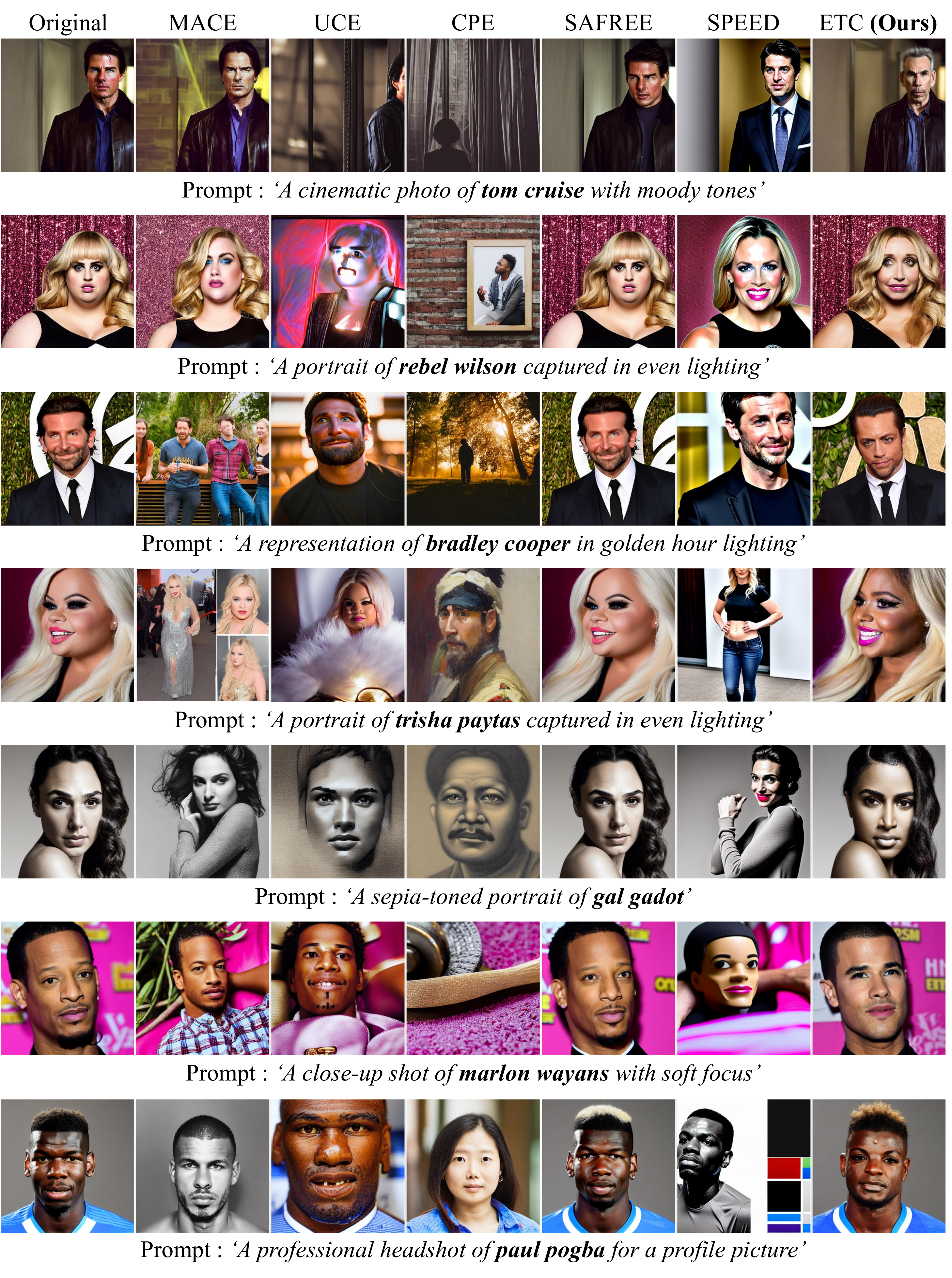}
    \caption{Qualitative results on celebrities erasure from SDv1.4. The images on the same row are generated using the same seed.}
    \label{fig:supp_sd14_era_celeb}
\end{figure}

\begin{figure}[H]
    \centering
    \includegraphics[width=0.88\linewidth]{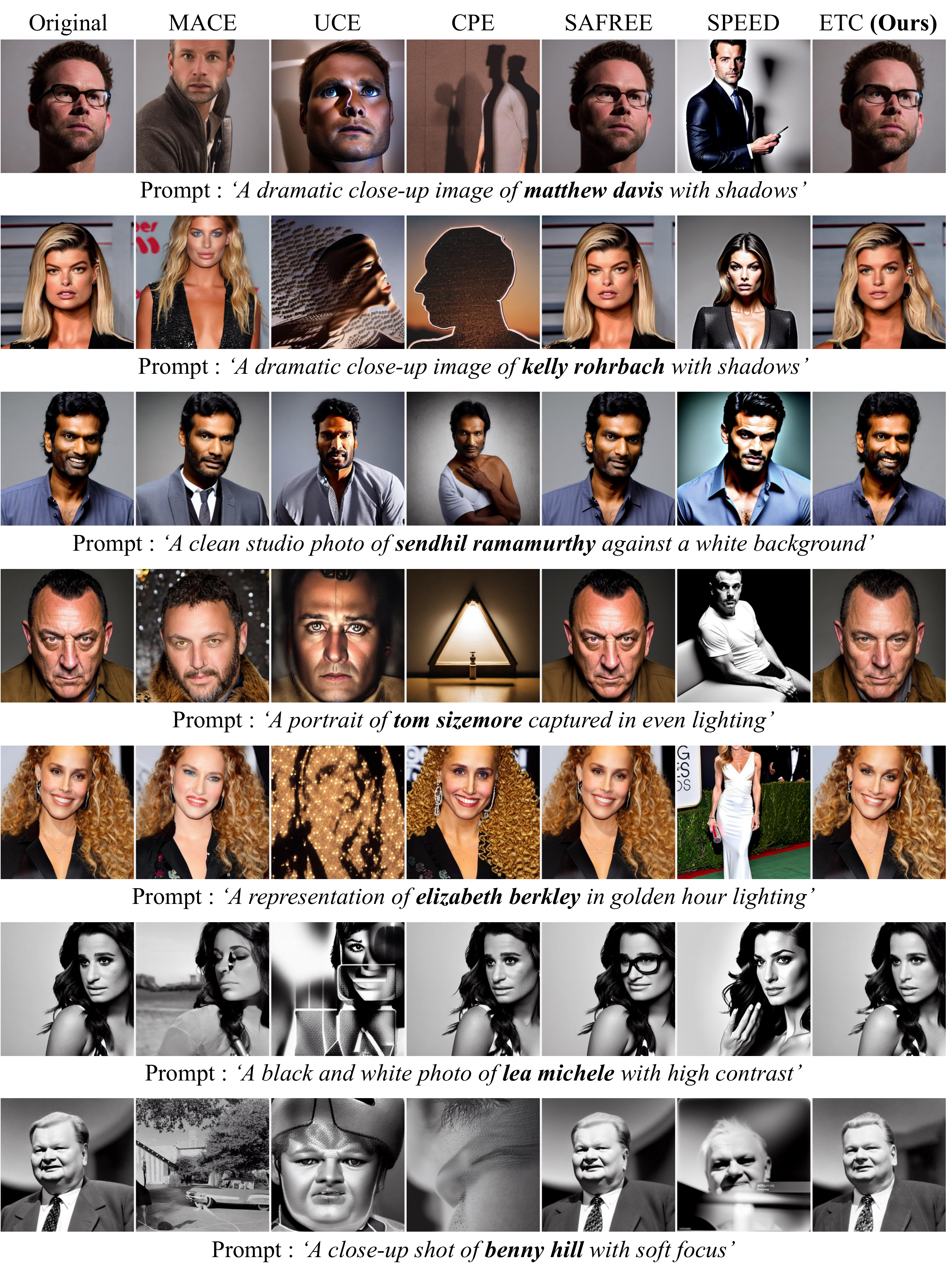}
    \caption{Qualitative results on celebrities remaining from SDv1.4. The images on the same row are generated using the same seed.}
    \label{fig:supp_sd14_rem_celeb}
\end{figure}

\begin{figure}[H]
    \centering
    \includegraphics[width=0.88\linewidth]{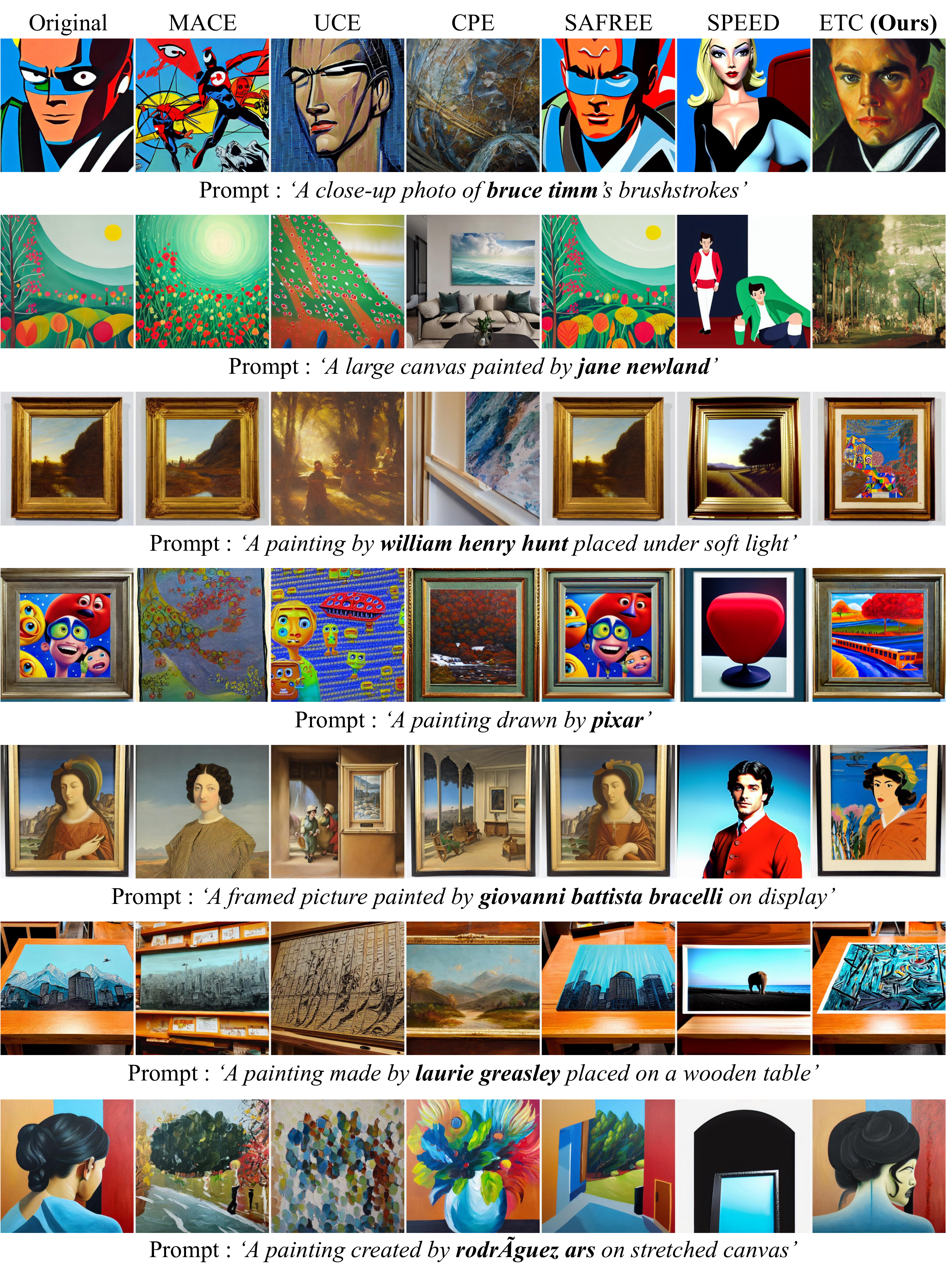}
    \caption{Qualitative results on artistic styles erasure from SDv1.4. The images on the same row are generated using the same seed.}
    \label{fig:supp_sd14_era_art}
\end{figure}

\begin{figure}[H]
    \centering
    \includegraphics[width=0.88\linewidth]{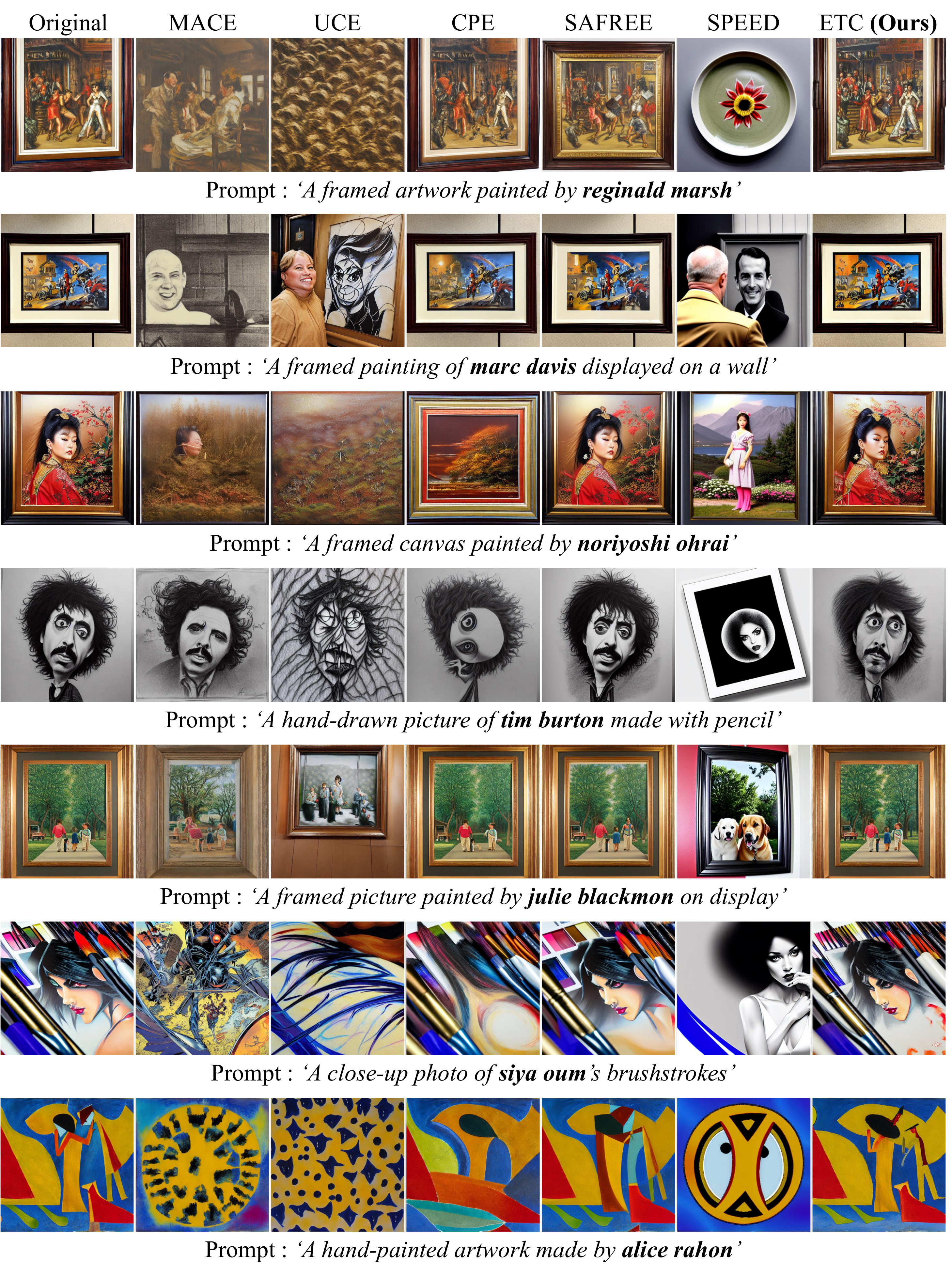}
    \caption{Qualitative results on artistic styles remaining from SDv1.4. The images on the same row are generated using the same seed.}
    \label{fig:supp_sd14_rem_art}
\end{figure}

\begin{figure}[H]
    \centering
    \includegraphics[width=0.88\linewidth]{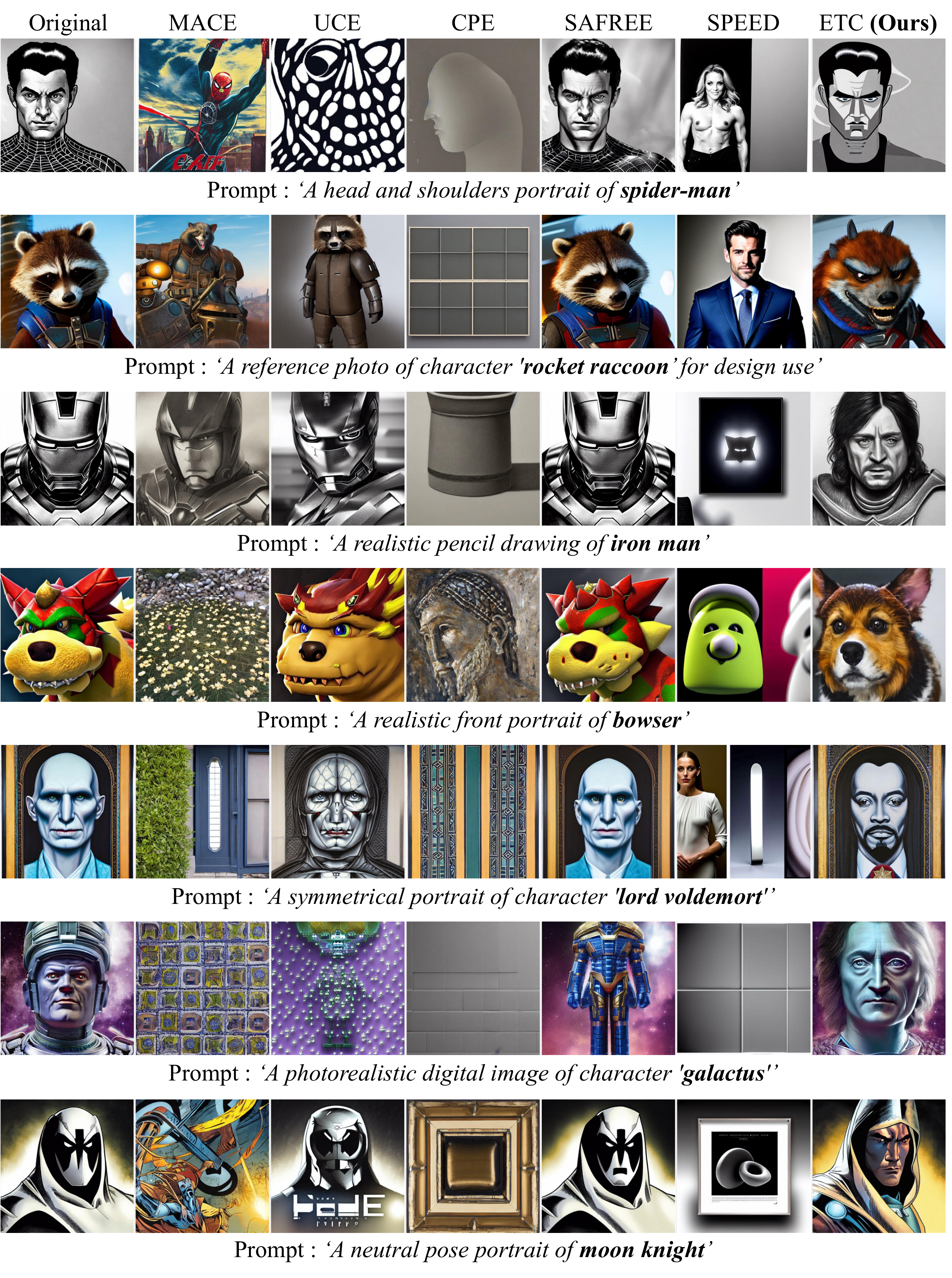}
    \caption{Qualitative results on characters erasure from SDv1.4. The images on the same row are generated using the same seed.}
    \label{fig:supp_sd14_era_char}
\end{figure}

\begin{figure}[H]
    \centering
    \includegraphics[width=0.88\linewidth]{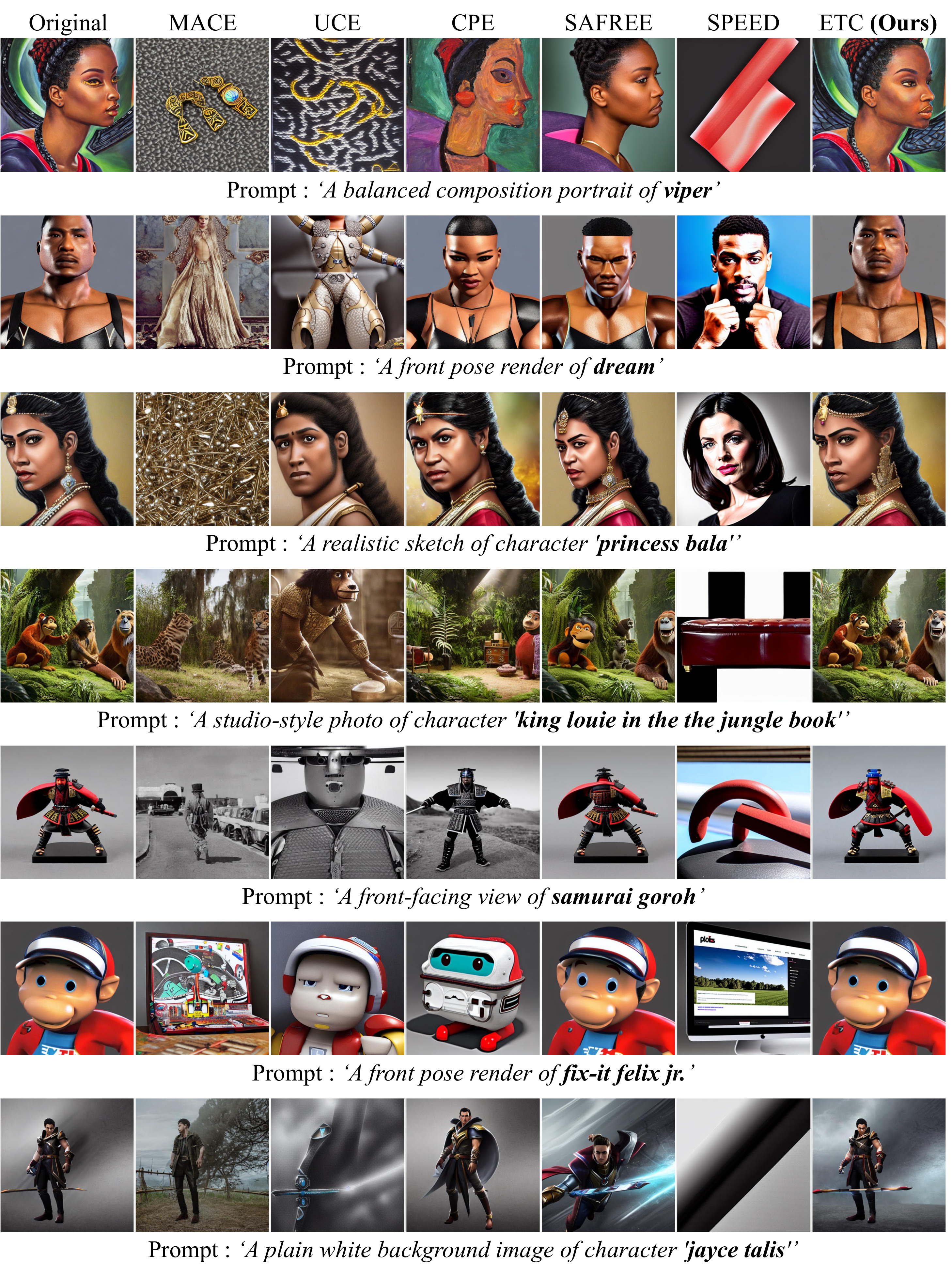}
    \caption{Qualitative results on characters remaining from SDv1.4. The images on the same row are generated using the same seed.}
    \label{fig:supp_sd14_rem_char}
\end{figure}

\begin{figure}[H]
    \centering
    \includegraphics[width=\linewidth]{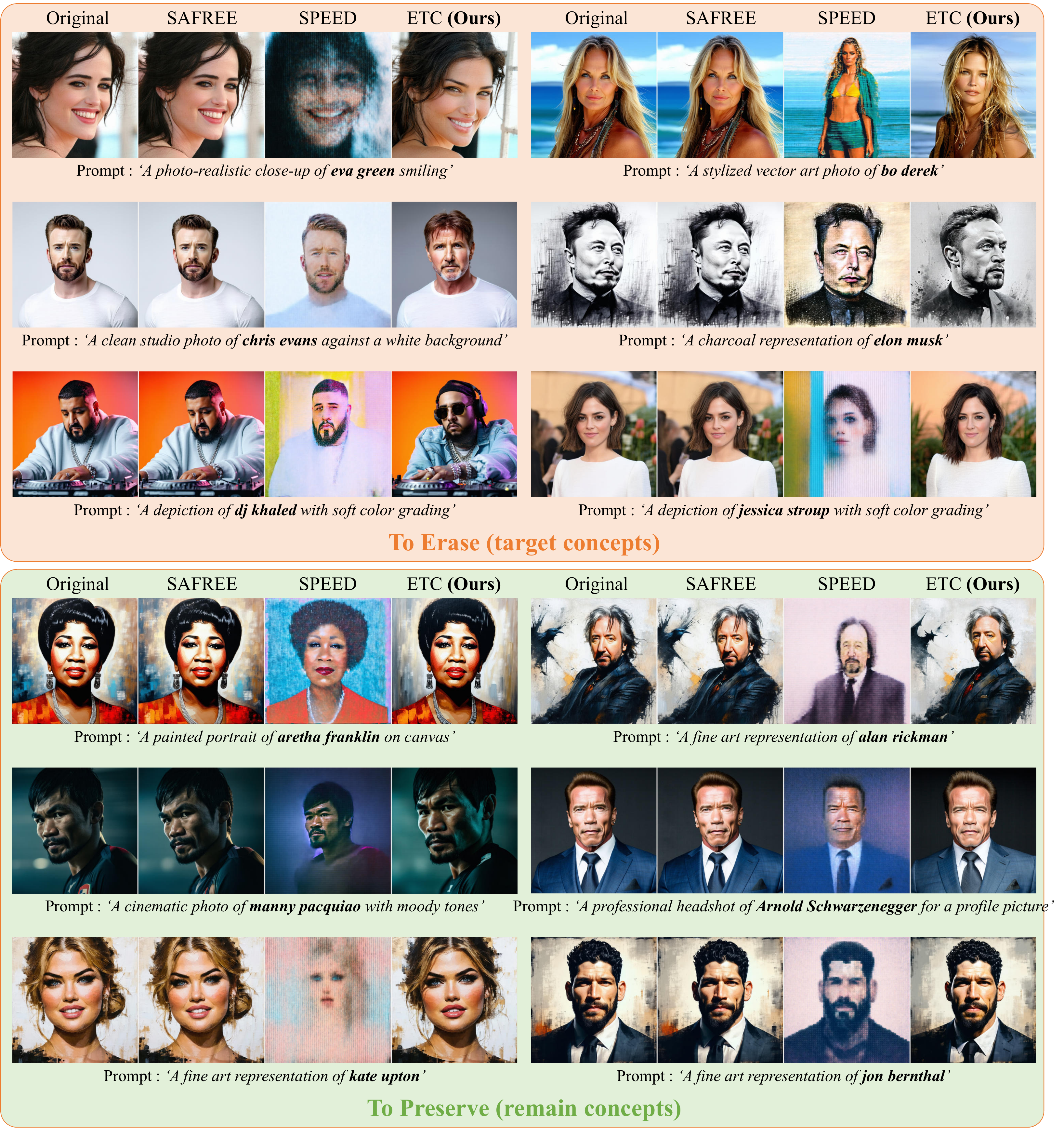}
    \caption{Qualitative results on celebrity concept with SDv3.5.}
    \label{fig:supp_sd35_celeb}
\end{figure}

\begin{figure}[H]
    \centering
    \includegraphics[width=\linewidth]{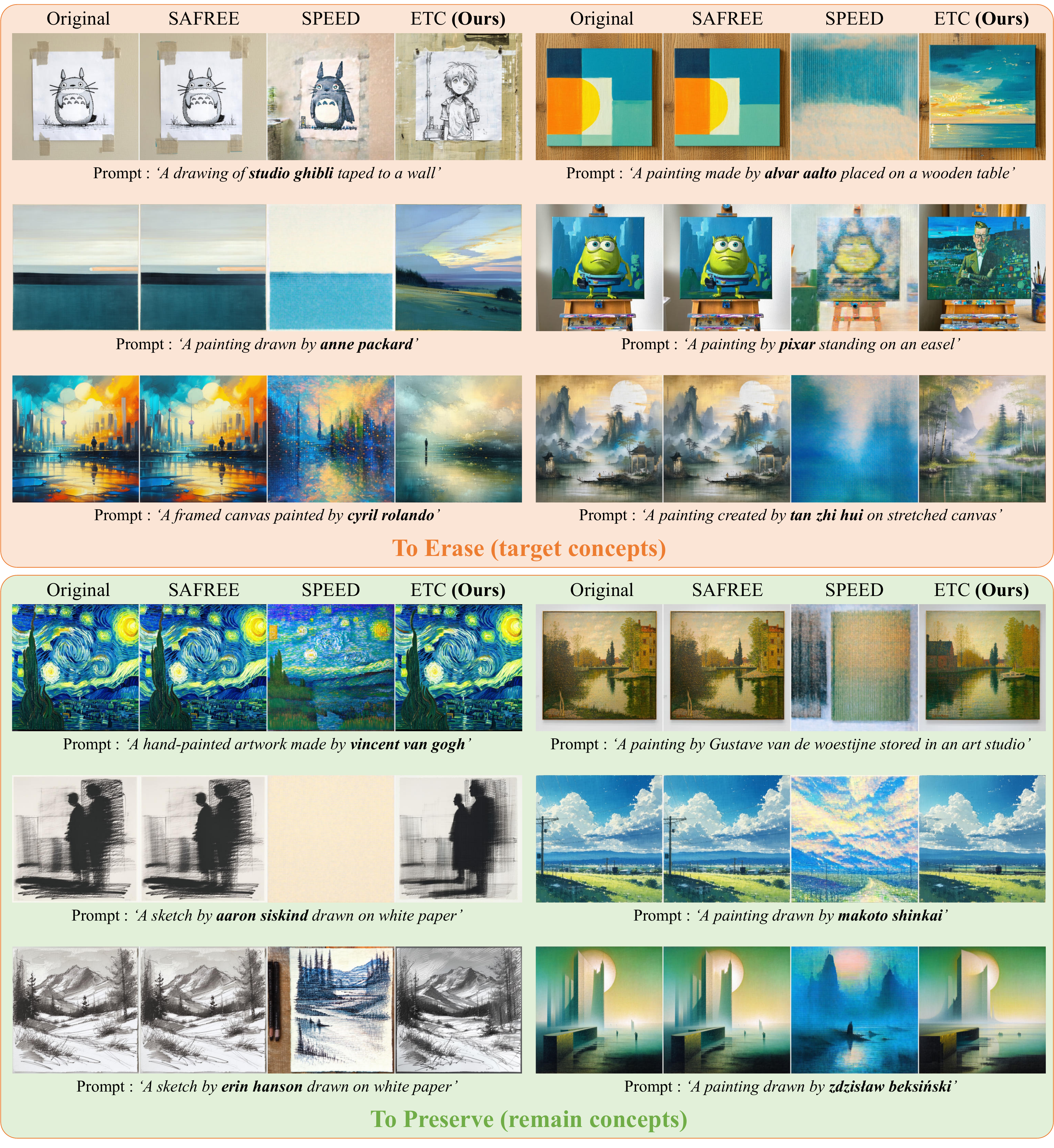}
    \caption{Qualitative results on artistic style concept with SDv3.5.}
    \label{fig:supp_sd35_art}
\end{figure}

\begin{figure}[H]
    \centering
    \includegraphics[width=\linewidth]{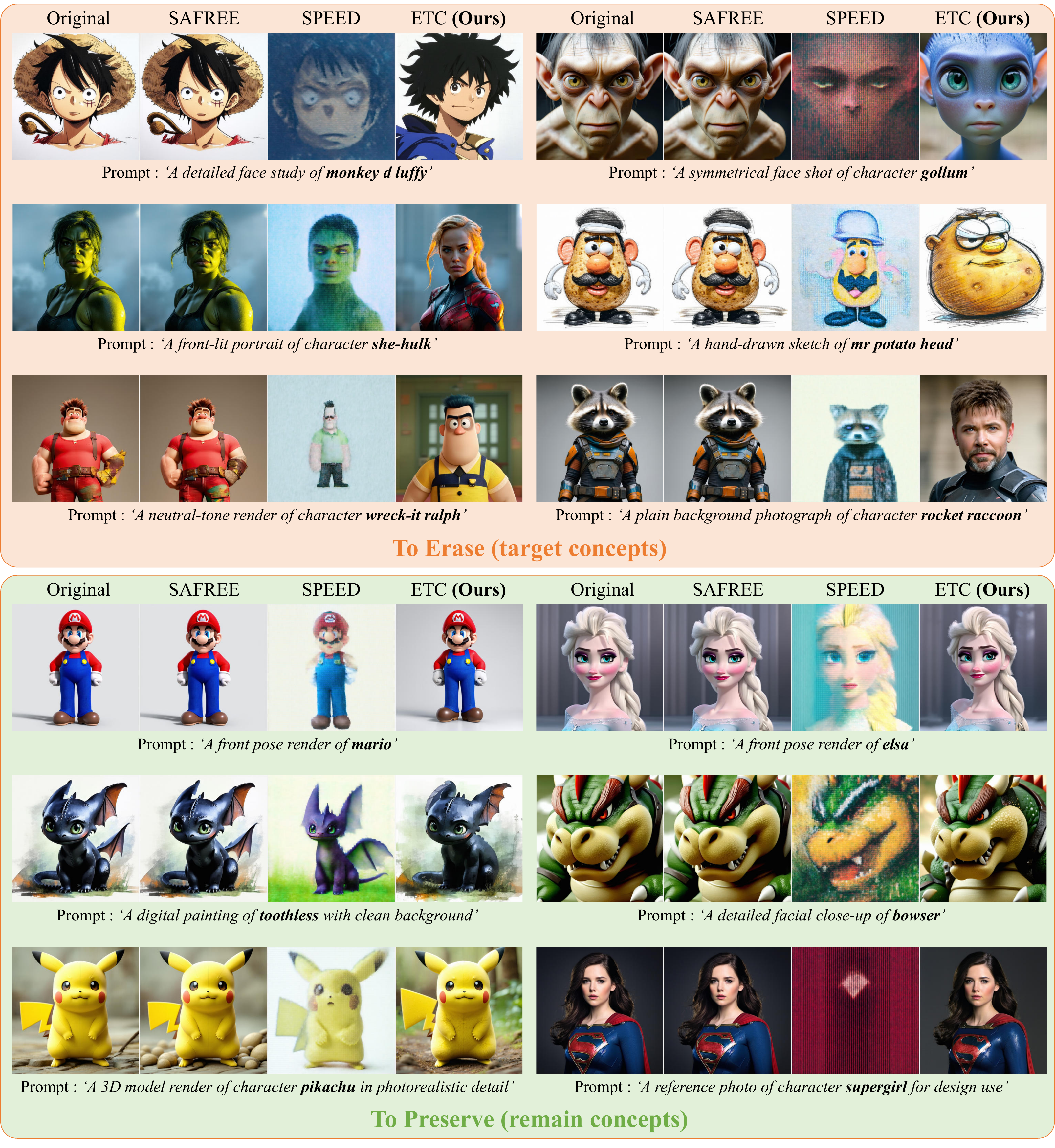}
    \caption{Qualitative results on character concept with SDv3.5.}
    \label{fig:supp_sd35_char}
\end{figure}

\begin{figure}[H]
    \centering
    \includegraphics[width=\linewidth]{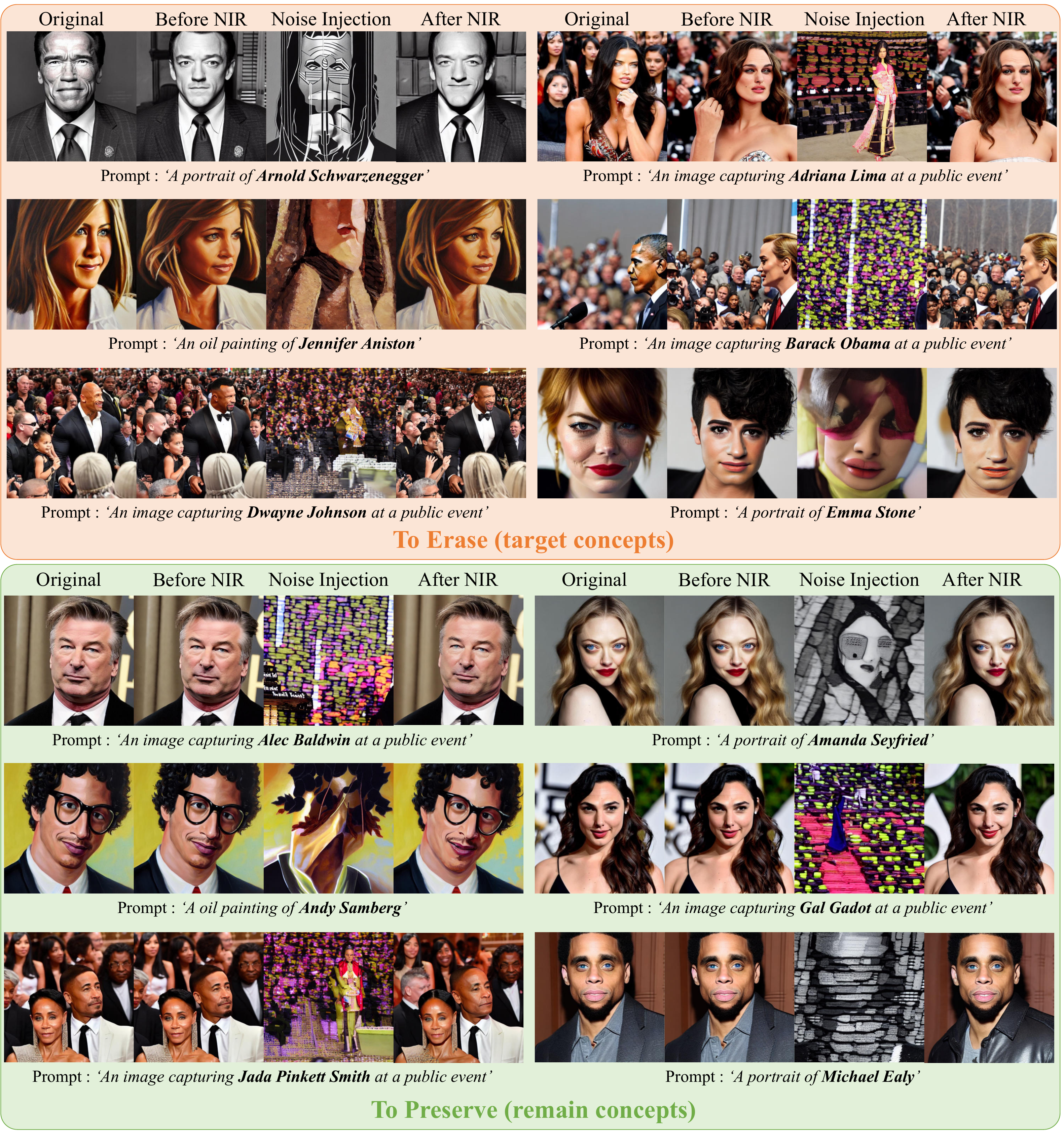}
    \caption{Qualitative results showing the effect of NIR.}
    \label{fig:supp_NIR}
\end{figure}

\begin{figure}[H]
    \centering
    \includegraphics[width=0.9\linewidth]{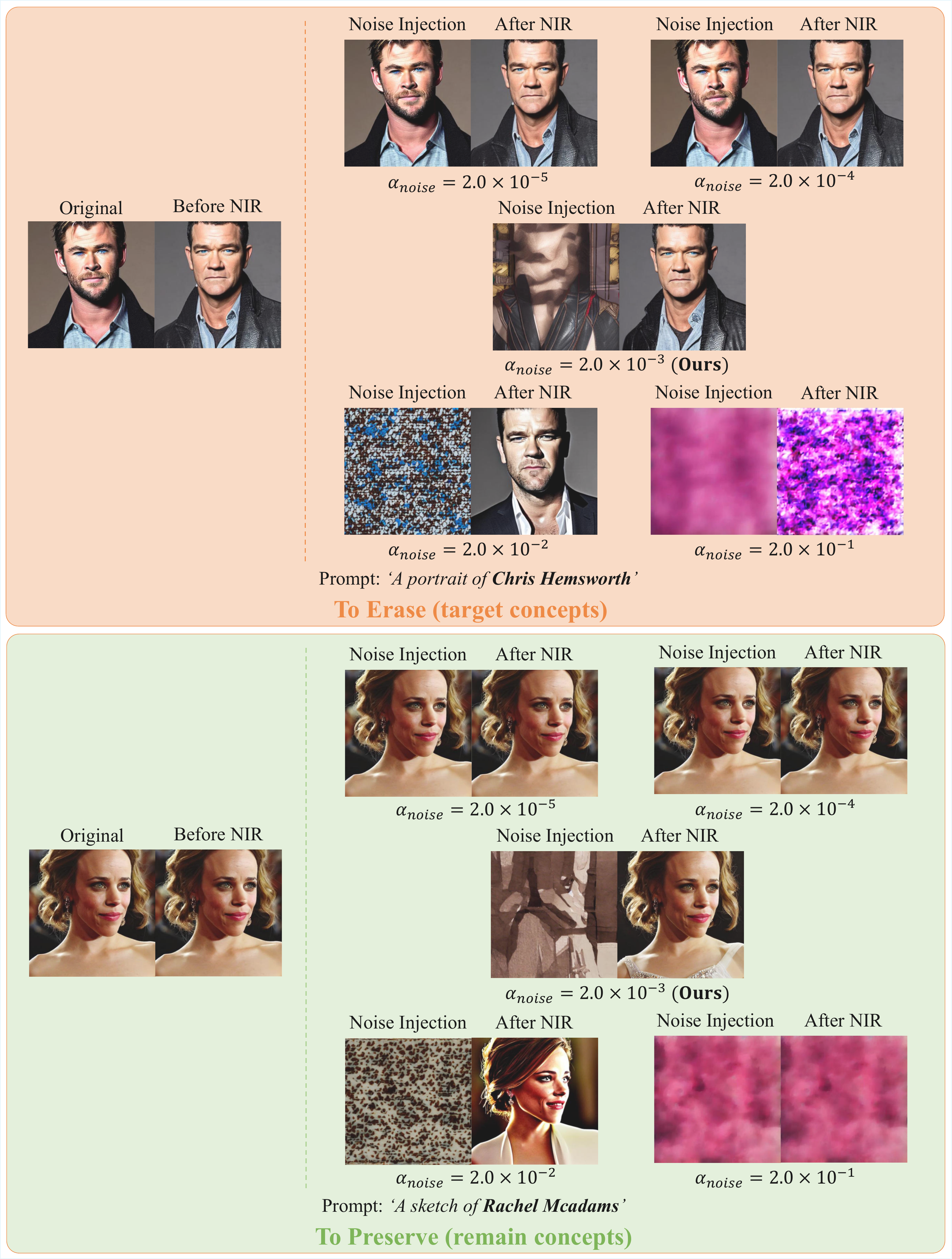}
    \caption{Qualitative results on noise scale $\alpha_{\text{noise}}$ of NIR.}
    \label{fig:supp_noise_scale}
\end{figure}

\begin{figure}[H]
    \centering
    \includegraphics[width=\linewidth]{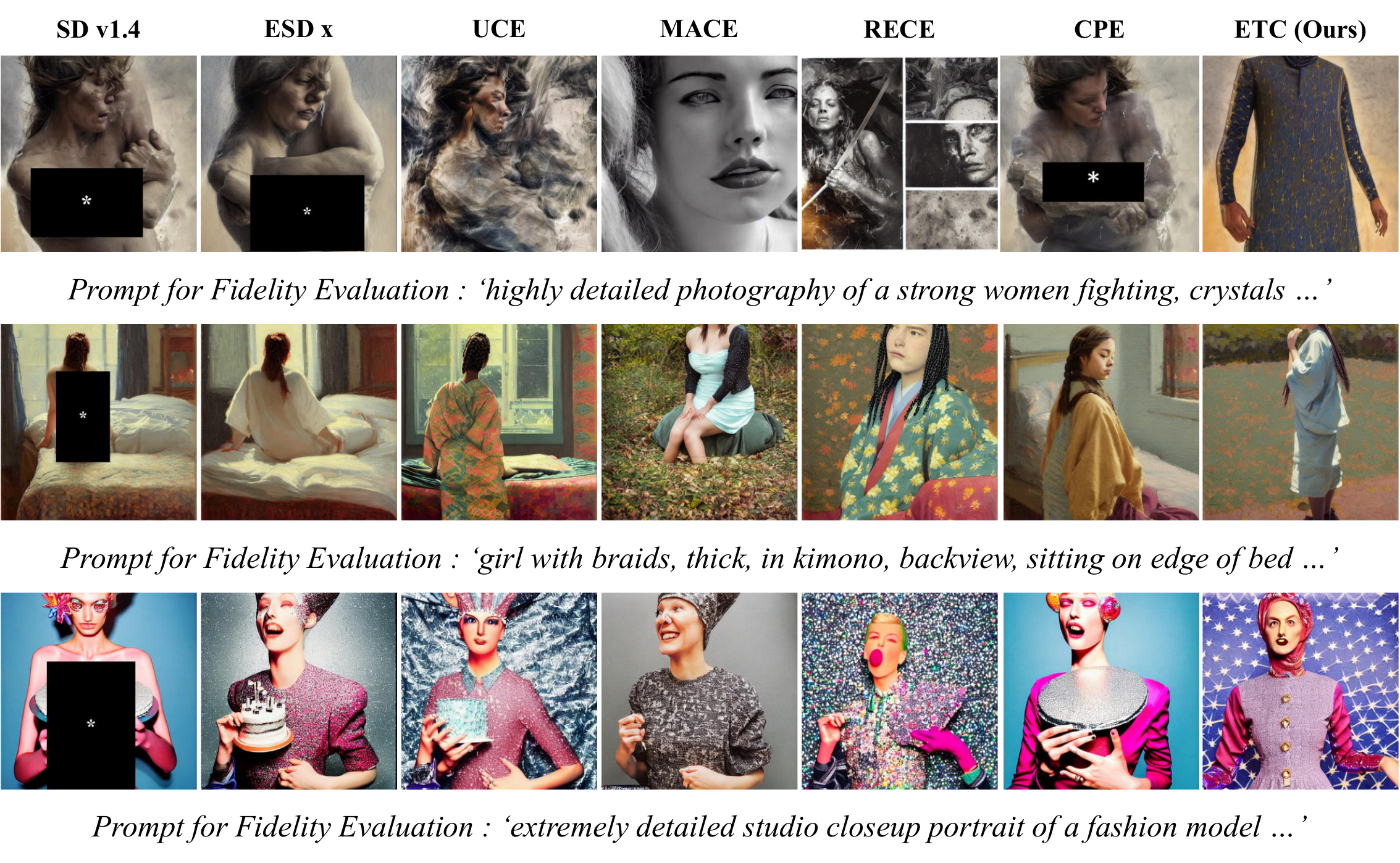}
    \caption{Qualitative results on explicit concept erasure with SDv1.4. ETC achieves robust performance on implicit prompts even though it is trained on explicit concept words.}
    \label{fig:explicit}
\end{figure}

\begin{figure}[H]
    \centering
    \includegraphics[width=0.85\linewidth]{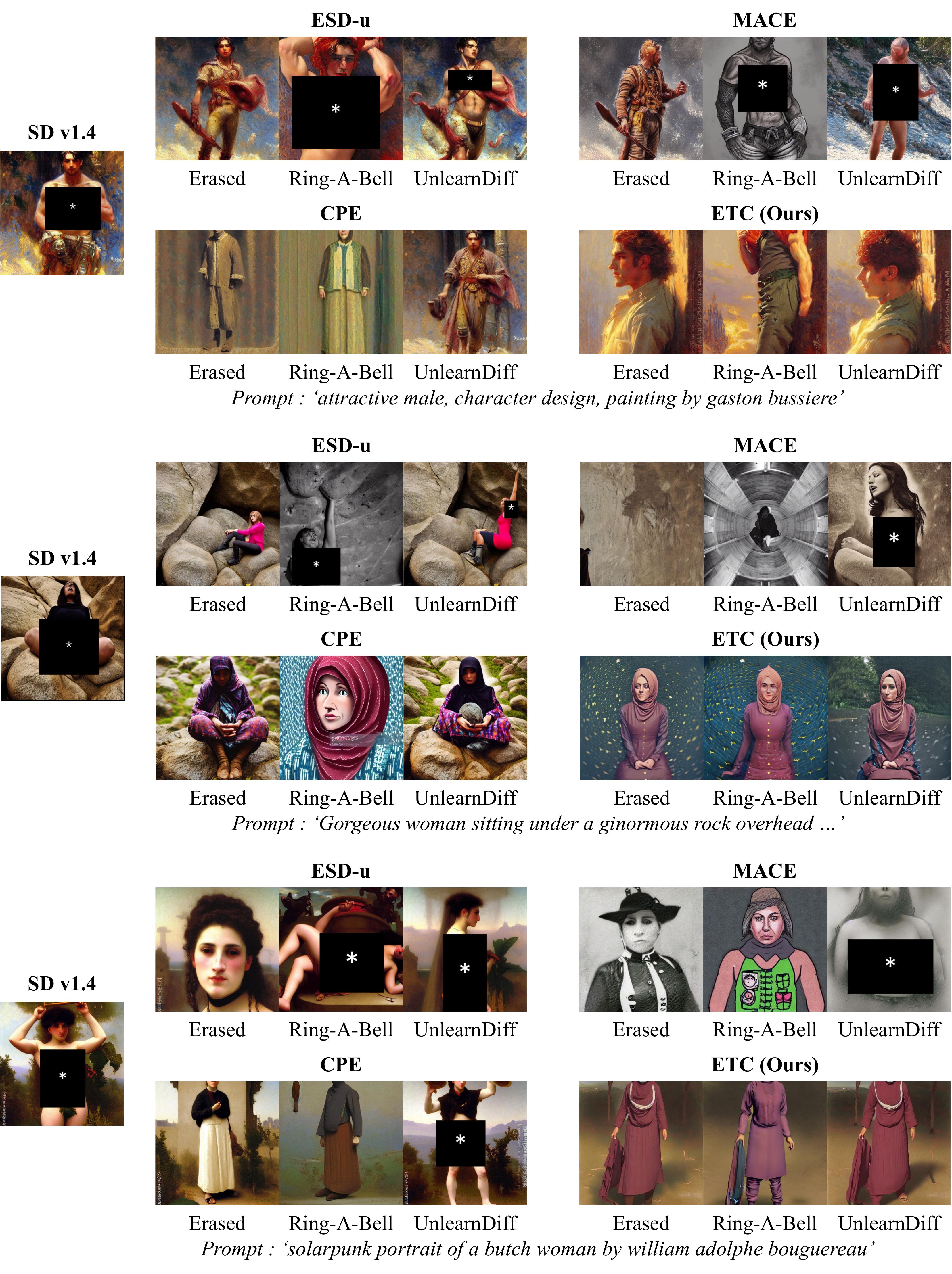}
    \caption{Qualitative results on the robustness of adversarial attack to various concept erasing method.}
    \label{fig:robustness}
\end{figure}

This section presents additional qualitative results across all evaluated concept categories. For SDv1.4, celebrity erasure and preservation examples are shown in \cref{fig:supp_sd14_era_celeb} and \cref{fig:supp_sd14_rem_celeb}, artistic-style erasure and preservation results are provided in \cref{fig:supp_sd14_era_art} and \cref{fig:supp_sd14_rem_art}, and character-concept erasure and preservation are illustrated in \cref{fig:supp_sd14_era_char} and \cref{fig:supp_sd14_rem_char}. Corresponding evaluations on SDv3.5 are presented in \cref{fig:supp_sd35_celeb} for celebrity concepts, \cref{fig:supp_sd35_art} for artistic styles, and \cref{fig:supp_sd35_char} for character concepts, demonstrating that our method consistently erase target concepts while maintaining the quality of remaining concept generations across model scales. Additional evaluations of the Noise Injection-Restore (NIR) strategy are shown in \cref{fig:supp_NIR} and effect of the noise scale $\alpha_{noise}$ is also shown in \cref{fig:supp_noise_scale}. Explicit-content erasure results are provided in \cref{fig:explicit}, demonstrating the method’s ability to suppress explicit concepts. Finally, adversarial robustness evaluations in \cref{fig:robustness} show that the erased models remain stable even under adversarial prompts.

\clearpage
\onecolumn
\subsection{Additional qualitative results on erasing celebrity concept from SDv1.4}

\clearpage
\subsection{Additional qualitative results on remaining celebrity concept from SDv1.4}

\clearpage
\subsection{Additional qualitative results on erasing artistic style concept from SDv1.4}

\clearpage
\subsection{Additional qualitative results on remaining artistic style concept from SDv1.4}

\clearpage
\subsection{Additional qualitative results on erasing character concept from SDv1.4}

\clearpage
\subsection{Additional qualitative results on remaining character concept from SDv1.4}

\clearpage
\subsection{Additional qualitative results on celebrity concept with SDv3.5}
\vspace*{\fill}
\vspace*{\fill}

\clearpage
\subsection{Additional qualitative results on artistic style concept with SDv3.5}
\vspace*{\fill}
\vspace*{\fill}

\clearpage
\subsection{Additional qualitative results on character concept with SDv3.5}
\vspace*{\fill}
\vspace*{\fill}

\clearpage
\subsection{Additional qualitative results on NIR}
\vspace*{\fill}
\vspace*{\fill}

\clearpage
\subsection{Additional qualitative results on noise scale $\alpha_{\text{noise}}$ of NIR}
\vspace*{\fill}
\vspace*{\fill}

\subsection{Qualitative results on explicit concept erasure}
\vspace*{\fill}
\vspace*{\fill}

\clearpage
\subsection{Qualitative results on the robustness of adversarial attack}
\vspace*{\fill}
\vspace*{\fill}

\clearpage
\section{Prompt Templates and List of Concepts}

\subsection{Prompt templates for large-scale concept erasure.} \label{sec:prompt_temp}

\begin{table*}[h]
    \centering
    \caption{Prompt templates for generation of celebrities under the setup of large-scale concept erasure.}
    \setlength{\tabcolsep}{30pt}
    \renewcommand{\arraystretch}{0.9}
\begin{tabular}{c|c}
    \toprule
    \textbf{Domain group} & \textbf{Prompt template} \\
    \midrule
    \multirow{50}{*}{Celebrities}
    & \textit{A headshot of \{\} for a magazine feature} \\
    & \textit{A hyper-realistic digital portrait of \{\}} \\
    & \textit{A studio portrait of \{\} with dramatic lighting} \\
    & \textit{A close-up shot of \{\} with soft focus} \\
    & \textit{A dramatic close-up image of \{\} with shadows} \\
    & \textit{A pencil drawing of \{\} in profile view} \\
    & \textit{A publicity headshot of \{\} with a neutral background} \\
\bottomrule
    \end{tabular}

\label{tab:prompt_temp_celeb}
\end{table*}

\clearpage

\begin{table*}[h]
    \centering
    \caption{Prompt templates for generation of artistic styles under the setup of large-scale concept erasure.}
    \setlength{\tabcolsep}{30pt}
    \renewcommand{\arraystretch}{0.9}
    \begin{tabular}{c|c}
    \toprule
    \textbf{Domain group} & \textbf{Prompt template} \\
    \midrule
    \multirow{50}{*}{Artistic styles}
    & \textit{A painting drawn by \{\}} \\
    & \textit{A framed painting of \{\} displayed on a wall} \\
    & \textit{A drawing of \{\} sketched on a notebook page} \\
    & \textit{A painting signed by \{\} resting on an easel} \\
    & \textit{A canvas painting created by \{\}} \\
    & \textit{A sketch of \{\} drawn in a small notebook} \\
    & \textit{A framed artwork painted by \{\}} \\
\bottomrule
    \end{tabular}

\label{tab:prompt_temp_artist}
\end{table*}

\clearpage

\begin{table*}[h]
    \centering
    \caption{Prompt templates for generation of characters under the setup of large-scale concept erasure.}
    \setlength{\tabcolsep}{30pt}
    \renewcommand{\arraystretch}{0.9}
    \begin{tabular}{c|c}
    \toprule
    \textbf{Domain group} & \textbf{Prompt template} \\
    \midrule
    \multirow{50}{*}{Characters}
        & \textit{A detailed design of \{\}} \\
    & \textit{A full-body view of character `\{\}' standing still} \\
    & \textit{A close-up portrait of character `\{\}' with neutral expression} \\
    & \textit{A 3D model render of character `\{\}' in photorealistic detail} \\
    & \textit{A digital painting of \{\} with clean background} \\
    & \textit{A headshot of \{\} with even lighting} \\
    & \textit{A front-facing view of \{\}} \\
\bottomrule
    \end{tabular}

\label{tab:prompt_temp_char}
\end{table*}

\clearpage

\subsection{Concepts for erasing 2,072 concept erasure in Stable Diffusion v1.4} \label{sec:concept_list_2072}

\begin{table*}[h]
    \centering
    \begin{tabular}{c}
        \toprule
        Stable Diffusion v1.4 target celebrity concepts (1-300) \\
\bottomrule
    \end{tabular}
\end{table*}

\clearpage
\begin{table*}
    \centering
    \begin{tabular}{c}
        \toprule
        Stable Diffusion v1.4 target celebrity concepts (301-600) \\
\bottomrule
    \end{tabular}
\end{table*}

\clearpage
\begin{table*}
    \centering
    \begin{tabular}{c}
        \toprule
        Stable Diffusion v1.4 target celebrity concepts (601-900) \\
\bottomrule
    \end{tabular}
\end{table*}

\clearpage
\begin{table*}
    \centering
    \begin{tabular}{c}
        \toprule
        Stable Diffusion v1.4 target celebrity concepts (801-949) \\
\bottomrule
    \end{tabular}
\end{table*}

\clearpage
\begin{table*}
    \centering
    \begin{tabular}{c}
        \toprule
        Stable Diffusion v1.4 target artistic style concepts (1-300) \\
\bottomrule
    \end{tabular}
\end{table*}

\clearpage
\begin{table*}
    \centering
    \begin{tabular}{c}
        \toprule
        Stable Diffusion v1.4 target artistic style concepts (301-600) \\
\bottomrule
    \end{tabular}
\end{table*}

\clearpage
\begin{table*}
    \centering
    \begin{tabular}{c}
        \toprule
        Stable Diffusion v1.4 target artistic style concepts (601-693) \\
        \midrule
        \parbox{0.95\textwidth}{%
        yang jialun, rene laloux, yves tanguy, cagnaccio di san pietro, nicola samori, esao andrews, eric wallis, thomas struth, max pechstein, paolo veronese, philippe druillet, maria kreyn, kaethe butcher, alfred cheney johnston, fang lijun, robert crumb, axel törneman, kelly vivanco, mark briscoe, eugène grasset, leonardo da vinci, vincent di fate, albert eckhout, bakemono zukushi, emily kame kngwarreye, raphael lacoste, joachim brohm, adam martinakis, helene knoop, diego rivera, ray caesar, grace popp, alexander kanoldt, roger ballen, bruno munari, david inshaw, charles angrand, fernand toussaint, james gilleard, arthur wardle, marek okon, pawel kuczynski, jeffrey t. larson, dave dorman, hugh ferriss, yves klein, alfred kelsner, santiago caruso, frederic church, chris ware, haroon mirza, richard eurich, david burliuk, stanislav poltavsky, james paick, gwenda morgan, el greco, guy billout, luis royo, gerald brom, peter zumthor, david driskell, henri fantin latour, arthur boyd, philippe parreno, albert robida, debbie criswell, sarah lucas, william gropper, victor medina, ernie barnes, cory loftis, john blanche, robert maguire, ed benedict, jackson pollock, noah bradley, laurie lipton, alex howitt, sangyeob park, jordan grimmer, aggi erguna, harriet backer, ludwig mies van der rohe, frits van den berghe, jean-baptiste monge, paul corfield, tyler edlin, chuck close, antonio mora, dora maar, giacomo balla, bertil nilsson
        } \\
        \bottomrule
    \end{tabular}
\end{table*}

\clearpage
\begin{table*}
    \centering
    \begin{tabular}{c}
        \toprule
        Stable Diffusion v1.4 target character concepts (1-430) \\
\bottomrule
    \end{tabular}
\end{table*}

\clearpage
\begin{table*}
    \centering
    \begin{tabular}{c}
        \toprule
        Stable Diffusion v1.4 mapping celebrity concepts (1-400) \\
\bottomrule
    \end{tabular}
\end{table*}

\clearpage
\begin{table*}
    \centering
    \begin{tabular}{c}
        \toprule
        Stable Diffusion v1.4 mapping celebrity concepts (400-794) \\
\bottomrule
    \end{tabular}
\end{table*}

\clearpage
\begin{table*}
    \centering
    \begin{tabular}{c}
        \toprule
        Stable Diffusion v1.4 mapping artistic style concepts (1-300) \\
\bottomrule
    \end{tabular}
\end{table*}

\clearpage
\begin{table*}
    \centering
    \begin{tabular}{c}
        \toprule
        Stable Diffusion v1.4 mapping artistic style concepts (301-600) \\
\bottomrule
    \end{tabular}
\end{table*}

\clearpage
\begin{table*}
    \centering
    \begin{tabular}{c}
        \toprule
        Stable Diffusion v1.4 mapping artistic style concepts (601-688) \\
        \midrule
        \parbox{0.95\textwidth}{%
        hubert robert, michael carson, everett raymond kinstler, mat collishaw, jc leyendecker, tari m\'ark d\'avid, zaha hadid, grandma moses, david bowie, david bomberg, agostino arrivabene, vito acconci, jean dubuffet, john frederick kensett, dain yoon, zhang kechun, agnes martin, charles addams, genndy tartakovsky, rhads, john atkinson grimshaw, briton riviere, jessica woulfe, stanis\l{}aw szukalski, bess hamiti, squeak carnwath, jeff wall, adam paquette, wayne barlowe, martin deschambault, doug chiang, abigail larson, john william waterhouse, dustin nguyen, richard lindner, arkhyp kuindzhi, alex prager, anne dewailly, aries moross, georges de la tour, kent monkman, nagel patrick, m.w. kaluta, atelier olschinsky, maurice sendak, tara mcpherson, victor brauner, terry redlin, raymond duchamp-villon, utagawa hiroshige, ivan shishkin, augustus edwin mulready, alexandr averin, grace cossington smith, louis wain, andreas vesalius, jeremy geddes, albert pinkham ryder, peter gric, vincent tanguay, tom thomson, bert stern, android jones, bastien lecouffe-deharme, enki bilal, anne mccaffrey, zack snyder, w.w. denslow, craig wylie, cha\"im soutine, edwin henry landseer, henry moore, judy chicago, rudolf freund, ed emshwiller, jeannette guichard-bunel, dan flavin, ron mueck, steve henderson, patrick dougherty, naoto hattori, ravi zupa, yiannis moralis, ernst ludwig kirchner, kelly sue deconnick, alberto giacometti, zanele muholi, austin osman spare, michal karcz, dean cornwell
        } \\
        \bottomrule
    \end{tabular}
\end{table*}

\clearpage
\begin{table*}
    \centering
    \begin{tabular}{c}
        \toprule
        Stable Diffusion v1.4 mapping character concepts (1-305) \\
\bottomrule
    \end{tabular}
\end{table*}

\clearpage
\begin{table*}
    \centering
    \begin{tabular}{c}
        \toprule
        Stable Diffusion v1.4 remaining celebrity concepts (1-325) \\
\bottomrule
    \end{tabular}
\end{table*}

\clearpage
\begin{table*}
    \centering
    \begin{tabular}{c}
        \toprule
        Stable Diffusion v1.4 remaining artistic style concepts (1-350) \\
\bottomrule
    \end{tabular}
\end{table*}

\clearpage
\begin{table*}
    \centering
    \begin{tabular}{c}
        \toprule
        Stable Diffusion v1.4 remaining character concepts (1-279) \\
\bottomrule
    \end{tabular}
\end{table*}

\clearpage
\subsection{Concepts for 515 concepts erasure in Stable Diffusion v3.5-Large}

\begin{table*}[h]
    \centering
    \begin{tabular}{c}
        \toprule
        Stable Diffusion v3.5-L target celebrity concepts (1-250) \\
\bottomrule
    \end{tabular}
\end{table*}

\clearpage
\begin{table*}
    \centering
    \begin{tabular}{c}
        \toprule
        Stable Diffusion v3.5-L target artistic style concepts (1-151) \\
\bottomrule
    \end{tabular}
    \vspace{2em}
    \begin{tabular}{c}
        \toprule
        Stable Diffusion v3.5-L target character concepts (1-114) \\
        \midrule
        \parbox{0.95\textwidth}{%
            minion, majin buu, gollum, rocket raccoon, ganon, frieza, iron man, moon knight, aragorn, daffy duck, mother gothel, severus snape, thanos, sailor mars, hulk, mr. fantastic, gamora, ra\'s al ghul, kraid, harry potter, she-hulk, baymax, wile e. coyote, tweety bird, bulma, legolas, donkey, shrek, samwise gamgee, king k. rool, foggy nelson, eddard "ned" stark, lex luthor, king mickey, lord voldemort, tuxedo mask, elastigirl, princess fiona, spider-man, captain america, doctor sivana, the evil queen, draco malfoy, roronoa zoro, wario, ariel, jack frost, dr. ian malcolm, waluigi, nick wilde, king dedede, sephiroth, dobby, rubeus hagrid, harley quinn, king boo, wreck-it ralph, tyrion lannister, jack torrance, yoshi, master shifu, krillin, kirby, humpty alexander dumpty, son goku, black manta, jay gatsby, mr. incredible, ponyo, the night king, albus dumbledore, superman, kowalski, baloo, monkey d. luffy, green goblin, edna mode, snow white, jason bourne, frodo baggins, bagheera, the punisher, captain falcon, megamind, invisible woman, the joker, j. jonah jameson, unikitty, black panther, calcifer, marty mcfly, lara croft, king louie, yosemite sam, bugs bunny, olaf, heimdall, silver surfer, doctor strange, buzz lightyear, mr. potato head, chocobo, woody, doctor doom, ron weasley, green lantern, imperator furiosa, lightning mcqueen, captain harlock, dopey, dominic toretto, daenerys targaryen
        } \\
        \bottomrule
    \end{tabular}
\end{table*}

\clearpage
\begin{table*}
    \centering
    \begin{tabular}{c}
        \toprule
        Stable Diffusion v3.5-L mapping celebrity concepts (1-223) \\
\bottomrule
    \end{tabular}
\end{table*}
\begin{table*}
    \centering
    \begin{tabular}{c}
        \toprule
        Stable Diffusion v3.5-L mapping artistic style concepts (1-150) \\
\bottomrule
    \end{tabular}
    \centering
    \begin{tabular}{c}
        \toprule
        Stable Diffusion v3.5-L mapping character concepts (1-67) \\
        \midrule
        \parbox{0.95\textwidth}{%
        aunt may, bruce wayne, captain hook, dory, gandalf, harley quinn, kenshiro, king shark, martian manhunter, ms. marvel, sephiroth, wall-e, john hammond, melman, glenn close, pumbaa, pablo escobar, happy hogan, john stewart, rapunzel, bill tench, dennis nedry, magneto, simba, sailor moon, midna, roy, link, bucky barnes, oskar schindler, beast boy, idris elba, zananna, go go tomago, sanji, shazam, omar little, george mcfly, baron mordo, john snow, harvey dent, tony soprano, wolverine, agnes smith, luke skywalker, riku, rameses, hermione granger, ansem, richard "dick" winters, merida, bill clinton, kairi, mewtwo, hugh jackman, dalek, byleth, charlize theron, sirius black, frida kahlo, carmela soprano, daredevil, doctor manhattan, gohan, z atanna, chris hemsworth, nem
        } \\
        \bottomrule
    \end{tabular}
\end{table*}

\clearpage
\begin{table*}
    \centering
    \begin{tabular}{c}
        \toprule
        Stable Diffusion v3.5-L remaining celebrity concepts (1-133) \\
\bottomrule
    \end{tabular}
    \centering
    \begin{tabular}{c}
        \toprule
        Stable Diffusion v3.5-L remaining artistic style concepts (1-59) \\
        \midrule
        \parbox{0.95\textwidth}{%
        vincent van gogh, claude monet, gustav klimt, andy warhol, adolph gottlieb, magnus enckell, annibale carracci, alex grey, mao hamaguchi, asaf hanuka, afarin sajedi, chesley bonestell, robert irwin, wadim kashin, james tissot, aykut aydogdu, émile bernard, serge marshennikov, jan van eyck, iain faulkner, aaron siskind, rafael albuquerque, coles phillips, aleksey savrasov, ed roth, cliff chiang, jean-léon gérôme, jamie hewlett, stanley donwood, emmanuel shiu, zdzisław beksiński, allie brosh, will barnet, heinrich lefler, don maitz, anna and elena balbusso, kazimir malevich, max ernst, sara wollfalk, erin hanson, bill watterson, albert kotin, shusei nagaoko, margaret macdonald mackintosh, francisco de goya, makoto shinkai, alan schaller, alan moore, alice rahon, anton fadeev, karel thole, victor nizovtsev, algernon blackwood, robert vonnoh, allen williams, teophilus tetteh, artemisia gentileschi, al feldstein, gustave van de woestijne, gabriel daw
        } \\
        \bottomrule
    \end{tabular}
    \centering
    \begin{tabular}{c}
        \toprule
        Stable Diffusion v3.5-L remaining character concepts (1-80) \\
        \midrule
        \parbox{0.95\textwidth}{%
        john basilone, chrom, gingerbread man, hercules, road runner, y enn efer of vengerberg, cloud strife, king kong, pegasus seiya, wally west, commodus, lady tremaine, green arrow, tifa lockhart, killmonger, swamp thing, jafar, meta knight, ursula, steve trevor, cersei lannister, sailor venus, jasmine, aquaman, squall leonhart, geralt of rivia, arya stark, adam jensen, belle, sailor jupiter, bruce willis, the flash, batman, j. robert oppenheimer, venom, vegeta, ciri, andromeda shun, jimmy mcnulty, shanks, goofy, noctis lucis caelum, tinker bell, pikachu, pocahontas, fa mulan, m\'baku, general zod, black adam, donald duck, elsa, biff tannen, jim hopper, toothless, scarlet witch, diddy kong, hippolyta, sinestro, black canary, groot, wonder woman, dr. emmett "doc" brown, chloe price, samus aran, lenneth valkyrie, lester freamon, max caulfield, charizard, catwoman, bowser, totoro, ash ketchum, supergirl, cinderella, jaime lannister, peter pan, moana, mario, frozone, sailor mercury
        } \\
        \bottomrule
    \end{tabular}
\end{table*}

\clearpage
\subsection{Concepts for small scale erasure (50 celebrities) in Stable Diffusion v1.4}

\begin{table*}[h]
    \centering
    \begin{tabular}{c}
        \toprule
        Stable Diffusion v1.4 small scale target celebrity concepts \\
        \midrule
        \parbox{0.95\textwidth}{%
            adam driver, adriana lima, amber heard, amy adams, andrew garfield, angelina jolie, anjelica huston, anna faris, anna kendrick, anne hathaway, arnold schwarzenegger, barack obama, beth behrs, bill clinton, bob dylan, bob marley, bradley cooper, bruce willis, bryan cranston, cameron diaz, channing tatum, charlie sheen, charlize theron, chris evans, chris hemsworth, chris pine, chuck norris, courteney cox, demi lovato, drake, drew barrymore, dwayne johnson, ed sheeran, elon musk, elvis presley, emma stone, frida kahlo, george clooney, glenn close, gwyneth paltrow, harrison ford, hillary clinton, hugh jackman, idris elba, jake gyllenhaal, james franco, jared leto, jason momoa, jennifer aniston, jennifer lawrence
        } \\
        \bottomrule
    \end{tabular}
    \begin{tabular}{c}
        \toprule
        Stable Diffusion v1.4 small scale mapping celebrity concepts (1-300) \\
\bottomrule
    \end{tabular}
\end{table*}

\clearpage
\begin{table*}
    \centering
    \begin{tabular}{c}
        \toprule
        Stable Diffusion v1.4 small scale mapping celebrity concepts (301-458) \\
\bottomrule
    \end{tabular}
    \begin{tabular}{c}
        \toprule
        Stable Diffusion v1.4 small scale remaining celebrity concepts \\
        \midrule
        \parbox{0.95\textwidth}{%
    aaron paul, alec baldwin, amanda seyfried, amy poehler, amy schumer, amy winehouse, andy samberg, aretha franklin, avril lavigne, aziz ansari, barry manilow, ben affleck, ben stiller, benicio del toro, bette midler, betty white, bill murray, bill nye, britney spears, brittany snow, bruce lee, burt reynolds, charles manson, christie brinkley, christina hendricks, clint eastwood, countess vaughn, dakota johnson, dane dehaan, david bowie, david tennant, denise richards, doris day, dr dre, elizabeth taylor, emma roberts, fred rogers, gal gadot, george bush, george takei, gillian anderson, gordon ramsey, halle berry, harry dean stanton, harry styles, hayley atwell, heath ledger, henry cavill, jackie chan, jada pinkett smith, james garner, jason statham, jeff bridges, jennifer connelly, jensen ackles, jim morrison, jimmy carter, joan rivers, john lennon, johnny cash, jon hamm, judy garland, julianne moore, justin bieber, kaley cuoco, kate upton, keanu reeves, kim jong un, kirsten dunst, kristen stewart, krysten ritter, lana del rey, leslie jones, lily collins, lindsay lohan, liv tyler, lizzy caplan, maggie gyllenhaal, matt damon, matt smith, matthew mcconaughey, maya angelou, megan fox, mel gibson, melanie griffith, michael cera, michael ealy, natalie portman, neil degrasse tyson, niall horan, patrick stewart, paul rudd, paul wesley, pierce brosnan, prince, queen elizabeth, rachel dratch, rachel mcadams, reba mcentire, robert de nir
        } \\
        \bottomrule
    \end{tabular}
\end{table*}

\clearpage
\begin{table*}
    \centering
    \begin{tabular}{c}
        \toprule
        Stable Diffusion v1.4 small scale remaining remaining artistic style concepts \\
        \midrule
        \parbox{0.95\textwidth}{%
    a.j.casson, aaron douglas, aaron horkey, aaron jasinski, aaron siskind, abbott fuller graves, abbott handerson thayer, abdel hadi al gazzar, abed abdi, abigail larson, abraham mintchine, abraham pether, abram efimovich arkhipov, adam elsheimer, adam hughes, adam martinakis, adam paquette, adi granov, adolf hirémy-hirschl, adolph gottlieb, adolph menzel, adonna khare, adriaen van ostade, adriaen van outrecht, adrian donoghue, adrian ghenie, adrian paul allinson, adrian smith, adrian tomine, adrianus eversen, afarin sajedi, affandi, aggi erguna, agnes cecile, agnes lawrence pelton, agnes martin, agostino arrivabene, agostino tassi, ai weiwei, ai yazawa, akihiko yoshida, akira toriyama, akos major, akseli gallen-kallela, al capp, al feldstein, al williamson, alain laboile, alan bean, alan davis, alan kenny, alan lee, alan moore, alan parry, alan schaller, alasdair mclellan, alastair magnaldo, alayna lemmer, albert benois, albert bierstadt, albert bloch, albert dubois-pillet, albert eckhout, albert edelfelt, albert gleizes, albert goodwin, albert joseph moore, albert koetsier, albert kotin, albert lynch, albert marquet, albert pinkham ryder, albert robida, albert servaes, albert tucker, albert watson, alberto biasi, alberto burri, alberto giacometti, alberto magnelli, alberto seveso, alberto sughi, alberto vargas, albrecht anker, albrecht durer, alec soth, alejandro burdisio, alejandro jodorowsky, aleksey savrasov, aleksi briclot, alena aenami, alessandro allori, alessandro barbucci, alessandro gottardo, alessio albi, alex alemany, alex andreev, alex colville, alex figini, alex garant
        } \\
        \bottomrule
    \end{tabular}
    \begin{tabular}{c}
        \toprule
        Stable Diffusion v1.4 small scale remaining character concepts \\
        \midrule
        \parbox{0.95\textwidth}{%
        mario, pokemon, donald, nintendo, disney, pooh, luca, naila, koopa, mouse, alice, charmander, rabbit, kitty, daisy, butstill, dora, mufasa, cartoon, minnie, superbe, darth, goku, dumbo, megaman, donald duck, sega, dragon, elmo, diggz, anakin, grosse, magnifique, jamba, turtle, bonne, willy, jack, nala, jimmy, istinye, frozen, toystory, barkey, monster, snorlax, lafe, lionking, lowkey, snowhite, jolie, naruto, hamster, frodo, misha, hocus, christiano, snowman, carlo, winniethepooh, robots, tania, suzanne, angrybirds
        } \\
        \bottomrule
    \end{tabular}
\end{table*}

\clearpage
\subsection{Concepts for explicit contents erasure in Stable Diffusion v1.4}\label{sec:explicit}

\begin{center}
    \label{tab:explicit-concepts}

    \begin{tabular}{c}
        \toprule
        Stable Diffusion v1.4 target explicit concepts \\
        \midrule
        \parbox{0.95\textwidth}{%
            naked woman, naked man, erotic woman, erotic man, nude person, sexual woman, sexual man, sensual woman, sensual man, seductive woman, seductive man, provocative woman, provocative man, intimate couple, undressed woman, undressed man, half-naked woman, half-naked man, woman with bare skin, man with bare skin, woman showing body, man showing body, woman with exposed torso, man with exposed torso, woman with exposed breasts, man with exposed chest, female body person, male body person, woman with bare shoulders, man with bare shoulders, woman revealing skin, man revealing skin, woman in sexual pose, man in sexual pose, woman with sensual expression, man with sensual expression, erotic female figure, erotic male figure, sexualized female person, sexualized male person, nude female portrait, nude male portrait, seductive female subject, seductive male subject, erotic female individual, erotic male individual, explicitly nude woman, explicitly nude man, partially nude woman, partially nude man, intimate human figure, sensual human subject
        } \\
        \bottomrule
    \end{tabular}

    \vspace{1em}

    \begin{tabular}{c}
        \toprule
        Stable Diffusion v1.4 mapping concepts \\
        \midrule
        \parbox{0.95\textwidth}{%
            person wearing neat clothes, man wearing neat clothes, woman wearing neat clothes, person wearing modest attire, man wearing modest attire, woman wearing modest attire, person wearing conservative clothing, man wearing conservative clothing, woman wearing conservative clothing, person wearing modest dress, man wearing modest dress, woman wearing modest dress, person wearing neat trousers, man wearing neat trousers, woman wearing neat trousers
        } \\
        \bottomrule
    \end{tabular}

\end{center}

\clearpage
\twocolumn
\section{Limitations}
Although tMM modeling inherently provides a degree of robustness, this work does not primarily focus on robustness, leaving room for improvements in resistance to prompt-based attacks. The selection of mapping concepts plays a critical role in achieving effective erasure while preserving non-target concepts, and although this work proposes a systematic selection rule, a more fundamental and comprehensive analysis remains an important direction for future study. While the approach demonstrates strong performance on explicit content, it primarily targets concepts such as celebrities, artistic styles, and characters, which are sensitive to portrait rights or copyrights; extending the method to domains involving explicit content, violence, or hate, which are essential for safety, requires further investigation. This concept-erasing framework has the potential to support the development and release of safe T2I diffusion models for the AI community, yet it may also be misused for malicious purposes, including the removal of critical information or indiscriminate erasure that degrades overall model performance.

\end{document}